\documentclass[pdflatex,sn-mathphys-num]{sn-jnl}

\usepackage[utf8]{inputenc}
\usepackage{graphicx}%
\usepackage{multirow}%
\usepackage{amsmath,amssymb,amsfonts}%
\usepackage{amsthm}%
\usepackage{mathrsfs}%
\usepackage[title]{appendix}%
\usepackage{xcolor}%
\usepackage{textcomp}%
\usepackage{manyfoot}%
\usepackage{booktabs}%
\usepackage{algorithm}%
\usepackage{algorithmicx}%
\usepackage{algpseudocode}%
\usepackage{listings}%
\usepackage{lineno}
\usepackage{longtable}

\begin{document}


\title[Article Title]{Higher-Order Neuromorphic Ising Machines - Autoencoders and Fowler-Nordheim Annealers are all you need for Scalability}

\author[1]{\fnm{Faiek} \sur{Ahsan}}\email{a.faiek@wustl.edu}
\author[2]{\fnm{Saptarshi} \sur{Maiti}}\email{saptarshimai@iisc.ac.in}
\author[1]{\fnm{Zihao} \sur{Chen}}\email{czihao@wustl.edu}
\author[3]{\fnm{Jakob} \sur{Kaiser}}\email{jakob.kaiser@kip.uni-heidelberg.de}
\author[2]{\fnm{Ankita} \sur{Nandi}}\email{ankitanandi@iisc.ac.in}
\author[2]{\fnm{Madhuvanthi} \sur{Srivatsav}}\email{madhuvanthis@iisc.ac.in}
\author[4]{\fnm{Johannes} \sur{Schemmel}}\email{schemmel@kip.uni-heidelberg.de}
\author[5]{\fnm{Andreas G.} \sur{Andreou}}\email{aandreo1@jhu.edu}
\author[6]{\fnm{Jason} \sur{Eshraghian}}\email{jsn@ucsc.edu}
\author[2]{\fnm{Chetan Singh} \sur{Thakur}}\email{csthakur@iisc.ac.in}
\author*[1]{\fnm{Shantanu} \sur{Chakrabartty}}\email{shantanu@wustl.edu}

\affil[1]{\orgdiv{Department of Electrical and Systems Engineering}, \orgname{Washington University in St. Louis}, \orgaddress{\street{One Brookings Drive}, \city{St. Louis}, \postcode{63130}, \state{MO}, \country{USA}}}

\affil[2]{\orgdiv{Department of Electronic Systems Engineering}, \orgname{Indian Institute of Science}, \orgaddress{\city{Bangalore}, \postcode{560012}, \country{India}}}

\affil[3]{\orgdiv{Kirchhoff Institute for Physics}, \orgname{Heidelberg University}, \country{Germany}}
\affil[4]{\orgdiv{Institute of Computer Engineering}, \orgname{Heidelberg University}, \country{Germany}}
\affil[5]{\orgdiv{Department of Electrical and computer engineering}, \orgname{Johns Hopkins University}, \orgaddress{\street{3400 N. Charles Street}, \city{Baltimore}, \postcode{21218}, \state{MD}, \country{USA}}}
\affil[6]{\orgdiv{Department of Electrical and Computer Engineering}, \orgname{University of California, Santa Cruz}, \orgaddress{\street{1156 High Street}, \city{Santa Cruz}, \postcode{95064}, \state{CA}, \country{USA}}}

\abstract{
We report a higher-order neuromorphic Ising machine that exhibits superior scalability compared to architectures based on quadratization, while also achieving state-of-the-art quality and reliability in solutions with competitive time-to-solution metrics. At the core of the proposed machine is an asynchronous autoencoder architecture that captures higher-order interactions by directly manipulating Ising clauses instead of Ising spins, thereby maintaining resource complexity independent of interaction order. Asymptotic convergence to the Ising ground state is ensured by sampling the autoencoder latent space defined by the spins, based on the annealing dynamics of the Fowler–Nordheim quantum mechanical tunneling. To demonstrate the advantages of the proposed higher-order neuromorphic Ising machine, we systematically solved benchmark combinatorial optimization problems such as MAX-CUT and MAX-SAT, comparing the results to those obtained using a second-order Ising machine employing the same annealing process. Our findings indicate that the proposed architecture consistently provides higher quality solutions in shorter time frames compared to the second-order model across multiple runs. Additionally, we show that the techniques based on the sparsity of the interconnection matrix, such as graph coloring, can be effectively applied to higher-order neuromorphic Ising machines, enhancing the solution quality and the time-to-solution. The time-to-solution can be further improved through hardware co-design, as demonstrated in this paper using a field-programmable gate array (FPGA). The results presented in this paper provide further evidence that autoencoders and Fowler–Nordheim annealers are sufficient to achieve reliability and scaling of any-order neuromorphic Ising machines.
}

\keywords{Neuromorphic Computing, Ising Machines, Higher-order Ising, Fowler-Nordheim Annealing, MAX-CUT, MAX-SAT, Combinatorial Optimization Problems}

\maketitle

\section{Introduction}\label{sec1}

Recently, there has been considerable interest in application domains where neuromorphic architectures can demonstrate significant performance advantages compared to conventional hardware architectures based on central processing units (CPUs), graphics processing units (GPUs), and tensor processing units (TPUs)~\cite{Neural_inference_Modha,Merolla_2014,Esser2016}. Combinatorial optimization is one such domain, where the physics of the problem can be directly mapped onto a neuromorphic Ising architecture~\cite{Chen2025}. Similar to other Ising solvers based on superconducting quantum annealers~\cite{Johnson2011,King2023}, optical and CMOS oscillators~\cite{Marandi2014,McMahon2016,Wang2019,Chou2019}, and stochastic magnetic-tunnel junctions~\cite{Sutton2017,Aadit2022}, neuromorphic Ising machines capture pairwise interactions between spin variables as synaptic weights. The neurons encode the spin variables, and the attractor states of the neuronal network represent possible solutions to the underlying optimization problem~\cite{Hopfield1982,Chen2025}. When the Lyapunov dynamics of the network are combined with noise, annealing, and random fluctuations to escape local attractor states, an Ising machine epitomizes several unique and advantageous features of neuromorphic architecture compared to other hardware architectures which includes: (a) asynchronous event-based computing; (b) massive parallelism; and (c) noise-driven dynamics. As a result, the neuromorphic advantage for Ising problems has been demonstrated in terms of time-to-solution metrics~\cite{Cai2020,Maher2024CMOSVOISING}, energy-to-solution metrics~\cite{Chen2025,Cai2020,Yik2025,Maher2024CMOSVOISING} and reliability/quality of solution metrics~\cite{Chen2025}.

\begin{figure}[t]
\centering
\includegraphics[width=1\textwidth]{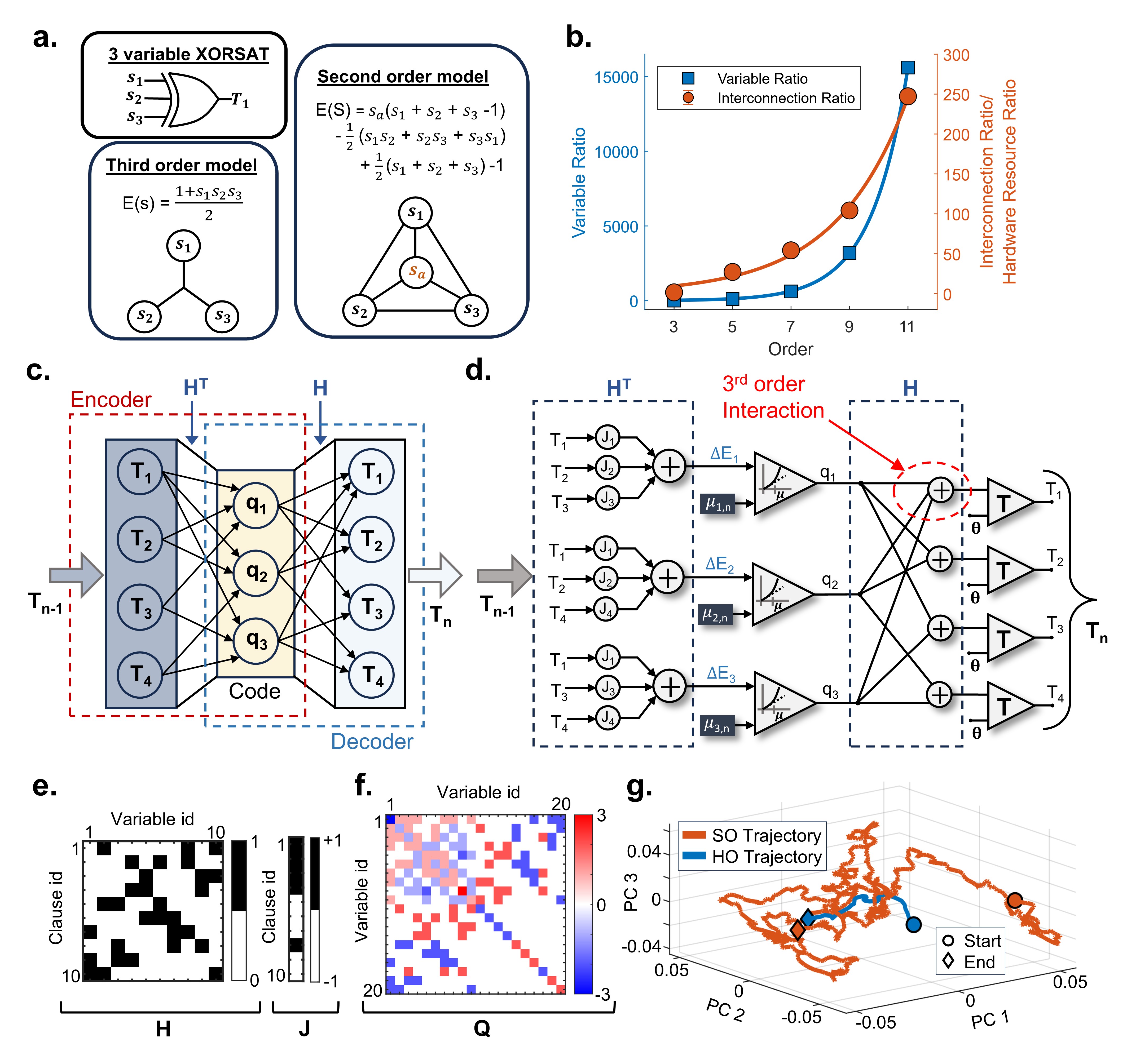}
\caption{
\textbf{Higher-order neuromorphic Ising machines based on an autoencoder architecture :}
\textbf{(a)} Mapping of a 3-spin, single clause XOR-SAT problem into a third-order Ising model. Quadratization leads to six second-order interactions and an additional auxiliary variable ($s_a$). 
\textbf{(b)} Exponential growth in the ratio of variables and the interconnection ratio in the Ising energy function, post-quadratization, compared to directly optimizing higher-order interactions. 
\textbf{(c)} An autoencoder architecture with the outer layers comprising of clause neurons with outputs $T_k$'s and the latent layer comprises of the spin neurons with outputs $q_i$'s.
\textbf{(d)} A higher-order neuromorphic Ising machine defined by the 
interconnection matrix $\mathbf{H}$, which specifies the variables involved in each clause. 
The non-linear integration is achieved by summing the clause outputs and comparing it against a noisy threshold ($\mu_{i,n}$) generated by a Fowler-Nordheim Annealer.
Spin outputs are generated by toggle neurons, which switch their states when the input exceeds a threshold $\theta$. \textbf{(e)(f)(g)} depict the higher-order dynamics for a small 3R-3X problem with 10 variables and 10 clauses. In \textbf{(e)}, the interconnection matrix $\mathbf{H}$ and weight vector $\mathbf{J}$ used to solve this problem directly using higher-order interactions. The post-quadratization Hamiltonian matrix $\mathbf{Q}$ is shown in \textbf{(f)} illustrating the increased precision and range of the resulting pairwise coupling coefficients. \textbf{(g)} shows a PCA-based trajectory of the higher-order (HO) Ising machine and a second-order (SO) Ising machine~\cite{Chen2025} when solving the same problem using an identical annealing schedule.
} \label{fig1}
\end{figure}

Most Ising machines (both neuromorphic and non-neuromorphic) support only pairwise interactions, which are straightforward to implement in hardware via resistive~\cite{Mallick2023,Sharma2023,Cai2020,Albertsson2023} and capacitive~\cite{Maher2024CMOSVOISING,Vaidya2022,Dutta2021} couplings in analog hardware, or via point-to-point connections in digital hardware~\cite{Yamaoka2016, Takemoto2019, Loihi,mayr2019spinnaker210million}. However, the direct implementation of higher-order (three-body or beyond) terms is more challenging, even though higher-order terms inevitably appear when encoding general combinatorial objectives as Ising energy polynomials~\cite{Lucas2014}. To address this, numerous quadratization techniques have been developed to reduce higher-order polynomials to second-order by introducing auxiliary variables~\cite{BOROS2002155,Freedman2005,Ishikawa2011}. From an implementation standpoint, this requires increasing both the total number of spins and the precision and dynamic range of the pairwise coupling coefficients compared to the original higher-order weights.
This concept is depicted in Fig.~\ref{fig1}.a, using a 3-variable XOR-SAT problem where the Ising Hamiltonian is described using a single third-order term. Quadratization yields a second-order network with one auxiliary spin and six pairwise couplings, as shown in Fig.~\ref{fig1}.a. While the quadratization overhead incurred for this and other small-scale examples may be modest, it grows rapidly with the order of the original optimization objective polynomial. As shown in Fig.~\ref{fig1}.b, the ratio of variables and the interconnection ratio in the Ising model's energy function (after quadratization) grow exponentially with the interaction order for MAX-SAT problems of orders 3, 5, 7, 9, and 11 respectively (see Methods Section~\ref{sec_methods_resource}). The exponential increase in the variable ratio signifies that quadratization expands the search space exponentially and may introduce additional ruggedness in the energy landscape, impacting solver performance in terms of convergence and time-to-solution. Additionally, because the interconnection ratio directly correlates with hardware resource requirements, the hardware resources to implement a quadratized Ising solver also grows exponentially. In this regard, a direct implementation of higher-order interactions (without quadratization) offers a more scalable and resource efficient solution. 

Recent approaches for implementing higher-order Ising machines are based on two computing methods: classical annealing and dynamical system evolution. Examples of these include probabilistic bit (p-bit) based higher-order Ising machines~\cite{Nikhar2024} and memristor-based higher-order Hopfield optimization solvers for classical annealing~\cite{Memristor_based_HOIM}, and coupled oscillator-based higher-order Ising machines for dynamical system evolution~\cite{Bybee2023}. Classical annealing methods like p-bit stochastic solvers relax binary constraints and leverage intrinsic noise to find solutions, using the stochastic properties of p-bits for efficient computation, and memristor-based networks exploit device stochasticity to achieve low metrics in terms of energy-to-solution and time-to-solution. Dynamical system evolution methods, such as coupled oscillator-based approaches, evolve more rapidly than thermal or adiabatic timescales, driving the system towards low-energy configurations quickly. While the benefits of these higher-order approaches can be fully realized when implemented in analog hardware, computational precision, device mismatch and dynamic range for readout circuits limit the scaling to large problems. Also, there is no theoretical assurance of outperforming current state-of-the-art solutions or consistently converging to the true ground state. Digital implementations can avoid some of these issues~\cite{Tatsumura2021,Wang2025, Yamaoka2016, Yamamoto2021} but require careful calibration of noise statistics to achieve competitive performance. A promising recent development is the neuromorphic Ising machine named {\it NeuroSA}~\cite{Chen2025}, that leverages advances in asynchronous neuromorphic hardware to achieve scalability for solving second-order Ising problems. Furthermore, the {\it NeuroSA} Ising machines provide an asymptotic guarantee of converging to the Ising ground state which leads to higher reliability in producing and discovering state-of-the-art solutions.

In this work, we propose a scalable higher-order Neuromorphic Ising machine that overcomes the limitations of quadratization. We also show that the Fowler-Nordheim annealer which was important for producing state-of-the-art solutions in the {\it NeuroSA} architecture, is also the key for the proposed higher-order Ising machine to achieve state-of-the-art results. We show this through an isomorphic functional mapping of simulated annealing to a general $N$-spin combinatorial optimization problem based on the minimization of the Hamiltonian
\begin{equation}\label{eq:General_Ising_min}
\min_{\mathbf{s}\in\{\pm1\}^N}E(\mathbf{s})
=-\sum_{k=1}^M J_k\prod_{i=1}^N s_i^{H_{k,i}}.
\end{equation}
Here $\mathbf{s}=(s_1,\dots,s_N)$ represent Ising spins and $M$ represents  the total number of interaction terms that are indexed by $k=1,\dots,M$. $J_k$ denotes the weight for $k$'th spin tuple. $\mathbf{H}\in\{0,1\}^{M\times N}$ indicates clause membership, i.e., the contributing spins in $k$'th spin tuple. For instance, the single XOR-SAT clause in Fig.~\ref{fig1}.a corresponds to a simple third-order Ising Hamiltonian. This framework was also used in~\cite{Nandi2024XORSAT} for modeling parity-check constraints. 
In the standard Ising‐based minimization of Eq.~\eqref{eq:General_Ising_min}, one explores the space of spin states to identify the configuration that yields the lowest energy. At each step of this sampling process, the change in energy, \(\Delta E\), must be recalculated for a potentially new spin configuration. For a \(p\)-th order Hamiltonian, computing \(\Delta E\) directly from the full spin vector \(\mathbf{s}\) incurs a cost of \(\mathcal{O}(n^p)\), which rapidly becomes intractable as both the total spin count \(n\) and the interaction order \(p\) increase. We note, however, that if \(\Delta E\) can be reformulated in terms of the clause outputs \(T_k\), then it becomes possible to navigate the energy landscape directly within the \(T_k\)-space ($\mathbf{T}\in\{-1,1\}^{M}$). By doing so, the exact higher‐order interactions of the original Ising model are retained while circumventing the \(\mathcal{O}(n^p)\) bottleneck. However, the clause outputs \(T_k\) are not independent, so the naive way of sampling directly in the \(M\)-dimensional \(T_k\)-space to minimize energy is inefficient—adjusting one output usually forces changes in several others. To address this, the proposed autoencoder‐based model (Fig.~\ref{fig1}.c) projects the high‐dimensional \(T_k\) vector into a lower‐dimensional spin vector of size \(N\), with \(N < M\). In this reduced spin space, each spin is considered to be independent, so sampling an \(N\)-element spin vector immediately yields a consistent clause‐output configuration. These spin variables thus serve as the latent space (or bottleneck) of our autoencoder. 

To efficiently sample the spin space as described, we must compute the energy change based on the clause outputs for each potential spin flip.
If, $s_{i,n}$ denotes the $i$'th spin at time instant $n$, and $T_{k,n}=\prod_i s_{i,n}^{H_{k,i}}$ denotes the $k$'th clause output at time $n$ then flipping the spin $s_{i,n}$ leads to a decrease in energy $E$ if and only if
\begin{equation}\label{eq_intro_del_E}
\Delta E_{i,n}<0
\;\iff\;
\sum_{k=1}^{M}H_{k,i}\,J_{k}\,T_{k,n-1}<0.
\end{equation}

Methods Section~\ref{sec_asynchronous_ising_model} presents a detailed derivation of Eq.~\eqref{eq_intro_del_E}. In the Methods Section~\ref{sec_autoencoder_model}, we show how the Hamiltonian in Eq.~\eqref{eq:General_Ising_min} can be minimized using the autoencoder architecture (shown in Fig.~\ref{fig1}.c). Sampling in the latent space uses the projection of clause outputs to spin space according to Eq.~\eqref{eq_intro_del_E}. Concretely, the outer layers of neurons of the autoencoder correspond to the $M$ clauses and the inner layer (or, latent layer) of neurons correspond to the $N$ spin variables.  
Note that the summation in Eq.~\eqref{eq_intro_del_E} is exactly the weighted integration of clause outputs in the encoder stage (Fig.~\ref{fig1}.c). In particular, the \(i\)-th column of \(\mathbf{H}\) indicates which previous clause outputs \(T_{k,n-1}\) connect to the \(i\)-th latent neuron—each contribution being weighted by its coupling strength \(J_k\). Each latent neuron does more than simply sum its inputs: if its aggregated input exceeds a noisy threshold \(\mu_{i,n}\), it emits a spike \(q_{i,n}\). This thresholded nonlinearity precisely mirrors the acceptance‐rejection step of the Simulated Annealing algorithm.

In the decoder stage, each \(p\)-th order clause output \(T_k\) is reconstructed by combining the spikes from the \(p\) associated latent neurons. At time step \(n\), let \(q_{i,n}\) denote the binary output of the \(i\)-th latent neuron after thresholding by \(\mu_{i,n}\), which is chosen so that only one latent neuron can fire at a time (ensuring asynchronous operation). Similarly, let \(T_{k,n}\) represent the \(k\)-th clause output, as determined by the \(k\)-th decoder neuron. The resulting neuron models are described by:

\begin{equation}
\begin{aligned}
\textbf{Latent Neuron Model:}&\quad
q_{i,n} = 
\begin{cases}
1, & \text{if } \displaystyle \sum_{k=1}^{M} H_{k,i}\,J_{k}\,T_{k,n-1} \;<\; \mu_{i,n},\\[0.6em]
0,       & \text{otherwise},
\end{cases}
\\[1.5em]
\textbf{Decoder Neuron Model:}&\quad
 T_{k,n} = -T_{k,n-1}, 
\quad \text{if } \displaystyle \sum_{i=1}^{N} H_{k,i}\,q_{i,n} \;>\; \theta.
\end{aligned}
\end{equation}
\noindent
The decoder neuron implements a toggle operation which can be mapped onto a standard neuromorphic architecture using a pair of ON-OFF integrate-and-fire neurons~\cite{Chen2025} (see SI Section~\ref{sec_supp_ON_OFF_neuron} for details on this mapping). For the sake of brevity and clarity, the algorithm and implementation in this paper will directly use toggle neurons.
Also, in a practical implementation, the latent neuron outputs \(q_{i,n}\) could be a spike-train or a continuous value rather than the idealized binary (0/1) state. Consequently, applying the threshold \(\theta\) at the decoder neurons ensures that small, unintended variations in the latent neurons do not trigger a response.

Fig.~\ref{fig1}.d shows the architecture of a neuromorphic Ising machine  implementing the autoencoder corresponding to an interconnection matrix $\mathbf{H}$. Here clause outputs are implemented as toggle neurons that flip state whenever their input exceeds a threshold $\theta$. 
Note that in the depicted architecture, only clause \(T_1\) incorporates a third‐order interaction, as highlighted in the figure.

We anneal the noisy thresholds of each latent neuron ($\mu_{i,n}$) over time using a Fowler–Nordheim (FN) annealer (see Methods Section~\ref{sec_methods_FN_annealer}).  As shown in \cite{Chen2025} for a second-order Ising machine, the FN-annealer produces an optimal annealing schedule that guarantees asymptotic convergence to the ground state of the underlying combinatorial optimization problem. The FN dynamics may be realized either by a physical FN‐tunneling device~\cite{Zhou2017FNtimer} or via a digital emulation of the FN‐tunneling dynamical systems.  In this work, we employ the digital emulation to achieve the precision required for simulated annealing in the low‐temperature regime.

Similar to~\cite{Chen2025}, the FN dynamics are combined with independent and identically distributed (i.i.d.) noise variables, $\mathcal{N}_{i,n}^E$ and $\mathcal{N}_{i,n}^B$, to produce the time‐varying threshold $\mu_{i,n}$.  Here, $\mathcal{N}_{i,n}^E$ is drawn from an exponential distribution, while $\mathcal{N}_{i,n}^B$ is sampled from a Bernoulli distribution on $\{0,1\}$.  This combination ensures that every neuron maintains a nonzero probability of firing, satisfying the irreducibility and aperiodicity requirements of simulated annealing.

Since most practical combinatorial‐optimization problems lead to sparse interconnection matrices, our architecture leverages this sparsity via graph‐coloring of the spin variables.  Although computing an optimal coloring is NP-hard, we follow the approach of Aadit et al.~\cite{Aadit2022} by using a fast greedy heuristic during preprocessing. While it does not guarantee optimality, it is efficient for our purposes.  In the colored graph, latent neurons sharing the same color—or that are conditionally independent—are updated simultaneously, while neurons of other colors are held at a very high threshold.  Consequently, the latent neurons require only the annealed exponential noise as their effective threshold.  This graph‐coloring strategy enables massively parallel updates across these neurons at each time step.

In previously reported higher‐order Ising‐machine architectures—where the spins themselves are the dynamic state variables—encoding a \(p\)‐body interaction requires hardware resources that scale as \(\mathcal{O}(n^p)\), since each \(p\)-th order term must be implemented across \(n\) spins. In contrast, our higher‐order Ising machine treats clauses as the dynamic variables—effectively encoding higher‐order interactions—while still sampling the energy landscape efficiently in the spin domain. Consequently, the hardware complexity remains invariant regardless of interaction order. 
Moreover, by leveraging computational sparsity (omitting entries in $\mathbf{H}$ with zeros), we can further reduce the hardware footprint and enable massively parallel updates via graph coloring, which together enhance both solution quality and time-to-solution as demonstrated in our results. Since our architecture is functionally isomorphic to the optimal simulated‐annealing algorithm, it can asymptotically converge to the true Ising ground state facilitated by the optimum annealing using FN-annealer. As we show, for inherently higher‐order combinatorial problems such as MAX‐3SAT, even finite‐duration runs on our neuromorphic machine consistently yield higher‐quality solutions in shorter time-frames than the quadratized equivalent executed on a second‐order Ising machine.

\begin{figure}[h]
\centering
\includegraphics[width=\textwidth]{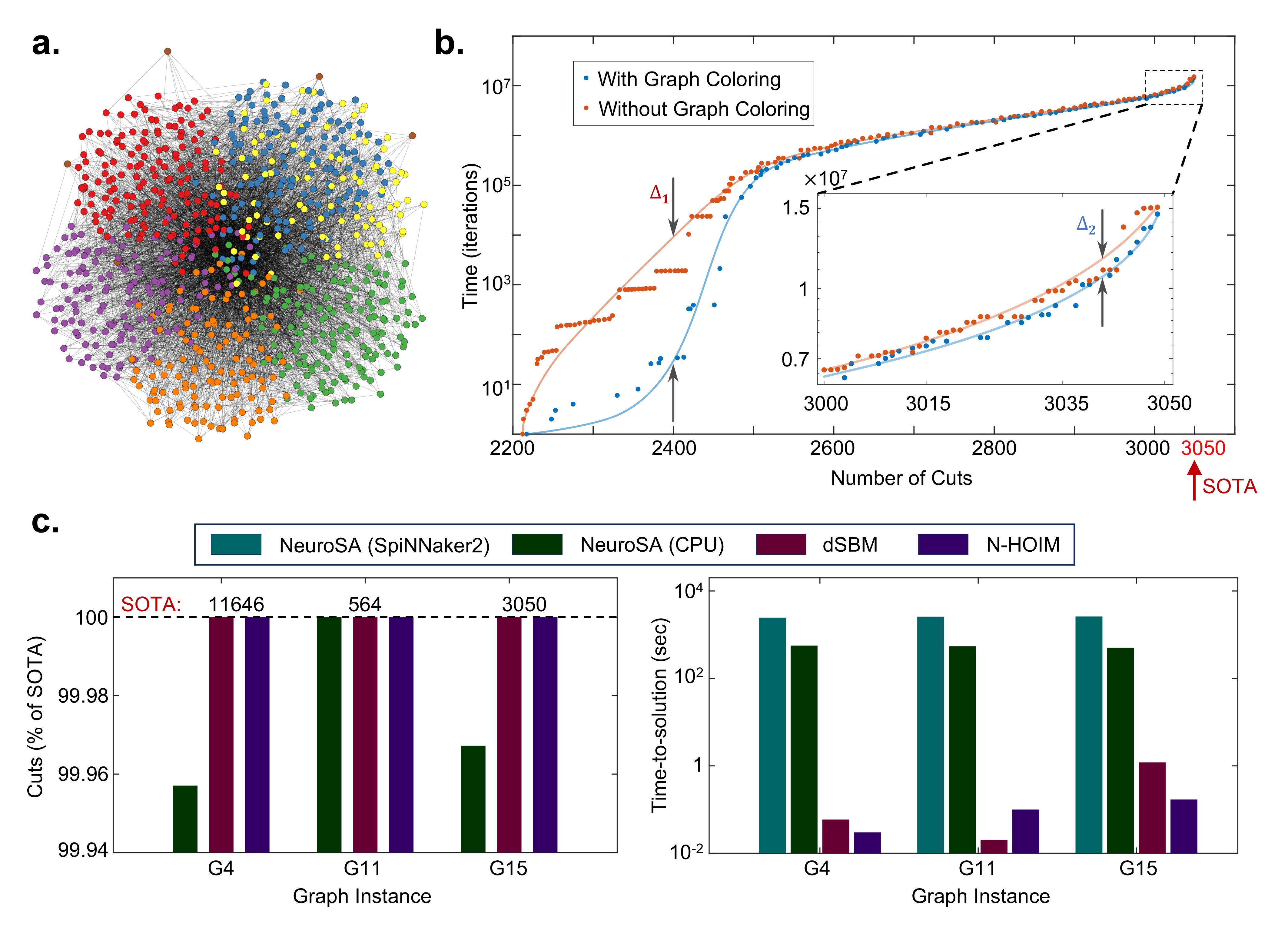}
\caption{\textbf{Results using MAX-CUT graphs from the Gset benchmark~\cite{Gset}:} 
\textbf{(a)} The G15 graph with 800 vertices and 4661 edges utilizes 7 colors to color the graph using graph coloring. 
\textbf{(b)} This plot presents the time taken to achieve a specific number of cuts and the steady improvement in the solution quality towards the 3050 cuts, which is the current state-of-the-art (SOTA) for this graph. Notably, in the initial (high-temperature) regime, the trajectory with graph coloring exhibits bigger jumps in cuts, following a faster path towards the SOTA compared to the trajectory without graph coloring (denoted by $\Delta_1$). The inset illustrates the trajectory near the SOTA, demonstrating that in the low-temperature regime, both trajectories approach the SOTA while maintaining an offset in time of approximately $~5 \times 10^5$ (a.u.) (denoted by $\Delta_2$). In \textbf{(c)}, we compare the best cuts and the time required to achieve those cuts on three graphs (G4, G11, G15) for various Ising machine implementations. We compare our FPGA implementation of the neuromorphic higher-order Ising Machine (N-HOIM) with the previously demonstrated neuromorphic second-order Ising solver, NeuroSA~\cite{Chen2025} (CPU and SpiNNaker2 implementations), and the state-of-the-art Discrete Simulated Bifurcation Machine by Toshiba (dSBM)~\cite{SBM_Toshiba}.}\label{fig2}
\end{figure}

\section{Results}\label{sec2}
{First, we analyze the dynamics of our neuromorphic higher-order Ising machine on a small 3R-3X problem, where the search dynamics is easily visualized~\cite{Krzakala_2010}. The main objective of this experiment is to compare against the second-order Ising machine. Next, we employ a well-established MAX-CUT benchmark to showcase the benefits of graph coloring, and we compare our hardware implementation (on an FPGA) against other reported solutions to demonstrate its performance even on inherently second-order problems. Finally, we assess our higher-order neuromorphic Ising machine on MAX-SAT instances to quantify its edge over a second-order solver when addressing optimization tasks with intrinsic higher-order interactions.}

\subsection{Experiment on small 3R-3X problem}\label{sec_3r3x}
Our neuromorphic higher‐order Ising machine was initially tested on a small 3‐Regular‐3‐XORSAT (3R‐3X) instance comprising 10 variables and 10 clauses. Although this problem can be solved in $\mathcal{O}(n^3)$ time by Gaussian elimination, its regular structure induces an NP‐hard, “egg‐carton”-like energy landscape~\cite{Glassy_without_hard}, making it an ideal problem for probing network‐dynamics trajectories even at small scales. The higher‐order optimization objective polynomial yields the interconnection matrix $\mathbf{H}$, and the corresponding clause weight matrix $\mathbf{J}$ appears in Fig.~\ref{fig1}.e. After applying quadratization, the resulting second‐order Hamiltonian $\mathbf{Q}$ is displayed in Fig.~\ref{fig1}.f. A detailed description of the equation‐planting procedure and the mapping from 3R‐3X to both the higher‐order and second‐order interconnection matrices is provided in SI Section~\ref{sec_supp_3R3X_Ising}. Comparison of Figs.~\ref{fig1}.e and~\ref{fig1}.f readily reveals the increased precision and expanded range of the pairwise coupling coefficients following quadratization.

We then solve this instance using two approaches: directly on the higher‐order interaction matrix $\mathbf{H}$ with the proposed higher‐order Ising machine, and on the quadratized Hamiltonian $\mathbf{Q}$ with NeuroSA~\cite{Chen2025}, a previously reported neuromorphic second‐order Ising machine. Since both implementations are asynchronous event-based, we can visualize their search dynamics by aggregating spike rates (or event-rates). For NeuroSA, we use the total spiking rate of each neuron, while for our architecture we use the aggregated spike rates from each latent neuron. These aggregated rates are computed over a moving time window. We then project the resulting population‐firing dynamics onto a reduced three‐dimensional space spanned by the three dominant principal components, as shown in Fig.~\ref{fig1}.g. This principal‐component‐analysis (PCA)–based approach is standard in spiking‐data analysis~\cite{Population_firing_dynamics}; details are described in Methods Section~\ref{sec_methods_PCA}. To ensure a fair comparison, the annealing schedules and initial spin configurations are identical for both Ising machines, eliminating any artifacts from differing parameters. Due to the unique energy landscape of the 3R‐3X problem, the solvers do not exhibit long dwell times even upon reaching the global minimum. Therefore, we plot the population‐firing trajectories for each Ising machine up to the moment it first attains the ground state and remains there briefly. As illustrated in Fig.~\ref{fig1}.g, the higher‐order trajectory (labeled “HO Trajectory”) rapidly converges to the ground state, whereas the second‐order trajectory (labeled “SO Trajectory”) undergoes an extended random‐walk before finding the global minimum. This behavior directly reflects the larger search space introduced by quadratization in the second‐order solver.

\subsection{Experiments using MAX-CUT}\label{sec_maxcut}
Next, our algorithm is applied to a problem of MAX-CUT, whose ground state is unknown but for which the State-of-the-art (SOTA) is well documented.  We select the G15 graph—an 800-node, binary-weighted planar instance~\cite{Gset}—for which the current SOTA is 3050 cuts~\cite{Goto2019SBM}. The mapping for MAX-CUT problems to the Ising formulation required for our architecture is detailed in SI Section~\ref{sec_supp_MAXCUT_Ising}. Since MAX-CUT is the simplest case of an XORSAT problem (two-variable clauses), it provides a clear means to analyze our algorithm’s behavior on a relatively large graph. To assess the impact of graph coloring, we compare the dynamics of our algorithm on G15 both with and without coloring. We color the graph using the DSATUR, as a heuristic graph-coloring algorithm~\cite{DSATUR}.  As shown in Fig.~\ref{fig2}.a, DSATUR assigns seven distinct colors to G15. These simulations were performed on a CPU-based platform.  We observe that both implementations—colored and uncolored—converge steadily toward the SOTA of 3050 cuts as shown in Fig.~\ref{fig2}.b.

The dynamics naturally split into two regimes—high temperature and low temperature—based on the noisy threshold $\mu(t)$ provided by the FN annealer, which follows an annealing schedule $T_n\sim1/\log(n)$.  In the high-temperature phase (initial iterations), the noise threshold values ($T_n\log(u_n)$, see Methods Section~\ref{sec_autoencoder_model}) are large and negative, so the system evolves primarily along the gradient of the Ising energy.  With graph coloring, multiple independent variables can flip simultaneously, yielding larger gradient steps toward convergence compared to the asynchronous, single-spin flips in the uncolored case.  This effect is evident in the high-temperature trajectories ($\Delta_1$) shown in Fig.~\ref{fig2}.b.

As the temperature decreases, the search becomes confined to a local minimum.  The FN annealing dynamics maintain a finite probability of escape even in the low-temperature regime, so the behavior is dominated by a combination of random walks and sporadic escapes, without a preferred direction.  These occasional escapes allow the algorithm to continue exploring and potentially reach the global optimum.  However, in this regime, graph coloring offers less advantage: the inset of Fig.~\ref{fig2}.b shows that both colored and uncolored runs arrive at the SOTA level while maintaining an offset of approximately $5\times10^5$ a.u.\ ($\Delta_2$), and this offset remains roughly constant throughout the low-temperature region.

Next, we benchmark our neuromorphic higher-order Ising machine implementation on dedicated hardware for solving MAX-CUT problems. We implemented our higher‐order Ising machine with graph coloring on the RFSoC 4x2 FPGA Kit featuring the Zynq UltraScale+ RFSoC XCZU48DR‐2FFVG1517E core.  Design choices and architectural details are provided in Methods Section~\ref{sec_methods_sync_impl}. An important design detail to note here is that the noisy thresholds $\mu(t)$ in the FPGA implementation use 16‐bit signed integers. However, all CPU‐based algorithmic experiments reported in this work (unless otherwise specified) employed 64‐bit floating‐point thresholds.  As shown in \cite{Chen2025}, reducing noise‐threshold precision increases the time required to reach a given solution quality.  For all the experiments in this work, the FPGA runs at a 100 MHz clock frequency.

To evaluate the performance benefits of our implementation even on problems that naturally map to second‐order interactions, we compared the FPGA‐based higher‐order Ising machine with graph coloring against NeuroSA~\cite{Chen2025} (both CPU and SpiNNaker2 implementations) and the state‐of‐the‐art simulated bifurcation machine by Toshiba~\cite{SBM_Toshiba}. We conducted these comparisons on three graph instances from the Gset benchmark~\cite{Gset}. The left barplot in Fig.~\ref{fig2}.c compares the best cuts achieved—expressed as a percentage of the SOTA—using the reported results from the other two implementations. The right barplot in Fig.~\ref{fig2}.c presents the time‐to‐solution (TTS) (see Methods Section~\ref{sec_methods_benchmark_dataset} for the definition of TTS in this context) for each implementation to reach its respective best cut. Detailed results are provided in Table~\ref{tab:gset-cuts} in the SI Section. Our implementation attains substantially lower TTS than previously reported second‐order neuromorphic Ising machines and even outperforms NeuroSA on the G4 and G15 graphs in terms of best-cuts (achieving state‐of‐the‐art for all three instances, unlike NeuroSA). Remarkably, our TTS is comparable to—and, for certain graphs, better than—that of the simulated bifurcation machine~\cite{SBM_Toshiba}. This performance advantage arises from optimized annealing given by FN annealer, sparse problem embedding, efficient clause‐based traversal, and acceleration afforded by graph coloring during the high‐temperature phase. For the detailed methodology of the TTS evaluation, see Methods Section~\ref{sec_methods_benchmark_dataset}.

\begin{figure}[ht!]
\centering
\includegraphics[width=\textwidth]{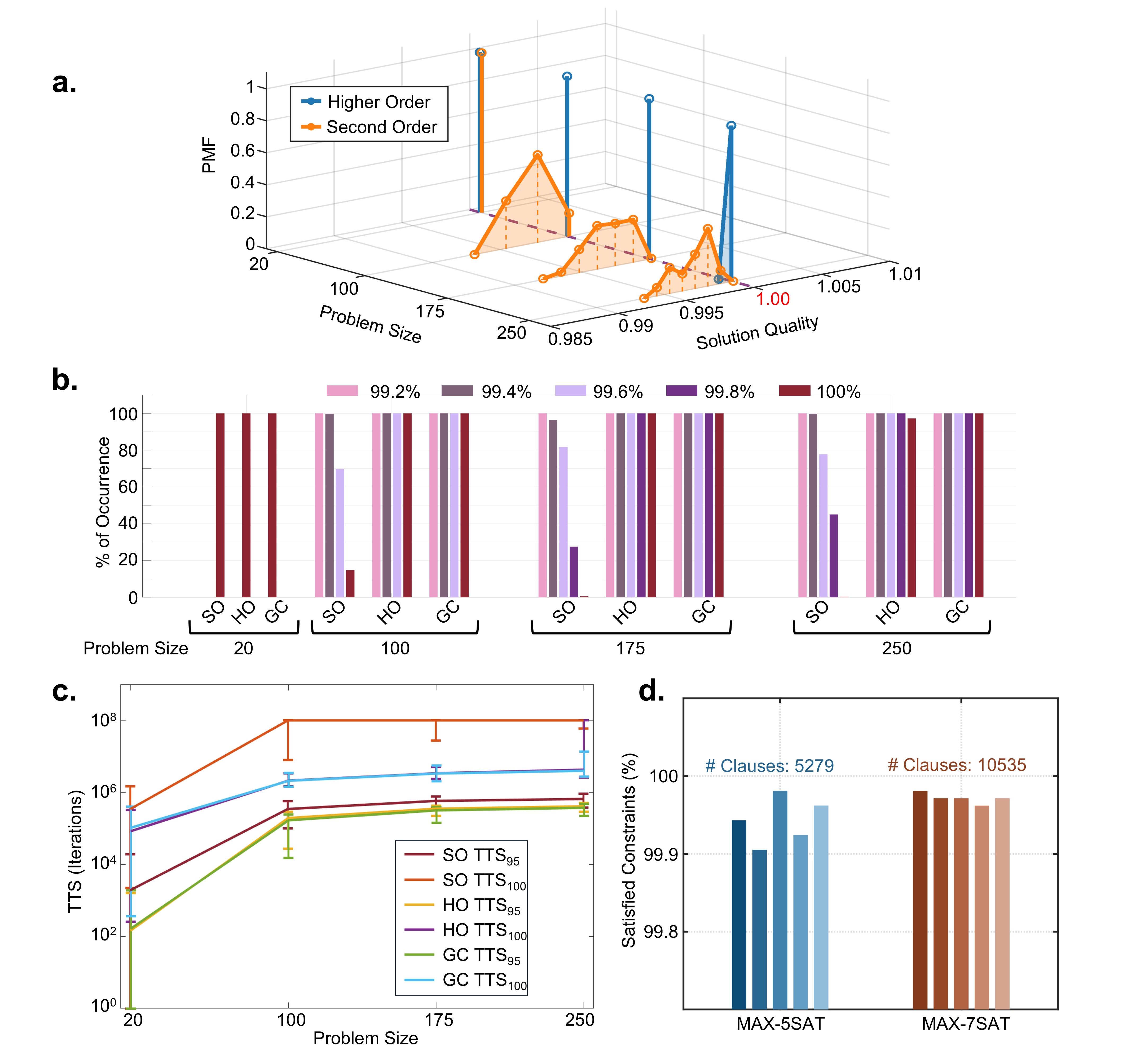}
\caption{\textbf{Results for MAX‐SAT problems:}  The first three panels present results on MAX‐3SAT instances from the UBC SATLIB benchmark~\cite{SATLIB}.  
\textbf{(a)} Probability mass functions (PMFs) of solution qualities for MAX‐3SAT problems of varying sizes (variables: 20, 100, 175, 250), where clauses count are 91, 430, 753, and 1065, respectively.  Solution quality is expressed as the fraction of clauses satisfied.  We observe consistent better-quality solutions for our higher-order implementation (blue) compared to a quadratic-Ising solver — NeuroSA~\cite{Chen2025} (orange).  
\textbf{(b)} Bar plot showing, for each problem size, the percentage of runs that achieve at least a given solution quality (99.2\%, 99.4\%, 99.6\%, 99.8\%, and 100\% of the optimal SAT).  Results are shown for the {\it NeuroSA} (SO), our higher‐order method (HO), and the higher‐order variant with graph coloring (GC). For larger problem sizes, SO’s success rate drops sharply as we intend to get a higher solution quality. In contrast, both HO and GC maintain almost flat occurrence profiles.
\textbf{(c)} Time‐to‐solution (TTS) curves for SO, HO, and GC across problem sizes. Here, $TTS_{95}$ denotes the time required to satisfy 95\% of the total clauses, and $TTS_{100}$ denotes the time to reach full satisfiability. The error bars, representing the full range of observed TTS, indicate that \(TTS_{100}\) for HO is almost an order of magnitude different from SO in those runs where SO achieved full satisfiability. To illustrate the feasibility of extending our architecture to any higher interaction orders, in \textbf{(d)} we present the percentage of satisfied clauses over five independent runs for one MAX-5SAT and one MAX-7SAT instance drawn from the SAT Competition 2018 dataset~\cite{SAT_competition_2018}.
}\label{fig3}
\end{figure}

\subsection{Evaluation on MAX-SAT problems}\label{sec_maxsat}

Next, we evaluated the algorithm on MAX‐3SAT instances to assess its performance on an inherently higher‐order problem compared to second‐order solvers. SI Section~\ref{sec_supp_MAXSAT_Ising} shows in detail how to map such higher-order SAT problems to Ising formulation, which is required for our algorithm. For a fair comparison, here as well we selected NeuroSA~\cite{Chen2025}— being a second‐order Ising solver using the same annealing schedule as ours—and tested it on quadratized MAX‐3SAT benchmarks (see SI Section \ref{sec_supp_quadratization_of_3SAT} for details on quadratization of MAX-3SAT problems).  We used the uniformly generated, satisfiable Random‐3SAT instances from the UBC SATLIB dataset~\cite{SATLIB}.  Fig.~\ref{fig3} presents a comprehensive evaluation of our algorithm across various MAX‐3SAT problem sizes, with all tests conducted on a conventional CPU platform. The FN annealer characteristics is kept identical for every benchmark, demonstrating that our algorithm’s effectiveness is consistent across instance sizes and obviating the need for individualized hyperparameter tuning for problem sizes of similar scales.

Fig.~\ref{fig3}.a shows the distribution of final solutions—normalized by the ground state (unity marker)—for both our higher‐order implementation and NeuroSA (second‐order) across problem sizes of 20, 100, 175, and 250 variables, which correspond to 91, 430, 753, and 1065 clauses, respectively. The probability mass function is empirically obtained using a sample of runs as a point estimate.  While NeuroSA solutions consistently fall within 99 \% of the ground state solution quality, their variance grows with problem size.  In contrast, our higher‐order solver produces a sharply peaked distribution around unity, indicating consistently better solutions on all benchmark sizes.

Next, we compare three variants: higher‐order with graph coloring (GC), higher‐order without graph coloring (HO), and NeuroSA’s second‐order model (SO).  Since the average time to reach a given solution quality depends on the probability of attaining that quality within a fixed time, Fig.~\ref{fig3}.b plots, for each problem size (20, 100, 175, 250), the percentage of runs that achieve at least 99.2 \%, 99.4 \%, 99.6 \%, 99.8 \%, and 100 \% of the clauses. For larger instance sizes, achieving full satisfaction of all clauses is particularly a hard problem, because only very few configurations in the entire space satisfy every clause. Only for size 20 does the second‐order model attain full satisfaction across all runs; for larger sizes, its success rate declines sharply as we intend to get a higher solution quality.  By contrast, both HO and GC maintain nearly flat occurrence profiles. In fact, our higher‐order implementation reliably finds satisfiable solutions for the largest 3SAT instances (250 variables), a feat that previous second‐order Ising machines were unable to achieve~\cite{Aadit2022}, and which aligns with prior results on higher‐order oscillator‐based Ising machines~\cite{Bybee2023}.  We also observe that, for size‐250 problems, the higher‐order solver without graph coloring occasionally falls short of 100\% clause satisfaction—visible as a slight dip in the occurrence profile for size 250—whereas the graph‐colored variant consistently reaches full satisfiability even in those instances, as shown by its flat profile in Fig.~\ref{fig3}.b.  This demonstrates that, on larger problem sizes, graph coloring not only improves time‐to‐solution but also enhances the reliability of finding optimal solutions compared to the non‐colored implementation. 

Fig.~\ref{fig3}.c reports the time to reach 95\% satisfied clauses ($TTS_{95}$) and 100\% satisfied clauses ($TTS_{100}$) for the three implementations (SO, HO, GC) across varying problem sizes.  Each instance was run for up to $10^8$ iterations, and error bars indicate the full range of observed TTS values for each size.  From these error bars, it is clear that, when the second‐order model does achieve 100\% satisfaction, its TTS exceeds that of our higher‐order implementations by more than an order of magnitude. As noted in Fig.~\ref{fig3}.b, the second‐order solver attains full satisfiability less frequently as the problem size increases within the $10^8$‐iteration limit.  Since all solvers share the same annealing schedule, the second‐order model is nonetheless guaranteed to converge asymptotically to the ground state; runs that fail to find 100\% clause satisfaction within the limit would require substantially more iterations to do so.  Accordingly, we assign a TTS of $10^8$ (Maximum iteration) to those unsatisfied runs when computing the median TTS for each size, causing $TTS_{100}$ for the second‐order implementation to approach the iteration cap for sizes beyond 20. We also note that the higher‐order implementation with graph coloring does perform better than the non‐colored version in time-to-solution. However, this advantage is not immediately apparent in the figure, since the difference in time (a.u.) is one order less than the absolute TTS.

For the next sub-figure Fig.~\ref{fig3}.d, we took one problem instance each for MAX-5SAT and MAX-7SAT problem from benchmark dataset \cite{SAT_competition_2018} to show the feasibility of extending our algorithm to any order of interaction. We report the number of clauses satisfied as a percentage of the total number of clauses (for MAX-5SAT: 5279 and, for MAX-7SAT: 10535) for each of these instances for five different runs after running the solver for $10^8$ iterations. We observe that, within the specified simulation time, our algorithm consistently produces solutions satisfying over 99\% of the constraints.

\begin{figure}[ht!]
\centering
\includegraphics[width=1\textwidth]{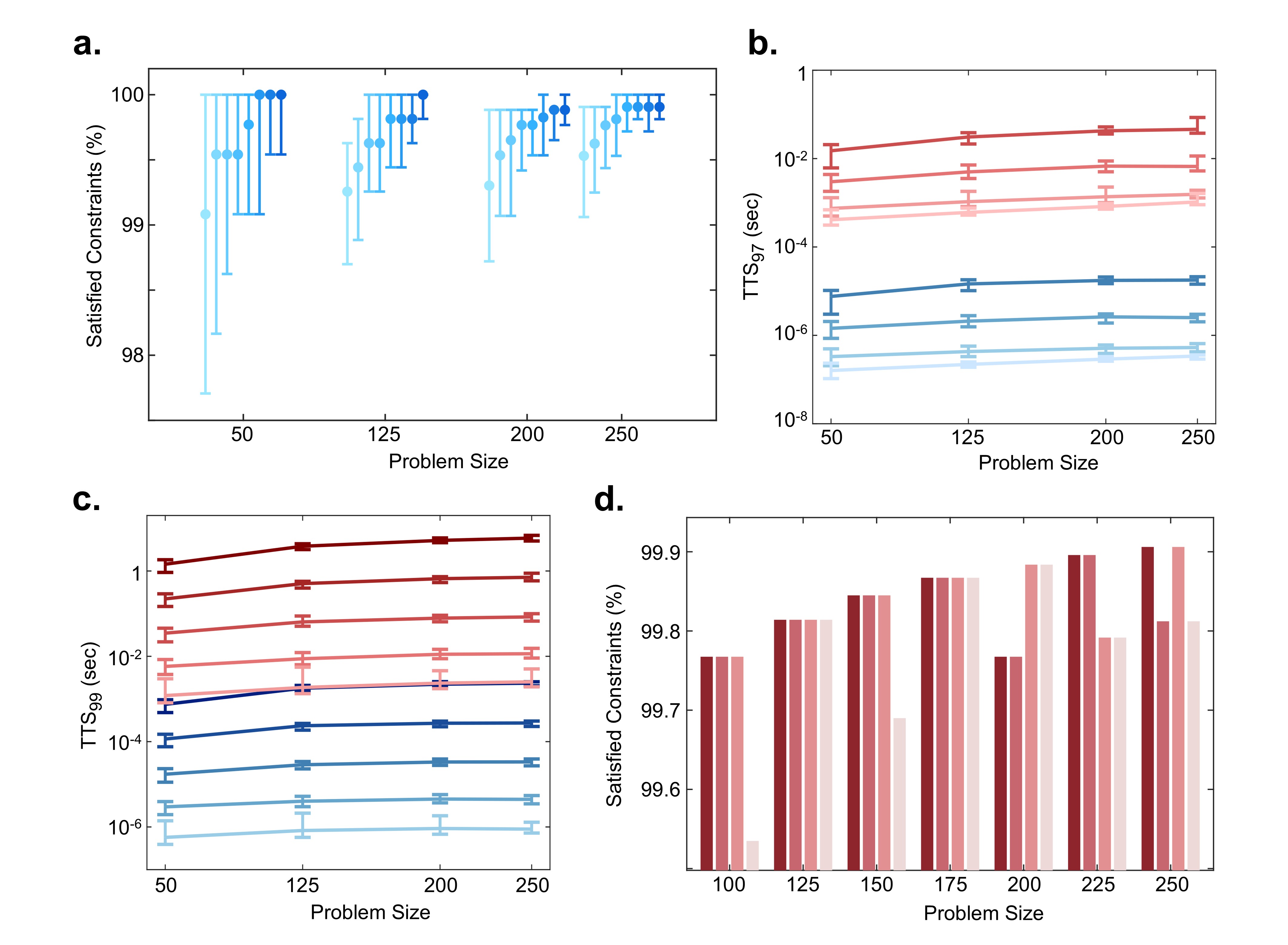}
\caption{\textbf{Results for MAX‐3SAT on FPGA:} The first three panels present FPGA experiments on satisfiable instances drawn from SATLIB~\cite{SATLIB}. Our objective in these experiments was to report the absolute Time-to-Solution (TTS) (in seconds) achieved on the FPGA for a specified solution quality. However, since the algorithm is probabilistic, there exists a trade-off of TTS and solution quality with the annealing schedule. Consequently, in \textbf{(a)} we illustrate how solution quality varies probabilistically with the annealing schedule. This shows the percentage of clauses satisfied after a 320 \(\mu\)s run for MAX‐3SAT instances of varying sizes (variables: 50, 125, 200, 250; clauses: 218, 538, 860, 1065, respectively). For each problem size, eight distinct annealing schedules are tested—depicted in shades of blue from darkest (slowest schedule) to lightest (fastest schedule)—where each darker shade corresponds to a schedule twice as slow as the next lighter one. Each instance undergoes 100 independent trials per annealing schedule. Markers indicate the median satisfaction percentage across trials, and the error bars span the full observed range. We observe that, for every problem size, slowing the annealing schedule shifts the median clause-satisfaction closer to 100\% and narrows the error bars, indicating a higher probability of achieving full satisfiability. However, slower annealing also increases the time required to reach a given solution quality. To illustrate this effect, panels \textbf{(b)} and \textbf{(c)} plot the time (in seconds) needed to attain 97\% (\(TTS_{97}\)) and 99\% (\(TTS_{99}\)) clause satisfaction, respectively, for all problem sizes and annealing schedules—darkest = slowest, lightest = fastest—where each darker schedule is ten times slower than the next lighter one. Red shades denote CPU‐based TTS (in seconds) and blue shades denote FPGA‐based TTS. Finally, since many professional SAT solvers exhibit different behavior on unsatisfiable instances, in \textbf{(d)} we report the best percentage of clauses satisfied—across multiple runs—for four distinct unsatisfiable SATLIB instances at each problem size, using the FPGA implementation under a slower annealing schedule (see Methods Section~\ref{sec_methods_FN_annealer}).
}\label{fig4}
\end{figure}

\subsection{Benchmarking MAX-SAT on FPGA}\label{sec_FPGA_result}

In this section, we benchmark MAX‐3SAT instances on our FPGA implementation of the higher‐order Ising machine with graph coloring. As with the MAX‐CUT experiments on FPGA (Section~\ref{sec_maxcut}), the noise thresholds are quantized to 16-bits and the FPGA operates at a 100 MHz clock. Our goal in these experiments is to report the absolute time‐to‐solution (TTS), in seconds, achieved on the FPGA for a given solution quality. However, since the algorithm is probabilistic, there is an inherent trade‐off of TTS and solution quality with the annealing schedule - which is illustrated in the next three sub-figures.

We start with illustrating how the solution quality varies
probabilistically with the annealing schedule. For Fig.~\ref{fig4}.a, we executed satisfiable SATLIB instances~\cite{SATLIB} of various sizes on the FPGA for 320 $\mu$s each.  Each instance was tested for 100 different trials for each annealing schedule, and we recorded the final percentage of satisfied clauses.  Methods Section~\ref{sec_methods_FN_annealer} explains how varying the sampling period $\Delta$ in the Fowler–Nordheim annealer schedule (Eq.~\eqref{eq_methods_optimal_temp2}) speeds up or slows down annealing.  For each problem size, we used eight sampling periods $\Delta \in \{4\times10^{-3},8\times10^{-3},\dots,512\times10^{-3}\}$, plotted with progressively lighter error‐bar shades. Thus, each schedule is twice as slow as the next lighter shade.  We observe that, for every problem size, slowing the annealing schedule shifts the median clause‐satisfaction closer to 100\%, and also narrows the error bars, indicating a higher probability of achieving full satisfiability.

However, slower annealing increases the time needed to attain a given solution quality.  This trade‐off appears in Fig.~\ref{fig4}.b, which shows the time to reach 97\% satisfaction ($TTS_{97}$), and in Fig.~\ref{fig4}.c, which shows the time to reach 99\% satisfaction ($TTS_{99}$), across various problem sizes.  In these experiments, we compare the FPGA’s TTS (shown in blue shades) against the TTS achieved from a CPU implementation (shown in red shades) that mimics the FPGA’s 16‐bit noise precision to quantify the FPGA speedup.  For Fig.~\ref{fig4}.b, we vary $\Delta \in \{4\times10^{-3},4\times10^{-2},\dots,4\}$, and for Fig.~\ref{fig4}.c, $\Delta \in \{4\times10^{-5},4\times10^{-4},\dots,0.4\}$, again illustrated using darker to lighter shades. For detailed TTS computation see Methods Section~\ref{sec_methods_benchmark_dataset}. We find that median TTS scales nearly linearly with annealing speed.  Notably, for size‐250 at the fastest schedule, median $TTS_{97}$ is 0.34 µs and median $TTS_{99}$ is as low as 0.89 µs.  Overall, the FPGA implementation achieves a 2500–3000 times speedup over the CPU version.

Since professional SAT solvers behave differently on unsatisfiable instances, we next benchmarked unsatisfiable SATLIB instances~\cite{SATLIB} on the FPGA.  We selected four instances for each size from 100 to 250 variables and ran them with a slower annealing schedule (see Methods Section~\ref{sec_methods_FN_annealer}) over extended durations.  Fig.~\ref{fig4}.d shows the best percentage of satisfied clauses achieved across multiple runs for each instance. Detailed results are provided in Table~\ref{tab:supp-satlib-results} in SI Section.

\section{Discussions}\label{sec_discussion}

In this work, we first showed that higher-order Ising machines are more resource efficient both in the number of Ising variables, as well as in the number of connections compared to second-order Ising machines for solving large combinatorial optimization problems. This is because, quadratization of the original optimization objective polynomial introduces extra auxiliary variables, which in turn exponentially increases the search space. Motivated by this, we presented a neuromorphic higher-order Ising machine using an autoencoder architecture. We extended the isomorphic implementation of simulated annealing onto higher-order Ising machine using a Fowler-Nordheim annealer. The functional isomorphism ensures that the Ising machine can theoretically reach the ground state of the problem; however, in practice, this feature also ensures high-quality solutions can be reliably generated across multiple runs.

The autoencoder-based architecture captures the higher-order interactions in the clause outputs and while the clause outputs are the dynamic variables, the resource complexity always scales as $\mathcal{O}(n^2)$ irrespective of the order of interaction. One may note that even this $\mathcal{O}(n^2)$ dependence can be unsustainable at a large scale. So, we use sparse problem embedding, as many combinatorial optimization problems map to polynomials which correspond to a very sparse higher-order interaction graph. For such sparse problem encoding, we just need $(n\times q)$ interconnection for the encoder stage, where $n$ is the number of spin variables and $q$ is the maximum neighborhood for that variable which encodes the sparsity. While the resource usage for the reconstruction at decoder stage will be in the order of $(m \times p)$, where $m$ is the number of clauses and $p$ is the interaction order of the Ising Hamiltonian. We also use this sparsity of the interconnection matrix to have massively parallel event generation across multiple latent neurons using techniques such as approximate graph coloring. We see from Fig.~\ref{fig2} that this graph coloring can help to quickly converge towards the solution in the high-temperature region. The limitation for this kind of technique is when the graph is sufficiently dense. For example, an all-to-all connected network would require $\mathcal{O}(n)$ colors, which would reduce the parallel event generation to a serial one. In those cases to satisfy the ergodicity property of simulated annealing we have to run it following the implementation we presented without coloring (see Algorithm~\ref{alg:pseudocode_without_color} in Supplementary Information) which uses the asynchronous update mechanism given by the noise threshold incorporating the Bernoulli random variable ($\mathcal{N}_{i,n}^B$) as described in Methods Section~\ref{sec_autoencoder_model}. Or, another potential way out is to sparsify the graph using sparsification techniques as a preprocessing step~\cite{Aadit2022}. But, since these techniques introduce auxiliary variables, this would be interesting to see whether parallel updates on sparse higher-order Ising machines with more variables would be advantageous than the one having serial updates with original objective polynomials.

To show the advantage of the higher-order Ising machine compared to the second-order Ising machine, we chose few satisfiable instances of MAX-3SAT problems from well-studied benchmark dataset SATLIB~\cite{SATLIB}. The clause to variable ratio being equal to the phase-transition ratio makes them a hard problem for systematic SAT solver, stochastic local search-based solver as well as Ising-based solver \cite{hard_problem_phase_transition}. We see from the results in Fig.~\ref{fig3}, that the higher-order Ising machine outperforms the second-order Ising machine (running with the same annealing schedule) in both solution quality as well as time-to-solution (TTS). We also see that graph coloring further improves both solution quality as well as the TTS. For the time-to-solution comparison in Fig.~\ref{fig3}.c, we compared the number of iterations taken as an arbitrary measure of time to reach the intended solution for both our higher-order implementation and second-order implementation based on NeuroSA~\cite{Chen2025} (the initial condition and annealing were kept the same). We note that this is rather a pessimistic measure as the difference in absolute time taken can be even more because of our sparse problem embedding in actual implementation. 
 
One may note that Eq.~\eqref{eq_intro_del_E}, came from the most general weighted Ising formula with any order of interactions, having many applications. Since $\mathbf{H}$ is a binary matrix, we can absorb the coupling coefficients or weights $J_k$ into the row of $\mathbf{H}$ as discussed in Methods Section~\ref{sec_autoencoder_model}. So, embedding this weighted Ising formula translates to weighted integration of the clause outputs at the latent neurons. We presented an FPGA-based implementation utilizing the parallel updates utilizing the sparsity. For these implementations, we can binarize the clause outputs as mentioned in SI Section~\ref{sec_supp_algorithmic_implementation}. This enables us to perform these weighted integrations at the inputs of latent neurons without the need for any multiplier. To compute this, we just need an adder tree of depth $log\,(q)$ for each latent neuron (where $q$ is the maximum neighborhood for that spin). 

In Section~\ref{sec_FPGA_result}, along with the results for the satisfiable SAT problem based on this FPGA implementation, we also show the results for unsatisfiable problem instances from SATLIB (see SI Section~\ref{supp_sec_unsatisfiable_result} for the detailed result). Many incomplete algorithms (Stochastic Local Search (SLS)) based Professional SAT solver perform worse on unsatisfiable problem instances compared to satisfiable SAT problems~\cite{local_search_for_unsat}. We note from our results that with an optimal annealing schedule, our higher-order Ising machine can find the ground state (since it reaches SAT which is = $\#$clauses-1) even for the problem instance of the largest size from SATLIB in a short time.

It has been reported that~\cite{Pierog2015pseudoEnergy}, generating high-quality random noise consumes significant energy and many neuromorphic architectures resort to physical noise (noise intrinsic in devices) as an efficient source of randomness. In our previous works~\cite{Zhou2017FNtimer,Mehta2022FNDAM,Mehta2020FNlog} we have reported a silicon-compatible device that is capable of producing $\mathcal{O}(1/\log)$ decay required by the FN annealer. This device integrates the current due to single electrons tunneling through a Fowler-Nordheim tunneling barrier to generate this annealing behavior. Future work will investigate how to leverage these discrete single-electron events to produce exponentially distributed random variable $\mathcal{N}^E_n$. Now, for the temperature annealing part, these physical FN-tunneling devices could be used as an extrinsic or intrinsic device. The sampling and scaling of the FN-decay can be chosen arbitrarily and is only limited by the resolution of the single-electron tunneling process (quantum uncertainty). This can be mitigated by choosing a larger size device or by modulating the voltage of the device. The time scale of the $\mathcal{O}(1/\log)$ decay can therefore be adjusted to fit the scheduling presented in the algorithm. Note that even with the optimal noise schedule, higher precision sampling would be required.  Due to the requirement of higher precision sampling of the FN device, we have resorted to a digital emulation of the FN-tunneling device on the FPGA platform. Note that our previously reported FN-devices~\cite{Zhou2017FNtimer} could have been used for this purpose, but only in conjunction with analog-to-digital converters with more than 16-bit precision. The scope of this work is limited to developing a neuromorphic architecture incorporating efficient embedding of higher-order interaction that can utilize the FN-annealing dynamics. 

The proposed autoencoder architecture can also be naturally mapped onto a dendritic computing model and hardware~\cite{dendrite_in_3D,Boahen2022}. This mapping is detailed in SI Section~\ref{autoencoder_to_dendritic}. In this mapping, the weighted aggregation of clause outputs at each latent neuron (encoder stage) corresponds directly to synaptic integration along individual dendritic branches. Likewise, the decoder stage—which reconstructs clause outputs from the latent code—parallels how the soma collects signals from multiple dendritic branches to generate each clause output. When the total synaptic input on a dendritic branch exceeds a noisy threshold, the local membrane potential undergoes a nonlinear amplification instead of remaining a purely linear sum of its inputs. Similar to our architecture in Fig.~\ref{fig1}.d, the soma integrates these nonlinear dendritic events and compares the result against a somatic threshold \(\theta\). Whenever the summed dendritic contributions cross \(\theta\), we can likewise interpret this as a discrete spike at the dendritic branches—precisely mirroring the gating mechanism in our autoencoder. A subtlety of such dendritic architectures is that the order of synaptic arrivals can affect integration (due to temporal dynamics and location‐dependent attenuation). However, for our autoencoder‐to‐dendrite mapping, we simplify by assuming that only the total synaptic charge within a fixed integration window matters. In other words, regardless of individual arrival times, if the cumulative input from all relevant clauses in that time window exceeds the branch threshold, the dendrite emits a spike. This implies a need for coarse synchrony—each dendritic branch must maintain an integration window so that inputs arriving in any order are summed appropriately. Beyond this requirement for synchronized integration windows, all other operations in our autoencoder correspond directly to dendritic processing stages.

Building on this dendritic architecture, implementing such architecture in hardware also can help us circumvent the limits of planar device scaling. As planar device miniaturization nears its fundamental limits, extending functionality into the third dimension has become an attractive path for boosting device density and expanding computational throughput. Notably, recent demonstrations of FeFET‐based nanodendritic structures~\cite{dendrite_in_3D} showcase how vertically stacked, dendrite‐inspired pathways can deliver thermally robust, massively parallel processing within a neuromorphic 3D architecture. We envision that adopting these nanodendritic structures—whose design faithfully emulates the arborized signal integration and local nonlinear processing of biological dendrites—will empower such mapping to dendritic‐computation‐based neuromorphic higher-order Ising machine to overcome the scaling ceilings imposed by purely two‐dimensional devices.

\section{Methods}\label{sec3_methods}

\subsection{Methodology for resource comparison} \label{sec_methods_resource}
For the resource comparison in Fig.~\ref{fig1}.b, we used ten instances each of MAX‐SAT of orders 3, 5, 7, 9, and 11. For MAX‐3SAT, we selected the first ten instances of size 250 from SATLIB~\cite{SATLIB}. We chose this largest problem size to generate the plot because, as discussed in SI Section~\ref{sec_supp_resource_vs_size}, the ratio of the interconnection matrix size approaches a constant value as the problem size increases. 

For MAX‐5SAT and MAX‐7SAT, we used ten instances from the uniform random benchmarks of the 2018 SAT Competition~\cite{SAT_competition_2018} with the following specifications: for MAX‐5SAT, 250 variables and 5279 clauses; for MAX‐7SAT, 120 variables and 10535 clauses. 

For MAX‐SAT of orders 9 and 11, we generated datasets by uniformly sampling \(k\)‐SAT instances at the phase‐transition = clause‐to‐variable ratio, making these instances hard~\cite{SAT_competition_2018}. The asymptotic phase‐transition ratio (satisfiability threshold) for random \(k\)‐SAT is given by
\[
r_{k\text{-SAT}} \;=\; 2^{k}\ln 2 \;-\; \frac{1}{2}\bigl(1 + \ln 2\bigr) \;+\; \varepsilon_{k}, 
\qquad \lim_{k \to \infty} \varepsilon_{k} = 0,
\]
as shown in~\cite{Phase_transition_kSAT}. Accordingly, we chose the following specifications: for MAX‐9SAT, 75  variables and 26553 clauses; for MAX‐11SAT, 50 variables and 70936 clauses.

Forming the higher‐order interconnection matrix sizes follows SI Section~\ref{sec_supp_MAXSAT_Ising}. Post‐quadratization costs were estimated using Chancellor’s transformation~\cite{Chancellor2016}, which introduces \(k\) auxiliary spins per clause when embedding MAX‐\(k\)‐SAT into a second‐order graph (for MAX‐3SAT, one auxiliary spin per clause; see SI Section~\ref{sec_supp_quadratization_of_3SAT}).

\subsection{Asynchronous Higher Order Ising Machine Model}\label{sec_asynchronous_ising_model}

Let $\mathbf{s}=(s_1,\dots,s_N)$ be Ising spins with $s_i\in\{-1,+1\}$.  We start from the most general higher-order $N$-spin Hamiltonian which can include any order of interactions from all possible 1-body up to N-body coupling:
\begin{equation}\label{eq:method_re_index}
E(\mathbf{s})
=-\sum_{k=1}^M J_k\prod_{i=1}^N s_i^{H_{k,i}}.
\end{equation}

\noindent
Here, $M$ is the total number of \emph{interaction terms}, i.e., $M = \sum_{p=1}^N \binom{N}{p} = 2^N - 1$.
\noindent
Each \emph{interaction term} is indexed by a single index $k=1,\dots,M$, with associated coupling $J_k$ and spin-tuples given by the binary matrix $\mathbf{H}\in\{0,1\}^{M\times N}$:
\[
H_{k,i} =
\begin{cases}
1, & \text{if } i\text{'th spin is present in } k\text{'th interaction term},\\
0, & \text{otherwise}.
\end{cases}
\]
Here, number of non-zero element (for instance $p$) in $k$'th row of $\mathbf{H}$ denotes the interaction order for that row (fields $p=1$, pairwise $p=2$, etc.).
Equivalently, we define clause output as \(T_k \;=\;\prod_{i=1}^N s_i^{\,H_{k,i}}\), 
so that
\begin{equation}\label{eq:E-T-def}
E(\mathbf{s}) \;=\;-\,\sum_{k=1}^M \;J_k\,T_k.
\end{equation}
\noindent
We now flip spin $s_i\to s_i'=-s_i$, holding all other $s_j$ ($j\neq i$) fixed. $s_i$ is involved in only those interaction terms $k$ for which $H_{k,i}=1$.  We define the \emph{neighborhood}
  \[
    \mathcal{N}(i) \;=\;\{\,k : H_{k,i}=1\}.
  \]
Let \(C_k \subseteq \{1,\dots,N\}\) denote the set of spins associated with clause \(k\). Then, the updated clause outputs \(T_k'\) can be written as
  \begin{align}\label{eq_method_odd_flip}
    T_k' =
    \begin{cases}
      \;(-s_i)\;\prod_{j\in C_k\setminus\{i\}}s_j
    \;=\;-\,T_k, & k\in\mathcal{N}(i),\\
      \phantom{-}T_k,       & k\notin\mathcal{N}(i).
    \end{cases}
  \end{align}
\noindent

\noindent
So, we can write the change in energy resulting from flipping spin $s_i$, denoted as $\Delta E_i$, as follows:
\begin{align}\label{eq:deltaE}
\Delta E_i = 2\sum_{k\in\mathcal{N}(i)}J_k\,T_k
= 2\sum_{k:\,H_{k,i}=1}J_k\;\prod_{j=1}^N s_j^{H_{k,j}}.
\end{align}

\noindent
We can rewrite the Equation \eqref{eq:deltaE} as,
\begin{align}
\Delta E_i
= 2 \sum_{k=1}^{M} H_{k,i}\,J_k\,T_k
= 2 \sum_{k=1}^{M} H_{k,i}\;J_k\;\prod_{j=1}^{N} s_j^{\,H_{k,j}}.
\end{align}

Now, for an underlying optimization problem, the problem of finding the ground state of the higher‐order Ising Hamiltonian \(E(\mathbf{s})\) can be stated as finding $\mathbf{s}^*$ such that,
\begin{equation}\label{eq_methods_ising_energy}
\mathbf{s}^* \;=\;\arg\min_{\mathbf{s}\in\{\pm1\}^N} E(\mathbf{s}).
\end{equation}
To go towards the minima, for each time step \(n\), we need to solve the subproblem of determining the subset of spins whose flip yields a new configuration \(\mathbf{s}'\) such that \(\Delta E_{n} < 0\). Finding this subset in itself is a combinatorial problem, and adopting asynchronous dynamics can simplify this problem \cite{Chen2025}. So, under asynchronous spiking dynamics—where at most one neuron spikes at each time step—the problem reduces to: $\forall n,\text{find a spin }s_{i,n}\text{ such that }\Delta E_{i,n}<0.$
This holds if and only if
\begin{align}\label{eq_methods_del_E_final_eq}
\sum_{k=1}^{M} H_{k,i}\,J_k\,T_{k,n-1} <0.
\end{align}

\subsection{Derivation of the Autoencoder Based Model }\label{sec_autoencoder_model}
As discussed earlier, our autoencoder architecture (Fig.~\ref{fig1}.c) projects the $M$-dimensional clause output vector $T_k$ into an $N$-dimensional spin space ($N<M$) for efficient sampling in the latent space. In the next section, we integrate this Metropolis (Boltzmann) sampling—via the Simulated Annealing algorithm—into this autoencoder framework.

As introduced in~\cite{Kirkpatrick1983}, the simulated annealing algorithm addresses Eq.~\eqref{eq_methods_ising_energy} by probabilistically accepting or rejecting the flip of spin \(s_{i,n}\) according to  
\begin{equation}
    \text{Accept flip at } s_{i,n}\text{:}
    \quad\text{if}\quad
    B\exp\left(\frac{-\Delta E_{i,n}}{\tau_n}\right)
    >u_n,
    \label{eq_methods_SAdynamics}
\end{equation}
where \(u_n\sim\mathrm{Uniform}[0,1]\) and \(B>0\) is a hyper-parameter, while \(\tau_n>0\) signifies the temperature at step \(n\).  Equivalently, one may write  
\begin{equation}
    \text{Accept flip at } s_{i,n}\text{:}
    \quad\text{if}\quad
    \Delta E_{i,n}
    < -\tau_n\log\left(\tfrac{u_n}{B} + \epsilon\right).
    \label{eq_methods_SA_dH}
\end{equation}
\noindent
The small constant \(\epsilon>0\) is introduced for numerical stability when \(u_n/B\) approaches zero; its value is set by the resolution of the hardware and thus treated as a hyperparameter. Now, substituting $\Delta E_{i,n}$ using
Eq.~\eqref{eq_methods_del_E_final_eq} gives,  
\begin{equation}
    \sum_{k=1}^{M} H_{k,i}\,J_k\,T_{k,n-1}
    < -\,\tau_n\,\mathcal{N}_n^E,
    \label{eq_methods_SA_ds}
\end{equation}
where \(\mathcal{N}_n^E = \log\!\bigl(\tfrac{u_n}{B} + \epsilon\bigr)\) follows an exponential distribution.

Noting that $\mathbf{H}$ is a binary matrix, we absorb the coupling coefficients $J_k$ into its rows by defining
$\widetilde{\mathbf{H}} \;=\;\mathrm{diag}(\mathbf{J})\,\mathbf{H}\,$. As shown in the encoder stage of our autoencoder model (Fig.~\ref{fig1}.c), the matrix \(\widetilde{\mathbf{H}}^{T}\) maps the clause outputs \(\{T_{1,n-1}, \dots, T_{M,n-1}\}\) to the “change in energy” in the spin‐dimensional space, i.e., \(\{\Delta E_{1,n-1}, \dots, \Delta E_{N,n-1}\}\). Equivalently, in both Fig.~\ref{fig1}.c and its equivalent architecture in Fig.~\ref{fig1}.d, the input to each neuron in the latent layer is given by the weighted sum of the preceding clause outputs \(T_{k,n-1}\). Concretely, for the \(i\)-th neuron at time step \(n\), the total input is $-\sum_{k=1}^{M} \widetilde{H}_{k,i}\,T_{k,n-1}$.
Whenever this integrated input exceeds the threshold $\tau_{n}\,\mathcal{N}_{n}^{E}$,
the neuron applies its nonlinearity or equivalently emits a spike \(q_{i,n}\), which corresponds to the “code” shown in Fig.~\ref{fig1}.c. 
This is shown as follows,

\begin{equation}
    q_{i,n}=
    \left\{ 
      \begin{array}{ c l }
        1, & \quad \textrm{if } -\sum_{k=1}^{M} \widetilde{H}_{k,i}\,\,T_{k,n-1}>\tau_n\mathcal{N}_n^E,\\
         0, & \quad \textrm{otherwise}.
      \end{array}
    \right.\label{eq_methods_discrete_dendritic_event}
\end{equation}

To ensure that each neuron in the latent layer is sampled uniformly—thereby preserving the ergodicity of the annealing process—we assign to each neuron \(i\) an independent Bernoulli random variable:
\begin{equation}
    \mathcal{N}_{i}^{B} =
    \begin{cases}
        1, & \text{with probability }1-\eta,\\
        0, & \text{with probability }\eta,
    \end{cases}
    \label{eq_methods_bernoulli_rv}
\end{equation}
which allows us to reformulate Eq.~\eqref{eq_methods_discrete_dendritic_event} as
\begin{equation}
    q_{i,n} =
    \begin{cases}
        1, & \text{if } -\sum_{k=1}^{M} \widetilde{H}_{k,i}\,T_{k,n-1} \;>\; \mu_{i,n},\\
        0, & \text{otherwise},
    \end{cases}
    \label{eq_methods_discrete_event_bernoulli_rv}
\end{equation}
where the effective threshold is $\mu_{i,n}=\tau_n\,\mathcal{N}_n^E +A\,\mathcal{N}_{i,n}^B$.
Here, the constant \(A\gg\bigl|\tau_n\log\bigl(\tfrac{u_n}{B}+\epsilon\bigr)\bigr|\) approximately ensures that spike events remain asynchronous across different neurons in the latent layer.

Now, in the decoder stage of our autoencoder model, the \(N\)-dimensional code is mapped back to the \(M\)-dimensional decoder neurons via the matrix \(\mathbf{H}\), as shown in Fig.~\ref{fig1}.c. Concretely, if \(\mathbf{q}_{n}\in
\{0,1\}^{N}\) denotes the spike‐based code at time \(n\), then the input to the \(k\)-th decoder neuron is $\sum_{i=1}^{N} H_{k,i}\,q_{i,n}$. We note that, following Eq.~\eqref{eq_method_odd_flip}, whenever an odd number of spins in clause \(k\) flip at time \(n\), the corresponding clause output \(T_{k}\) flips its sign. Thus, the decoder effectively reconstructs each \(T_{k}\) by checking the parity of the contributing spin flips.
  We therefore define the parity indicator
\begin{equation}
    \sigma_{k,n}
\;=\;\sum_{i=1}^N H_{k,i}\,q_{i,n}
\;\bmod 2. \label{eqn_methods_sigma_cal}
\end{equation}
When asynchronous operation is ensured by careful noise statistics, exactly one neuron spikes at each time step, so the set of \(\{q_{i,n}\}\) contains a single 1 and the rest zeros.  Moreover, for our graph‐coloring scheme, no two spins in the same clause can spike simultaneously.  In both scenarios the parity reduces to the simple sum
\[
\sigma_{k,n}=\sum_{i=1}^N H_{k,i}\,q_{i,n}.
\]
In a practical analog neuromorphic implementation, the latent neuron outputs \(q_{i,n}\) are not strictly binary (0/1) events. Therefore, we only pass \(\sigma_{k,n}\) to each decoder neuron if it exceeds a threshold \(\theta\), as illustrated in Fig.~\ref{fig1}.d. This gating mechanism prevents the decoder neurons from responding to spurious fluctuations in the latent layer. Once the values \(\sigma_{k,n}\) have been obtained and surpass \(\theta\), the update rule for the clause output is
\begin{equation}
    T_{k,n} \;=\;
    \begin{cases}
        -\,T_{k,n-1}, & \sigma_{k,n}\neq 0,\\
        \;\;T_{k,n-1}, & \sigma_{k,n}=0.
    \end{cases}
    \label{eq_methods_firing_criteria}
\end{equation}
Fig.~\ref{fig1}.d illustrates this mechanism: each clause \(T_k\) is represented by a toggle neuron that flips its state according to the accumulated parity \(\sigma_k\) of spike events from its associated latent neurons. These toggle neurons can be effectively implemented using ON-OFF integrate-and-fire neurons as discussed in the SI Section ~\ref{sec_supp_ON_OFF_neuron}. Note that, in the graph-coloring-based implementation, since we update only the conditionally independent latent neurons at any instance, we can use only $\tau_n\,\mathcal{N}_n^E$ as the noisy threshold $\mu_{i,n}$.

\subsection{FN Annealer based dynamical systems}\label{sec_methods_FN_annealer}

To generate the noisy thresholds for the latent layer neurons, we employ the dynamical system model of a Fowler-Nordheim (FN) Annealer as detailed in \cite{Chen2025}. This FN Annealer provides the required annealed random threshold in the form $\mu_{i,n} = \tau_n \mathcal{N}_n^E + A \mathcal{N}_{i,n}^B$, as needed for our autoencoder-based model discussed in Section~\ref{sec_autoencoder_model}.

In~\cite{Chen2025} it was shown that this dynamical system model produces an optimal temperature annealing schedule. This schedule ensures asymptotic convergence to the ground state of the underlying combinatorial optimization problem, which is defined as:
\begin{equation}
    \tau(t) = \frac{\tau_0}{\log\left(1 + \frac{t}{C}\right)}.
    \label{eq_methods_optimal_temp2}
\end{equation}

For experiments carried out on both the CPU and the FPGA, we discretize the continuous cooling schedule \( \tau(t) \) by sampling at \( t_n = 1 + n\,\Delta \), and we vary the sampling period \( \Delta \), which directly controls the speed of the annealing process. For all experiments, the parameter \( C \) was set to \( 8 \times 10^4 \). In the experiments illustrated in Fig.~\ref{fig3} and for the experiment in Fig.~\ref{fig4}.d, \( \tau_0 \) was set to \( 0.15625 \), and the mean value of the random variable \( \mathcal{N}_n^E \) was set to \( -0.083703 \). Also, for all the experiments in Fig.~\ref{fig3}, the sampling period $\Delta$ was chosen to be $2\times10^{-3}$. And, since for the unsatisfiable benchmarking experiment in Fig.~\ref{fig4}.d our goal was to search for the ground state for a particular instance of a problem across multiple runs in FPGA, we ran that experiment with a $10\times$ slower annealing schedule with sampling period $\Delta$ of $2\times10^{-4}$.  In all other experiments depicted in Fig.~\ref{fig4}, \( \tau_0 \) was set to \( 5 \times 10^{-4} \), and the mean value of the random variable \( \mathcal{N}_n^E \) was set to \( -1 \). Also the sampling period $\Delta$ was varied in a range for these experiments as mentioned in the Result Section~\ref{sec_FPGA_result}.

\subsection{Generation of the population firing dynamics using PCA}\label{sec_methods_PCA}

To illustrate and visualize the network dynamics for the 3R-3X problem (Section~\ref{sec_3r3x}), we applied principal component analysis (PCA) to perform dimensionality reduction on the population dynamics, following a procedure similar to that in~\cite{Population_firing_dynamics}. The population spiking activity, represented by aggregate spiking rates, directly reflects the search behavior based on latent layer neuron events.

Spike-iteration recordings were collected for both higher-order (HO) and second-order (SO) network architectures under the conditions described in Section~\ref{sec_3r3x}. Each neuron’s (only latent neuron's for HO) list of spike timestamps was converted into a continuous firing-rate representation using a sliding-window estimator: instantaneous firing rates were obtained by counting spikes within overlapping, fixed-width windows that traverse the iteration axis. This process produced time-indexed rate vectors for all neurons. The resulting high-dimensional firing-rate matrices were then subjected to PCA to extract the principal three-dimensional trajectory that captures the majority of the population variance.

Note that the SO solver’s spiking activity spanned 20 neurons, whereas the HO Ising machine’s activity involved 10 latent neurons. Finally, the real-valued vector sequence was projected onto the three principal components, yielding the 3D trajectories shown in Fig.~\ref{fig1}.g. To suppress high-frequency fluctuations, each of the three principal component coordinates was smoothed with a short moving-average filter. This PCA based projection approach provides a concise, low-dimensional representation of how collective spiking activity evolves over iterations for each network architecture.
 
\subsection{Acceleration on Synchronous Systems}\label{sec_methods_sync_impl}
While a fully asynchronous architecture would be ideal for our neuromorphic higher-order Ising machine, most existing accelerators are clocked—either fully synchronous with address-event-routing packet switching, or globally asynchronous but locally clocked via interrupt-driven units. In these clocked systems, we exploit the mutual independence and i.i.d. properties of the random variables \(\mathcal{N}^E\) and \(\mathcal{N}^B\). Accordingly, in both variants of our higher-order Ising machine (with and without graph coloring) we use only the annealed exponential random variable \(\mathcal{N}^E_n\) as the threshold \(\mu_n\) for each neuron in latent layer. At each simulation step, multiple latent neurons may generate spike; those that do so synchronously are termed “active latent-neurons.” 

In the colored version, conditionally independent active latent-neurons of the same color are processed in parallel, enabling a massively parallel update. In the uncolored version, to preserve ergodicity we employ a global arbiter: first, all latent neurons undergo parallel thresholding to identify active ones; then the arbiter selects one active latent-neuron uniformly at random and discards the others. This “rejection-free sampling” boosts acceptance probability compared to the original simulated-annealing framework and is used in solvers such as DAU~\cite{DAU}. The full algorithmic implementation for both variants is given in SI Section~\ref{sec_supp_algorithmic_implementation}.

For the results in Fig.~\ref{fig2}.c and Fig.~\ref{fig4}, we implemented the graph-colored higher-order Ising machine on an FPGA (see SI Section~\ref{supp_sec_FPGA} for details). The on-board Processing System (PS), controlled via PYNQ’s Python interface, configures, drives, and communicates with the Programmable Logic (PL) core solver. The PS master streams configuration bits and generated 16-bit random noise samples to the PL, emulating Fowler–Nordheim annealer behavior. Within the PL fabric, an AMD LogiCORE AXI DMA IP core manages bidirectional data transfer. Since the FPGA logic consumes one noise sample per clock cycle, the total clock cycle count in the core solver equals the number of samples dispatched by the PS. As each sample is sent, the Python controller increments a global counter.

In Fig.~\ref{fig2}.c and Fig.~\ref{fig4}.a, we run the FPGA for a fixed number of samples determined by this counter and read out the total satisfied clause count (SAT count) for every sample. For Fig.~\ref{fig4}.(b–c), we preload a target SAT count in the Python master at configuration time; on each cycle the PS master monitors the SAT count sent by the FPGA slave over DMA and if it matches with the target SAT count, records the counter value, yielding the time-to-solution (TTS). This measurement excludes communication overhead (e.g., DMA interrupt latency, software noise generation), which is included in the FPGA initialization phase.  
All the experiments on FPGA were run at a 10 ns clock period to meet timing constraints.

For comparison in Fig.~\ref{fig4}.(b–c), we implemented the same logic with same noise threshold precision in Python (VS Code v1.100.2, Python 3.11.9) on an Intel Ultra 9 Processor 185H (up to 5.10 GHz). TTS was measured using the built-in function \texttt{time.perf\_counter\_ns()} to obtain nanosecond precision for the time to reach the intended solution.

\subsection{Benchmark problems and simulation methodology}\label{sec_methods_benchmark_dataset}
For the MAX-CUT results in Section~\ref{sec_maxcut}, we employed the G4, G11, and G15 graphs from the G-set benchmarks~\cite{Gset}. The embedding procedure into our higher‐order Ising interconnection matrix is detailed in SI Section~\ref{sec_supp_MAXCUT_Ising}. In Fig.~\ref{fig2}.c, to ensure a fair comparison of time‐to‐solution (TTS), we adopted the same definition used in the dSBM work~\cite{SBM_Toshiba}.  Specifically, TTS is defined as the computation time required to obtain an optimal (or best‐known) solution with 99\% probability, so that $\mathrm{TTS} \;=\; T_{\mathrm{comp}}\;\frac{\log\bigl(1 - 0.99\bigr)}{\log\bigl(1 - P_{S}\bigr)},$
where \(T_{\mathrm{comp}}\) is the computation time per trial and \(P_{S}\) is the success probability of finding the optimal solution.  In our experiments, \(P_{S}\) was estimated from 1000 independent trials, and we reported the median TTS.  The details of the results of this TTS measurement are provided in SI Section~\ref{supp_sec_compare_Gset}. Note that if \(P_{S} > 0.99\), TTS is simply defined as \(T_{\mathrm{comp}}\).

However, applying the formula above alone does not account for the finding in~\cite{Chen2025} that obtaining even small improvements in solution quality can demand exponentially longer computation times. In SI Section~\ref{supp_sec_compare_Gset}, we demonstrate this by showing that the success rate for G15 remains unchanged despite extended runtimes. Consequently, for all TTS‐related experiments—except the MAX‐CUT benchmarks—we adopt the following approach: any run that fails to reach the target solution quality is assigned a large TTS value, and the median TTS is then calculated across all runs, with error bars included to reflect the observed variance.

All MAX-3SAT instances were obtained from SATLIB~\cite{SATLIB}. For Fig.~\ref{fig3}\,(a–c), we selected the first four satisfiable instances at variable counts of 20, 100, 175, and 250. These were mapped to our higher‐order Ising machine (see SI Section~\ref{sec_supp_MAXSAT_Ising}) both with graph coloring (GC) and without (HO), and—for comparison—quadratized (SO) following the gadget in SI Section~\ref{sec_supp_quadratization_of_3SAT} and solved on the second‐order Ising machine of Chen \emph{et al.}~\cite{Chen2025}. Each size–instance pairing was simulated over 100 trials with random initial spin assignments and noisy thresholds. For Fig.~\ref{fig3}.c, the error bars denote the full range of TTS values recorded over all trials.

In Fig.~\ref{fig4}\,(a–c), the first satisfiable MAX-3SAT instance at sizes 50, 125, 200, and 250 was run for 100 CPU and FPGA trials. In Fig.~\ref{fig4}.a, the error bars represent the entire range of satisfied-clause counts across every trial. Whereas, in Fig.~\ref{fig4}(b--c) we focus on the median TTS trend with different annealing, so the error bars extend from the 6\% to the 94\% percentile—thereby omitting extreme outliers. For Fig.~\ref{fig4}.d, the first four unsatisfiable MAX-3SAT instances for sizes \{100-250\} from SATLIB were each subjected to 10 FPGA trials, and the best-SAT assignment was recorded.

To demonstrate scalability to higher‐order clauses, we examined one MAX-5SAT and one MAX-7SAT instance from the 2018 SAT Competition~\cite{SAT_competition_2018}, with mappings described in SI Section~\ref{sec_supp_MAXSAT_Ising}.

\subsection{Data Availability}\label{sec_data}

The performance metrics for the CPU and FPGA implementation solving the benchmark problems, and other data supporting the figures within this paper,
are available from the corresponding author on a reasonable request.

\subsection{Code Availability}\label{sec_code}

The specific MATLAB, Python, and Verilog codes used in simulation/emulation studies on the CPU and FPGA platforms are available from the 
corresponding author upon reasonable request.

\section{Acknowledgements}\label{sec_acknowledgements}
This work is supported in part by research grants from the US National Science Foundation: ECCS:2332166 and FET:2208770. F.A., S.C. and A.N. would like to acknowledge the McDonnell International Scholars Academy seed grant and a Memorandum of Understanding between WashU and IISc. The initial work on XOR-SAT solvers was performed when A.N. was a visiting scholar at WashU under the Fulbright-Nehru Doctoral Fellowship program. J.S. and J.K. also acknowledge financial support from the Horizon Europe grant (Agreement No. 101147319, EBRAINS 2.0).

\section{Author Contributions}\label{sec_contributions}
F.A., S.M., A.N., J.K., J.S., C.S.T. and S.C. 
participated in a workgroup at the Bangalore Neuromorphic Engineering Workshop in 2025, and the outcomes from the workgroup have served as the motivation for this work. S.C., Z.C., M.S., C.S.T., J.E., and A.G.A. also participated in a workgroup at the Telluride Neuromorphic Engineering Workshop in 2024, which served as a motivation for the hardware architecture in this work. F.A. and S.C. formulated the autoencoder based Neuromorphic Ising Machines and designed the benchmark experiments; F.A. optimized the performance of the architecture using graph coloring and FN-annealing schedule, and wrote the Python code running the benchmark experiments. F.A. and S.M. devised the hardware architecture; S.M. refined the hardware architecture, implemented it in Verilog, and benchmarked the architecture on the FPGA. All authors/co-authors contributed to proof-reading and writing of the manuscript.  

\section{Competing interests}\label{sec_COI}
S.C. is named as an inventor on U.S. and international patents associated with FN-based dynamical systems, and the rights to the intellectual property are managed by Washington University in St. Louis. The remaining authors declare no competing interests.

\bibliography{sn-bibliography}


\begin{thebibliography}{58}
\ifx \bisbn   \undefined \def \bisbn  #1{ISBN #1}\fi
\ifx \binits  \undefined \def \binits#1{#1}\fi
\ifx \bauthor  \undefined \def \bauthor#1{#1}\fi
\ifx \batitle  \undefined \def \batitle#1{#1}\fi
\ifx \bjtitle  \undefined \def \bjtitle#1{#1}\fi
\ifx \bvolume  \undefined \def \bvolume#1{\textbf{#1}}\fi
\ifx \byear  \undefined \def \byear#1{#1}\fi
\ifx \bissue  \undefined \def \bissue#1{#1}\fi
\ifx \bfpage  \undefined \def \bfpage#1{#1}\fi
\ifx \blpage  \undefined \def \blpage #1{#1}\fi
\ifx \burl  \undefined \def \burl#1{\textsf{#1}}\fi
\ifx \doiurl  \undefined \def \doiurl#1{\url{https://doi.org/#1}}\fi
\ifx \betal  \undefined \def \betal{\textit{et al.}}\fi
\ifx \binstitute  \undefined \def \binstitute#1{#1}\fi
\ifx \binstitutionaled  \undefined \def \binstitutionaled#1{#1}\fi
\ifx \bctitle  \undefined \def \bctitle#1{#1}\fi
\ifx \beditor  \undefined \def \beditor#1{#1}\fi
\ifx \bpublisher  \undefined \def \bpublisher#1{#1}\fi
\ifx \bbtitle  \undefined \def \bbtitle#1{#1}\fi
\ifx \bedition  \undefined \def \bedition#1{#1}\fi
\ifx \bseriesno  \undefined \def \bseriesno#1{#1}\fi
\ifx \blocation  \undefined \def \blocation#1{#1}\fi
\ifx \bsertitle  \undefined \def \bsertitle#1{#1}\fi
\ifx \bsnm \undefined \def \bsnm#1{#1}\fi
\ifx \bsuffix \undefined \def \bsuffix#1{#1}\fi
\ifx \bparticle \undefined \def \bparticle#1{#1}\fi
\ifx \barticle \undefined \def \barticle#1{#1}\fi
\bibcommenthead
\ifx \bconfdate \undefined \def \bconfdate #1{#1}\fi
\ifx \botherref \undefined \def \botherref #1{#1}\fi
\ifx \url \undefined \def \url#1{\textsf{#1}}\fi
\ifx \bchapter \undefined \def \bchapter#1{#1}\fi
\ifx \bbook \undefined \def \bbook#1{#1}\fi
\ifx \bcomment \undefined \def \bcomment#1{#1}\fi
\ifx \oauthor \undefined \def \oauthor#1{#1}\fi
\ifx \citeauthoryear \undefined \def \citeauthoryear#1{#1}\fi
\ifx \endbibitem  \undefined \def \endbibitem {}\fi
\ifx \bconflocation  \undefined \def \bconflocation#1{#1}\fi
\ifx \arxivurl  \undefined \def \arxivurl#1{\textsf{#1}}\fi
\csname PreBibitemsHook\endcsname

\bibitem[\protect\citeauthoryear{Modha et~al.}{2023}]{Neural_inference_Modha}
\begin{barticle}
\bauthor{\bsnm{Modha}, \binits{D.S.}},
\bauthor{\bsnm{Akopyan}, \binits{F.}},
\bauthor{\bsnm{Andreopoulos}, \binits{A.}},
\bauthor{\bsnm{Appuswamy}, \binits{R.}},
\bauthor{\bsnm{Arthur}, \binits{J.V.}},
\bauthor{\bsnm{Cassidy}, \binits{A.S.}},
\bauthor{\bsnm{Datta}, \binits{P.}},
\bauthor{\bsnm{DeBole}, \binits{M.V.}},
\bauthor{\bsnm{Esser}, \binits{S.K.}},
\bauthor{\bsnm{Otero}, \binits{C.O.}},
\bauthor{\bsnm{Sawada}, \binits{J.}},
\bauthor{\bsnm{Taba}, \binits{B.}},
\bauthor{\bsnm{Amir}, \binits{A.}},
\bauthor{\bsnm{Bablani}, \binits{D.}},
\bauthor{\bsnm{Carlson}, \binits{P.J.}},
\bauthor{\bsnm{Flickner}, \binits{M.D.}},
\bauthor{\bsnm{Gandhasri}, \binits{R.}},
\bauthor{\bsnm{Garreau}, \binits{G.J.}},
\bauthor{\bsnm{Ito}, \binits{M.}},
\bauthor{\bsnm{Klamo}, \binits{J.L.}},
\bauthor{\bsnm{Kusnitz}, \binits{J.A.}},
\bauthor{\bsnm{McClatchey}, \binits{N.J.}},
\bauthor{\bsnm{McKinstry}, \binits{J.L.}},
\bauthor{\bsnm{Nakamura}, \binits{Y.}},
\bauthor{\bsnm{Nayak}, \binits{T.K.}},
\bauthor{\bsnm{Risk}, \binits{W.P.}},
\bauthor{\bsnm{Schleupen}, \binits{K.}},
\bauthor{\bsnm{Shaw}, \binits{B.}},
\bauthor{\bsnm{Sivagnaname}, \binits{J.}},
\bauthor{\bsnm{Smith}, \binits{D.F.}},
\bauthor{\bsnm{Terrizzano}, \binits{I.}},
\bauthor{\bsnm{Ueda}, \binits{T.}}:
\batitle{Neural inference at the frontier of energy, space, and time}.
\bjtitle{Science}
\bvolume{382}(\bissue{6668}),
\bfpage{329}--\blpage{335}
(\byear{2023})
\doiurl{10.1126/science.adh1174}
{\href{https://arxiv.org/abs/https://www.science.org/doi/pdf/10.1126/science.adh1174}{{https://www.science.org/doi/pdf/10.1126/science.adh1174}}}
\end{barticle}
\endbibitem

\bibitem[\protect\citeauthoryear{Merolla et~al.}{2014}]{Merolla_2014}
\begin{barticle}
\bauthor{\bsnm{Merolla}, \binits{P.A.}},
\bauthor{\bsnm{Arthur}, \binits{J.V.}},
\bauthor{\bsnm{Alvarez-Icaza}, \binits{R.}},
\bauthor{\bsnm{Cassidy}, \binits{A.S.}},
\bauthor{\bsnm{Sawada}, \binits{J.}},
\bauthor{\bsnm{Akopyan}, \binits{F.}},
\bauthor{\bsnm{Jackson}, \binits{B.L.}},
\bauthor{\bsnm{Imam}, \binits{N.}},
\bauthor{\bsnm{Guo}, \binits{C.}},
\bauthor{\bsnm{Nakamura}, \binits{Y.}},
\bauthor{\bsnm{Brezzo}, \binits{B.}},
\bauthor{\bsnm{Vo}, \binits{I.}},
\bauthor{\bsnm{Esser}, \binits{S.K.}},
\bauthor{\bsnm{Appuswamy}, \binits{R.}},
\bauthor{\bsnm{Taba}, \binits{B.}},
\bauthor{\bsnm{Amir}, \binits{A.}},
\bauthor{\bsnm{Flickner}, \binits{M.D.}},
\bauthor{\bsnm{Risk}, \binits{W.P.}},
\bauthor{\bsnm{Manohar}, \binits{R.}},
\bauthor{\bsnm{Modha}, \binits{D.S.}}:
\batitle{A million spiking-neuron integrated circuit with a scalable communication network and interface}.
\bjtitle{Science}
\bvolume{345}(\bissue{6197}),
\bfpage{668}--\blpage{673}
(\byear{2014})
\doiurl{10.1126/science.1254642}
{\href{https://arxiv.org/abs/https://www.science.org/doi/pdf/10.1126/science.1254642}{{https://www.science.org/doi/pdf/10.1126/science.1254642}}}
\end{barticle}
\endbibitem

\bibitem[\protect\citeauthoryear{Esser et~al.}{2016}]{Esser2016}
\begin{barticle}
\bauthor{\bsnm{Esser}, \binits{S.K.}},
\bauthor{\bsnm{Merolla}, \binits{P.A.}},
\bauthor{\bsnm{Arthur}, \binits{J.V.}},
\bauthor{\bsnm{Cassidy}, \binits{A.S.}},
\bauthor{\bsnm{Appuswamy}, \binits{R.}},
\bauthor{\bsnm{Andreopoulos}, \binits{A.}},
\bauthor{\bsnm{Berg}, \binits{D.J.}},
\bauthor{\bsnm{McKinstry}, \binits{J.L.}},
\bauthor{\bsnm{Melano}, \binits{T.}},
\bauthor{\bsnm{Barch}, \binits{D.R.}},
\bauthor{\bsnm{Nolfo}, \binits{C.}},
\bauthor{\bsnm{Datta}, \binits{P.}},
\bauthor{\bsnm{Amir}, \binits{A.}},
\bauthor{\bsnm{Taba}, \binits{B.}},
\bauthor{\bsnm{Flickner}, \binits{M.D.}},
\bauthor{\bsnm{Modha}, \binits{D.S.}}:
\batitle{Convolutional networks for fast, energy-efficient neuromorphic computing}.
\bjtitle{Proceedings of the National Academy of Sciences}
\bvolume{113}(\bissue{41}),
\bfpage{11441}--\blpage{11446}
(\byear{2016})
\doiurl{10.1073/pnas.1604850113}
{\href{https://arxiv.org/abs/https://www.pnas.org/doi/pdf/10.1073/pnas.1604850113}{{https://www.pnas.org/doi/pdf/10.1073/pnas.1604850113}}}
\end{barticle}
\endbibitem

\bibitem[\protect\citeauthoryear{Chen et~al.}{2025}]{Chen2025}
\begin{barticle}
\bauthor{\bsnm{Chen}, \binits{Z.}},
\bauthor{\bsnm{Xiao}, \binits{Z.}},
\bauthor{\bsnm{Akl}, \binits{M.}},
\bauthor{\bsnm{Leugring}, \binits{J.}},
\bauthor{\bsnm{Olajide}, \binits{O.}},
\bauthor{\bsnm{Malik}, \binits{A.}},
\bauthor{\bsnm{Dennler}, \binits{N.}},
\bauthor{\bsnm{Harper}, \binits{C.}},
\bauthor{\bsnm{Bose}, \binits{S.}},
\bauthor{\bsnm{Gonzalez}, \binits{H.A.}},
\bauthor{\bsnm{Samaali}, \binits{M.}},
\bauthor{\bsnm{Liu}, \binits{G.}},
\bauthor{\bsnm{Eshraghian}, \binits{J.}},
\bauthor{\bsnm{Pignari}, \binits{R.}},
\bauthor{\bsnm{Urgese}, \binits{G.}},
\bauthor{\bsnm{Andreou}, \binits{A.G.}},
\bauthor{\bsnm{Shankar}, \binits{S.}},
\bauthor{\bsnm{Mayr}, \binits{C.}},
\bauthor{\bsnm{Cauwenberghs}, \binits{G.}},
\bauthor{\bsnm{Chakrabartty}, \binits{S.}}:
\batitle{On-off neuromorphic ising machines using fowler-nordheim annealers}.
\bjtitle{Nature Communications}
\bvolume{16}(\bissue{1}),
\bfpage{3086}
(\byear{2025})
\doiurl{10.1038/s41467-025-58231-5}
\end{barticle}
\endbibitem

\bibitem[\protect\citeauthoryear{Johnson et~al.}{2011}]{Johnson2011}
\begin{barticle}
\bauthor{\bsnm{Johnson}, \binits{M.W.}},
\bauthor{\bsnm{Amin}, \binits{M.H.S.}},
\bauthor{\bsnm{Gildert}, \binits{S.}},
\bauthor{\bsnm{Lanting}, \binits{T.}},
\bauthor{\bsnm{Hamze}, \binits{F.}},
\bauthor{\bsnm{Dickson}, \binits{N.}},
\bauthor{\bsnm{Harris}, \binits{R.}},
\bauthor{\bsnm{Berkley}, \binits{A.J.}},
\bauthor{\bsnm{Johansson}, \binits{J.}},
\bauthor{\bsnm{Bunyk}, \binits{P.}},
\bauthor{\bsnm{Chapple}, \binits{E.M.}},
\bauthor{\bsnm{Enderud}, \binits{C.}},
\bauthor{\bsnm{Hilton}, \binits{J.P.}},
\bauthor{\bsnm{Karimi}, \binits{K.}},
\bauthor{\bsnm{Ladizinsky}, \binits{E.}},
\bauthor{\bsnm{Ladizinsky}, \binits{N.}},
\bauthor{\bsnm{Oh}, \binits{T.}},
\bauthor{\bsnm{Perminov}, \binits{I.}},
\bauthor{\bsnm{Rich}, \binits{C.}},
\bauthor{\bsnm{Thom}, \binits{M.C.}},
\bauthor{\bsnm{Tolkacheva}, \binits{E.}},
\bauthor{\bsnm{Truncik}, \binits{C.J.S.}},
\bauthor{\bsnm{Uchaikin}, \binits{S.}},
\bauthor{\bsnm{Wang}, \binits{J.}},
\bauthor{\bsnm{Wilson}, \binits{B.}},
\bauthor{\bsnm{Rose}, \binits{G.}}:
\batitle{Quantum annealing with manufactured spins}.
\bjtitle{Nature}
\bvolume{473}(\bissue{7346}),
\bfpage{194}--\blpage{198}
(\byear{2011})
\doiurl{10.1038/nature10012}
\end{barticle}
\endbibitem

\bibitem[\protect\citeauthoryear{King et~al.}{2023}]{King2023}
\begin{barticle}
\bauthor{\bsnm{King}, \binits{A.D.}},
\bauthor{\bsnm{Raymond}, \binits{J.}},
\bauthor{\bsnm{Lanting}, \binits{T.}},
\bauthor{\bsnm{Harris}, \binits{R.}},
\bauthor{\bsnm{Zucca}, \binits{A.}},
\bauthor{\bsnm{Altomare}, \binits{F.}},
\bauthor{\bsnm{Berkley}, \binits{A.J.}},
\bauthor{\bsnm{Boothby}, \binits{K.}},
\bauthor{\bsnm{Ejtemaee}, \binits{S.}},
\bauthor{\bsnm{Enderud}, \binits{C.}},
\bauthor{\bsnm{Hoskinson}, \binits{E.}},
\bauthor{\bsnm{Huang}, \binits{S.}},
\bauthor{\bsnm{Ladizinsky}, \binits{E.}},
\bauthor{\bsnm{MacDonald}, \binits{A.J.R.}},
\bauthor{\bsnm{Marsden}, \binits{G.}},
\bauthor{\bsnm{Molavi}, \binits{R.}},
\bauthor{\bsnm{Oh}, \binits{T.}},
\bauthor{\bsnm{Poulin-Lamarre}, \binits{G.}},
\bauthor{\bsnm{Reis}, \binits{M.}},
\bauthor{\bsnm{Rich}, \binits{C.}},
\bauthor{\bsnm{Sato}, \binits{Y.}},
\bauthor{\bsnm{Tsai}, \binits{N.}},
\bauthor{\bsnm{Volkmann}, \binits{M.}},
\bauthor{\bsnm{Whittaker}, \binits{J.D.}},
\bauthor{\bsnm{Yao}, \binits{J.}},
\bauthor{\bsnm{Sandvik}, \binits{A.W.}},
\bauthor{\bsnm{Amin}, \binits{M.H.}}:
\batitle{Quantum critical dynamics in a 5,000-qubit programmable spin glass}.
\bjtitle{Nature}
\bvolume{617}(\bissue{7959}),
\bfpage{61}--\blpage{66}
(\byear{2023})
\doiurl{10.1038/s41586-023-05867-2}
\end{barticle}
\endbibitem

\bibitem[\protect\citeauthoryear{Marandi et~al.}{2014}]{Marandi2014}
\begin{barticle}
\bauthor{\bsnm{Marandi}, \binits{A.}},
\bauthor{\bsnm{Wang}, \binits{Z.}},
\bauthor{\bsnm{Takata}, \binits{K.}},
\bauthor{\bsnm{Byer}, \binits{R.L.}},
\bauthor{\bsnm{Yamamoto}, \binits{Y.}}:
\batitle{Network of time-multiplexed optical parametric oscillators as a coherent ising machine}.
\bjtitle{Nature Photonics}
\bvolume{8}(\bissue{12}),
\bfpage{937}--\blpage{942}
(\byear{2014})
\doiurl{10.1038/nphoton.2014.249}
\end{barticle}
\endbibitem

\bibitem[\protect\citeauthoryear{McMahon et~al.}{2016}]{McMahon2016}
\begin{barticle}
\bauthor{\bsnm{McMahon}, \binits{P.L.}},
\bauthor{\bsnm{Marandi}, \binits{A.}},
\bauthor{\bsnm{Haribara}, \binits{Y.}},
\bauthor{\bsnm{Hamerly}, \binits{R.}},
\bauthor{\bsnm{Langrock}, \binits{C.}},
\bauthor{\bsnm{Tamate}, \binits{S.}},
\bauthor{\bsnm{Inagaki}, \binits{T.}},
\bauthor{\bsnm{Takesue}, \binits{H.}},
\bauthor{\bsnm{Utsunomiya}, \binits{S.}},
\bauthor{\bsnm{Aihara}, \binits{K.}},
\bauthor{\bsnm{Byer}, \binits{R.L.}},
\bauthor{\bsnm{Fejer}, \binits{M.M.}},
\bauthor{\bsnm{Mabuchi}, \binits{H.}},
\bauthor{\bsnm{Yamamoto}, \binits{Y.}}:
\batitle{A fully programmable 100-spin coherent ising machine with all-to-all connections}.
\bjtitle{Science}
\bvolume{354}(\bissue{6312}),
\bfpage{614}--\blpage{617}
(\byear{2016})
\doiurl{10.1126/science.aah5178}
{\href{https://arxiv.org/abs/https://www.science.org/doi/pdf/10.1126/science.aah5178}{{https://www.science.org/doi/pdf/10.1126/science.aah5178}}}
\end{barticle}
\endbibitem

\bibitem[\protect\citeauthoryear{Wang and Roychowdhury}{2019}]{Wang2019}
\begin{bchapter}
\bauthor{\bsnm{Wang}, \binits{T.}},
\bauthor{\bsnm{Roychowdhury}, \binits{J.}}:
\bctitle{Oim: Oscillator-based ising machines for solving combinatorial optimisation problems}.
In: \beditor{\bsnm{McQuillan}, \binits{I.}},
\beditor{\bsnm{Seki}, \binits{S.}} (eds.)
\bbtitle{Unconventional Computation and Natural Computation},
pp. \bfpage{232}--\blpage{256}.
\bpublisher{Springer},
\blocation{Cham}
(\byear{2019})
\end{bchapter}
\endbibitem

\bibitem[\protect\citeauthoryear{Chou et~al.}{2019}]{Chou2019}
\begin{barticle}
\bauthor{\bsnm{Chou}, \binits{J.}},
\bauthor{\bsnm{Bramhavar}, \binits{S.}},
\bauthor{\bsnm{Ghosh}, \binits{S.}},
\bauthor{\bsnm{Herzog}, \binits{W.}}:
\batitle{Analog coupled oscillator based weighted ising machine}.
\bjtitle{Scientific Reports}
\bvolume{9}(\bissue{1}),
\bfpage{14786}
(\byear{2019})
\doiurl{10.1038/s41598-019-49699-5}
\end{barticle}
\endbibitem

\bibitem[\protect\citeauthoryear{Sutton et~al.}{2017}]{Sutton2017}
\begin{barticle}
\bauthor{\bsnm{Sutton}, \binits{B.}},
\bauthor{\bsnm{Camsari}, \binits{K.Y.}},
\bauthor{\bsnm{Behin-Aein}, \binits{B.}},
\bauthor{\bsnm{Datta}, \binits{S.}}:
\batitle{Intrinsic optimization using stochastic nanomagnets}.
\bjtitle{Scientific Reports}
\bvolume{7}(\bissue{1}),
\bfpage{44370}
(\byear{2017})
\doiurl{10.1038/srep44370}
\end{barticle}
\endbibitem

\bibitem[\protect\citeauthoryear{Aadit et~al.}{2022}]{Aadit2022}
\begin{barticle}
\bauthor{\bsnm{Aadit}, \binits{N.A.}},
\bauthor{\bsnm{Grimaldi}, \binits{A.}},
\bauthor{\bsnm{Carpentieri}, \binits{M.}},
\bauthor{\bsnm{Theogarajan}, \binits{L.}},
\bauthor{\bsnm{Martinis}, \binits{J.M.}},
\bauthor{\bsnm{Finocchio}, \binits{G.}},
\bauthor{\bsnm{Camsari}, \binits{K.Y.}}:
\batitle{Massively parallel probabilistic computing with sparse ising machines}.
\bjtitle{Nature Electronics}
\bvolume{5}(\bissue{7}),
\bfpage{460}--\blpage{468}
(\byear{2022})
\doiurl{10.1038/s41928-022-00774-2}
\end{barticle}
\endbibitem

\bibitem[\protect\citeauthoryear{Hopfield}{1982}]{Hopfield1982}
\begin{barticle}
\bauthor{\bsnm{Hopfield}, \binits{J.J.}}:
\batitle{Neural networks and physical systems with emergent collective computational abilities.}
\bjtitle{Proceedings of the National Academy of Sciences}
\bvolume{79}(\bissue{8}),
\bfpage{2554}--\blpage{2558}
(\byear{1982})
\doiurl{10.1073/pnas.79.8.2554}
{\href{https://arxiv.org/abs/https://www.pnas.org/doi/pdf/10.1073/pnas.79.8.2554}{{https://www.pnas.org/doi/pdf/10.1073/pnas.79.8.2554}}}
\end{barticle}
\endbibitem

\bibitem[\protect\citeauthoryear{Cai et~al.}{2020}]{Cai2020}
\begin{barticle}
\bauthor{\bsnm{Cai}, \binits{F.}},
\bauthor{\bsnm{Kumar}, \binits{S.}},
\bauthor{\bsnm{Van~Vaerenbergh}, \binits{T.}},
\bauthor{\bsnm{Sheng}, \binits{X.}},
\bauthor{\bsnm{Liu}, \binits{R.}},
\bauthor{\bsnm{Li}, \binits{C.}},
\bauthor{\bsnm{Liu}, \binits{Z.}},
\bauthor{\bsnm{Foltin}, \binits{M.}},
\bauthor{\bsnm{Yu}, \binits{S.}},
\bauthor{\bsnm{Xia}, \binits{Q.}},
\bauthor{\bsnm{Yang}, \binits{J.J.}},
\bauthor{\bsnm{Beausoleil}, \binits{R.}},
\bauthor{\bsnm{Lu}, \binits{W.D.}},
\bauthor{\bsnm{Strachan}, \binits{J.P.}}:
\batitle{Power-efficient combinatorial optimization using intrinsic noise in memristor hopfield neural networks}.
\bjtitle{Nature Electronics}
\bvolume{3}(\bissue{7}),
\bfpage{409}--\blpage{418}
(\byear{2020})
\doiurl{10.1038/s41928-020-0436-6}
\end{barticle}
\endbibitem

\bibitem[\protect\citeauthoryear{Maher et~al.}{2024}]{Maher2024CMOSVOISING}
\begin{barticle}
\bauthor{\bsnm{Maher}, \binits{O.}},
\bauthor{\bsnm{Jim{\'e}nez}, \binits{M.}},
\bauthor{\bsnm{Delacour}, \binits{C.}},
\bauthor{\bsnm{Harnack}, \binits{N.}},
\bauthor{\bsnm{N{\'u}{\~{n}}ez}, \binits{J.}},
\bauthor{\bsnm{Avedillo}, \binits{M.J.}},
\bauthor{\bsnm{Linares-Barranco}, \binits{B.}},
\bauthor{\bsnm{Todri-Sanial}, \binits{A.}},
\bauthor{\bsnm{Indiveri}, \binits{G.}},
\bauthor{\bsnm{Karg}, \binits{S.}}:
\batitle{A cmos-compatible oscillation-based vo2 ising machine solver}.
\bjtitle{Nature Communications}
\bvolume{15}(\bissue{1}),
\bfpage{3334}
(\byear{2024})
\doiurl{10.1038/s41467-024-47642-5}
\end{barticle}
\endbibitem

\bibitem[\protect\citeauthoryear{Yik et~al.}{2025}]{Yik2025}
\begin{barticle}
\bauthor{\bsnm{Yik}, \binits{J.}},
\bauthor{\bsnm{Berghe}, \binits{K.}},
\bauthor{\bsnm{Blanken}, \binits{D.}},
\bauthor{\bsnm{Bouhadjar}, \binits{Y.}},
\bauthor{\bsnm{Fabre}, \binits{M.}},
\bauthor{\bsnm{Hueber}, \binits{P.}},
\bauthor{\bsnm{Ke}, \binits{W.}},
\bauthor{\bsnm{Khoei}, \binits{M.A.}},
\bauthor{\bsnm{Kleyko}, \binits{D.}},
\bauthor{\bsnm{Pacik-Nelson}, \binits{N.}},
\bauthor{\bsnm{Pierro}, \binits{A.}},
\bauthor{\bsnm{Stratmann}, \binits{P.}},
\bauthor{\bsnm{Sun}, \binits{P.-S.V.}},
\bauthor{\bsnm{Tang}, \binits{G.}},
\bauthor{\bsnm{Wang}, \binits{S.}},
\bauthor{\bsnm{Zhou}, \binits{B.}},
\bauthor{\bsnm{Ahmed}, \binits{S.H.}},
\bauthor{\bsnm{Vathakkattil~Joseph}, \binits{G.}},
\bauthor{\bsnm{Leto}, \binits{B.}},
\bauthor{\bsnm{Micheli}, \binits{A.}},
\bauthor{\bsnm{Mishra}, \binits{A.K.}},
\bauthor{\bsnm{Lenz}, \binits{G.}},
\bauthor{\bsnm{Sun}, \binits{T.}},
\bauthor{\bsnm{Ahmed}, \binits{Z.}},
\bauthor{\bsnm{Akl}, \binits{M.}},
\bauthor{\bsnm{Anderson}, \binits{B.}},
\bauthor{\bsnm{Andreou}, \binits{A.G.}},
\bauthor{\bsnm{Bartolozzi}, \binits{C.}},
\bauthor{\bsnm{Basu}, \binits{A.}},
\bauthor{\bsnm{Bogdan}, \binits{P.}},
\bauthor{\bsnm{Bohte}, \binits{S.}},
\bauthor{\bsnm{Buckley}, \binits{S.}},
\bauthor{\bsnm{Cauwenberghs}, \binits{G.}},
\bauthor{\bsnm{Chicca}, \binits{E.}},
\bauthor{\bsnm{Corradi}, \binits{F.}},
\bauthor{\bsnm{Croon}, \binits{G.}},
\bauthor{\bsnm{Danielescu}, \binits{A.}},
\bauthor{\bsnm{Daram}, \binits{A.}},
\bauthor{\bsnm{Davies}, \binits{M.}},
\bauthor{\bsnm{Demirag}, \binits{Y.}},
\bauthor{\bsnm{Eshraghian}, \binits{J.}},
\bauthor{\bsnm{Fischer}, \binits{T.}},
\bauthor{\bsnm{Forest}, \binits{J.}},
\bauthor{\bsnm{Fra}, \binits{V.}},
\bauthor{\bsnm{Furber}, \binits{S.}},
\bauthor{\bsnm{Furlong}, \binits{P.M.}},
\bauthor{\bsnm{Gilpin}, \binits{W.}},
\bauthor{\bsnm{Gilra}, \binits{A.}},
\bauthor{\bsnm{Gonzalez}, \binits{H.A.}},
\bauthor{\bsnm{Indiveri}, \binits{G.}},
\bauthor{\bsnm{Joshi}, \binits{S.}},
\bauthor{\bsnm{Karia}, \binits{V.}},
\bauthor{\bsnm{Khacef}, \binits{L.}},
\bauthor{\bsnm{Knight}, \binits{J.C.}},
\bauthor{\bsnm{Kriener}, \binits{L.}},
\bauthor{\bsnm{Kubendran}, \binits{R.}},
\bauthor{\bsnm{Kudithipudi}, \binits{D.}},
\bauthor{\bsnm{Liu}, \binits{S.-C.}},
\bauthor{\bsnm{Liu}, \binits{Y.-H.}},
\bauthor{\bsnm{Ma}, \binits{H.}},
\bauthor{\bsnm{Manohar}, \binits{R.}},
\bauthor{\bsnm{Margarit-Taul{\'e}}, \binits{J.M.}},
\bauthor{\bsnm{Mayr}, \binits{C.}},
\bauthor{\bsnm{Michmizos}, \binits{K.}},
\bauthor{\bsnm{Muir}, \binits{D.R.}},
\bauthor{\bsnm{Neftci}, \binits{E.}},
\bauthor{\bsnm{Nowotny}, \binits{T.}},
\bauthor{\bsnm{Ottati}, \binits{F.}},
\bauthor{\bsnm{Ozcelikkale}, \binits{A.}},
\bauthor{\bsnm{Panda}, \binits{P.}},
\bauthor{\bsnm{Park}, \binits{J.}},
\bauthor{\bsnm{Payvand}, \binits{M.}},
\bauthor{\bsnm{Pehle}, \binits{C.}},
\bauthor{\bsnm{Petrovici}, \binits{M.A.}},
\bauthor{\bsnm{Posch}, \binits{C.}},
\bauthor{\bsnm{Renner}, \binits{A.}},
\bauthor{\bsnm{Sandamirskaya}, \binits{Y.}},
\bauthor{\bsnm{Schaefer}, \binits{C.J.S.}},
\bauthor{\bsnm{Schaik}, \binits{A.}},
\bauthor{\bsnm{Schemmel}, \binits{J.}},
\bauthor{\bsnm{Schmidgall}, \binits{S.}},
\bauthor{\bsnm{Schuman}, \binits{C.}},
\bauthor{\bsnm{Seo}, \binits{J.-s.}},
\bauthor{\bsnm{Sheik}, \binits{S.}},
\bauthor{\bsnm{Shrestha}, \binits{S.B.}},
\bauthor{\bsnm{Sifalakis}, \binits{M.}},
\bauthor{\bsnm{Sironi}, \binits{A.}},
\bauthor{\bsnm{Stewart}, \binits{K.}},
\bauthor{\bsnm{Stewart}, \binits{M.}},
\bauthor{\bsnm{Stewart}, \binits{T.C.}},
\bauthor{\bsnm{Timcheck}, \binits{J.}},
\bauthor{\bsnm{T{\"o}men}, \binits{N.}},
\bauthor{\bsnm{Urgese}, \binits{G.}},
\bauthor{\bsnm{Verhelst}, \binits{M.}},
\bauthor{\bsnm{Vineyard}, \binits{C.M.}},
\bauthor{\bsnm{Vogginger}, \binits{B.}},
\bauthor{\bsnm{Yousefzadeh}, \binits{A.}},
\bauthor{\bsnm{Zohora}, \binits{F.T.}},
\bauthor{\bsnm{Frenkel}, \binits{C.}},
\bauthor{\bsnm{Reddi}, \binits{V.J.}}:
\batitle{The neurobench framework for benchmarking neuromorphic computing algorithms and systems}.
\bjtitle{Nature Communications}
\bvolume{16}(\bissue{1}),
\bfpage{1545}
(\byear{2025})
\doiurl{10.1038/s41467-025-56739-4}
\end{barticle}
\endbibitem

\bibitem[\protect\citeauthoryear{Mallick et~al.}{2023}]{Mallick2023}
\begin{barticle}
\bauthor{\bsnm{Mallick}, \binits{A.}},
\bauthor{\bsnm{Zhao}, \binits{Z.}},
\bauthor{\bsnm{Bashar}, \binits{M.K.}},
\bauthor{\bsnm{Alam}, \binits{S.}},
\bauthor{\bsnm{Islam}, \binits{M.M.}},
\bauthor{\bsnm{Xiao}, \binits{Y.}},
\bauthor{\bsnm{Xu}, \binits{Y.}},
\bauthor{\bsnm{Aziz}, \binits{A.}},
\bauthor{\bsnm{Narayanan}, \binits{V.}},
\bauthor{\bsnm{Ni}, \binits{K.}},
\bauthor{\bsnm{Shukla}, \binits{N.}}:
\batitle{Cmos-compatible ising machines built using bistable latches coupled through ferroelectric transistor arrays}.
\bjtitle{Scientific Reports}
\bvolume{13}(\bissue{1}),
\bfpage{1515}
(\byear{2023})
\doiurl{10.1038/s41598-023-28217-8}
\end{barticle}
\endbibitem

\bibitem[\protect\citeauthoryear{Sharma et~al.}{2023}]{Sharma2023}
\begin{barticle}
\bauthor{\bsnm{Sharma}, \binits{A.}},
\bauthor{\bsnm{Burns}, \binits{M.}},
\bauthor{\bsnm{Hahn}, \binits{A.}},
\bauthor{\bsnm{Huang}, \binits{M.}}:
\batitle{Augmenting an electronic ising machine to effectively solve boolean satisfiability}.
\bjtitle{Scientific Reports}
\bvolume{13}(\bissue{1}),
\bfpage{22858}
(\byear{2023})
\doiurl{10.1038/s41598-023-49966-6}
\end{barticle}
\endbibitem

\bibitem[\protect\citeauthoryear{Albertsson and Rusu}{2023}]{Albertsson2023}
\begin{barticle}
\bauthor{\bsnm{Albertsson}, \binits{D.I.}},
\bauthor{\bsnm{Rusu}, \binits{A.}}:
\batitle{Highly reconfigurable oscillator-based ising machine through quasiperiodic modulation of coupling strength}.
\bjtitle{Scientific Reports}
\bvolume{13}(\bissue{1}),
\bfpage{4005}
(\byear{2023})
\doiurl{10.1038/s41598-023-31155-0}
\end{barticle}
\endbibitem

\bibitem[\protect\citeauthoryear{Vaidya et~al.}{2022}]{Vaidya2022}
\begin{barticle}
\bauthor{\bsnm{Vaidya}, \binits{J.}},
\bauthor{\bsnm{Surya~Kanthi}, \binits{R.S.}},
\bauthor{\bsnm{Shukla}, \binits{N.}}:
\batitle{Creating electronic oscillator-based ising machines without external injection locking}.
\bjtitle{Scientific Reports}
\bvolume{12}(\bissue{1}),
\bfpage{981}
(\byear{2022})
\doiurl{10.1038/s41598-021-04057-2}
\end{barticle}
\endbibitem

\bibitem[\protect\citeauthoryear{Dutta et~al.}{2021}]{Dutta2021}
\begin{barticle}
\bauthor{\bsnm{Dutta}, \binits{S.}},
\bauthor{\bsnm{Khanna}, \binits{A.}},
\bauthor{\bsnm{Assoa}, \binits{A.S.}},
\bauthor{\bsnm{Paik}, \binits{H.}},
\bauthor{\bsnm{Schlom}, \binits{D.G.}},
\bauthor{\bsnm{Toroczkai}, \binits{Z.}},
\bauthor{\bsnm{Raychowdhury}, \binits{A.}},
\bauthor{\bsnm{Datta}, \binits{S.}}:
\batitle{An ising hamiltonian solver based on coupled stochastic phase-transition nano-oscillators}.
\bjtitle{Nature Electronics}
\bvolume{4}(\bissue{7}),
\bfpage{502}--\blpage{512}
(\byear{2021})
\doiurl{10.1038/s41928-021-00616-7}
\end{barticle}
\endbibitem

\bibitem[\protect\citeauthoryear{Yamaoka et~al.}{2016}]{Yamaoka2016}
\begin{barticle}
\bauthor{\bsnm{Yamaoka}, \binits{M.}},
\bauthor{\bsnm{Yoshimura}, \binits{C.}},
\bauthor{\bsnm{Hayashi}, \binits{M.}},
\bauthor{\bsnm{Okuyama}, \binits{T.}},
\bauthor{\bsnm{Aoki}, \binits{H.}},
\bauthor{\bsnm{Mizuno}, \binits{H.}}:
\batitle{A 20k-spin ising chip to solve combinatorial optimization problems with cmos annealing}.
\bjtitle{IEEE Journal of Solid-State Circuits}
\bvolume{51}(\bissue{1}),
\bfpage{303}--\blpage{309}
(\byear{2016})
\doiurl{10.1109/JSSC.2015.2498601}
\end{barticle}
\endbibitem

\bibitem[\protect\citeauthoryear{Takemoto et~al.}{2019}]{Takemoto2019}
\begin{bchapter}
\bauthor{\bsnm{Takemoto}, \binits{T.}},
\bauthor{\bsnm{Hayashi}, \binits{M.}},
\bauthor{\bsnm{Yoshimura}, \binits{C.}},
\bauthor{\bsnm{Yamaoka}, \binits{M.}}:
\bctitle{2.6 a 2 ×30k-spin multichip scalable annealing processor based on a processing-in-memory approach for solving large-scale combinatorial optimization problems}.
In: \bbtitle{2019 IEEE International Solid-State Circuits Conference - (ISSCC)},
pp. \bfpage{52}--\blpage{54}
(\byear{2019}).
\doiurl{10.1109/ISSCC.2019.8662517}
\end{bchapter}
\endbibitem

\bibitem[\protect\citeauthoryear{Davies et~al.}{2018}]{Loihi}
\begin{barticle}
\bauthor{\bsnm{Davies}, \binits{M.}},
\bauthor{\bsnm{Srinivasa}, \binits{N.}},
\bauthor{\bsnm{Lin}, \binits{T.-H.}},
\bauthor{\bsnm{Chinya}, \binits{G.}},
\bauthor{\bsnm{Cao}, \binits{Y.}},
\bauthor{\bsnm{Choday}, \binits{S.H.}},
\bauthor{\bsnm{Dimou}, \binits{G.}},
\bauthor{\bsnm{Joshi}, \binits{P.}},
\bauthor{\bsnm{Imam}, \binits{N.}},
\bauthor{\bsnm{Jain}, \binits{S.}},
\bauthor{\bsnm{Liao}, \binits{Y.}},
\bauthor{\bsnm{Lin}, \binits{C.-K.}},
\bauthor{\bsnm{Lines}, \binits{A.}},
\bauthor{\bsnm{Liu}, \binits{R.}},
\bauthor{\bsnm{Mathaikutty}, \binits{D.}},
\bauthor{\bsnm{McCoy}, \binits{S.}},
\bauthor{\bsnm{Paul}, \binits{A.}},
\bauthor{\bsnm{Tse}, \binits{J.}},
\bauthor{\bsnm{Venkataramanan}, \binits{G.}},
\bauthor{\bsnm{Weng}, \binits{Y.-H.}},
\bauthor{\bsnm{Wild}, \binits{A.}},
\bauthor{\bsnm{Yang}, \binits{Y.}},
\bauthor{\bsnm{Wang}, \binits{H.}}:
\batitle{Loihi: A neuromorphic manycore processor with on-chip learning}.
\bjtitle{IEEE Micro}
\bvolume{38}(\bissue{1}),
\bfpage{82}--\blpage{99}
(\byear{2018})
\doiurl{10.1109/MM.2018.112130359}
\end{barticle}
\endbibitem

\bibitem[\protect\citeauthoryear{Mayr et~al.}{2019}]{mayr2019spinnaker210million}
\begin{botherref}
\oauthor{\bsnm{Mayr}, \binits{C.}},
\oauthor{\bsnm{Hoeppner}, \binits{S.}},
\oauthor{\bsnm{Furber}, \binits{S.}}:
SpiNNaker 2: A 10 Million Core Processor System for Brain Simulation and Machine Learning
(2019).
\url{https://arxiv.org/abs/1911.02385}
\end{botherref}
\endbibitem

\bibitem[\protect\citeauthoryear{Lucas}{2014}]{Lucas2014}
\begin{botherref}
\oauthor{\bsnm{Lucas}, \binits{A.}}:
Ising formulations of many np problems.
Frontiers in Physics
\textbf{Volume 2 - 2014}
(2014)
\doiurl{10.3389/fphy.2014.00005}
\end{botherref}
\endbibitem

\bibitem[\protect\citeauthoryear{Boros and Hammer}{2002}]{BOROS2002155}
\begin{barticle}
\bauthor{\bsnm{Boros}, \binits{E.}},
\bauthor{\bsnm{Hammer}, \binits{P.L.}}:
\batitle{Pseudo-boolean optimization}.
\bjtitle{Discrete Applied Mathematics}
\bvolume{123}(\bissue{1}),
\bfpage{155}--\blpage{225}
(\byear{2002})
\doiurl{10.1016/S0166-218X(01)00341-9}
\end{barticle}
\endbibitem

\bibitem[\protect\citeauthoryear{Freedman and Drineas}{2005}]{Freedman2005}
\begin{bchapter}
\bauthor{\bsnm{Freedman}, \binits{D.}},
\bauthor{\bsnm{Drineas}, \binits{P.}}:
\bctitle{Energy minimization via graph cuts: settling what is possible}.
In: \bbtitle{2005 IEEE Computer Society Conference on Computer Vision and Pattern Recognition (CVPR'05)},
vol. \bseriesno{2},
pp. \bfpage{939}--\blpage{9462}
(\byear{2005}).
\doiurl{10.1109/CVPR.2005.143}
\end{bchapter}
\endbibitem

\bibitem[\protect\citeauthoryear{Ishikawa}{2011}]{Ishikawa2011}
\begin{barticle}
\bauthor{\bsnm{Ishikawa}, \binits{H.}}:
\batitle{Transformation of general binary mrf minimization to the first-order case}.
\bjtitle{IEEE Transactions on Pattern Analysis and Machine Intelligence}
\bvolume{33}(\bissue{6}),
\bfpage{1234}--\blpage{1249}
(\byear{2011})
\doiurl{10.1109/TPAMI.2010.91}
\end{barticle}
\endbibitem

\bibitem[\protect\citeauthoryear{Nikhar et~al.}{2024}]{Nikhar2024}
\begin{barticle}
\bauthor{\bsnm{Nikhar}, \binits{S.}},
\bauthor{\bsnm{Kannan}, \binits{S.}},
\bauthor{\bsnm{Aadit}, \binits{N.A.}},
\bauthor{\bsnm{Chowdhury}, \binits{S.}},
\bauthor{\bsnm{Camsari}, \binits{K.Y.}}:
\batitle{All-to-all reconfigurability with sparse and higher-order ising machines}.
\bjtitle{Nature Communications}
\bvolume{15}(\bissue{1}),
\bfpage{8977}
(\byear{2024})
\doiurl{10.1038/s41467-024-53270-w}
\end{barticle}
\endbibitem

\bibitem[\protect\citeauthoryear{Hizzani et~al.}{2024}]{Memristor_based_HOIM}
\begin{bchapter}
\bauthor{\bsnm{Hizzani}, \binits{M.}},
\bauthor{\bsnm{Heittmann}, \binits{A.}},
\bauthor{\bsnm{Hutchinson}, \binits{G.}},
\bauthor{\bsnm{Dobrynin}, \binits{D.}},
\bauthor{\bsnm{Van~Vaerenbergh}, \binits{T.}},
\bauthor{\bsnm{Bhattacharya}, \binits{T.}},
\bauthor{\bsnm{Renaudineau}, \binits{A.}},
\bauthor{\bsnm{Strukov}, \binits{D.}},
\bauthor{\bsnm{Strachan}, \binits{J.P.}}:
\bctitle{Memristor-based hardware and algorithms for higher-order hopfield optimization solver outperforming quadratic ising machines}.
In: \bbtitle{2024 IEEE International Symposium on Circuits and Systems (ISCAS)},
pp. \bfpage{1}--\blpage{5}
(\byear{2024}).
\doiurl{10.1109/ISCAS58744.2024.10558658}
\end{bchapter}
\endbibitem

\bibitem[\protect\citeauthoryear{Bybee et~al.}{2023}]{Bybee2023}
\begin{barticle}
\bauthor{\bsnm{Bybee}, \binits{C.}},
\bauthor{\bsnm{Kleyko}, \binits{D.}},
\bauthor{\bsnm{Nikonov}, \binits{D.E.}},
\bauthor{\bsnm{Khosrowshahi}, \binits{A.}},
\bauthor{\bsnm{Olshausen}, \binits{B.A.}},
\bauthor{\bsnm{Sommer}, \binits{F.T.}}:
\batitle{Efficient optimization with higher-order ising machines}.
\bjtitle{Nature Communications}
\bvolume{14}(\bissue{1}),
\bfpage{6033}
(\byear{2023})
\doiurl{10.1038/s41467-023-41214-9}
\end{barticle}
\endbibitem

\bibitem[\protect\citeauthoryear{Tatsumura et~al.}{2021}]{Tatsumura2021}
\begin{barticle}
\bauthor{\bsnm{Tatsumura}, \binits{K.}},
\bauthor{\bsnm{Yamasaki}, \binits{M.}},
\bauthor{\bsnm{Goto}, \binits{H.}}:
\batitle{Scaling out ising machines using a multi-chip architecture for simulated bifurcation}.
\bjtitle{Nature Electronics}
\bvolume{4}(\bissue{3}),
\bfpage{208}--\blpage{217}
(\byear{2021})
\doiurl{10.1038/s41928-021-00546-4}
\end{barticle}
\endbibitem

\bibitem[\protect\citeauthoryear{Wang et~al.}{2025}]{Wang2025}
\begin{barticle}
\bauthor{\bsnm{Wang}, \binits{H.}},
\bauthor{\bsnm{Liu}, \binits{Z.}},
\bauthor{\bsnm{Xie}, \binits{Z.}},
\bauthor{\bsnm{Li}, \binits{L.}},
\bauthor{\bsnm{Miao}, \binits{Z.}},
\bauthor{\bsnm{Cui}, \binits{W.}},
\bauthor{\bsnm{Pan}, \binits{Y.}}:
\batitle{Parallel ising annealer via gradient-based hamiltonian monte carlo}.
\bjtitle{Quantum Machine Intelligence}
\bvolume{7}(\bissue{1}),
\bfpage{6}
(\byear{2025})
\doiurl{10.1007/s42484-024-00231-7}
\end{barticle}
\endbibitem

\bibitem[\protect\citeauthoryear{Yamamoto et~al.}{2021}]{Yamamoto2021}
\begin{barticle}
\bauthor{\bsnm{Yamamoto}, \binits{K.}},
\bauthor{\bsnm{Kawamura}, \binits{K.}},
\bauthor{\bsnm{Ando}, \binits{K.}},
\bauthor{\bsnm{Mertig}, \binits{N.}},
\bauthor{\bsnm{Takemoto}, \binits{T.}},
\bauthor{\bsnm{Yamaoka}, \binits{M.}},
\bauthor{\bsnm{Teramoto}, \binits{H.}},
\bauthor{\bsnm{Sakai}, \binits{A.}},
\bauthor{\bsnm{Takamaeda-Yamazaki}, \binits{S.}},
\bauthor{\bsnm{Motomura}, \binits{M.}}:
\batitle{Statica: A 512-spin 0.25m-weight annealing processor with an all-spin-updates-at-once architecture for combinatorial optimization with complete spin–spin interactions}.
\bjtitle{IEEE Journal of Solid-State Circuits}
\bvolume{56}(\bissue{1}),
\bfpage{165}--\blpage{178}
(\byear{2021})
\doiurl{10.1109/JSSC.2020.3027702}
\end{barticle}
\endbibitem

\bibitem[\protect\citeauthoryear{Nandi et~al.}{2024}]{Nandi2024XORSAT}
\begin{botherref}
\oauthor{\bsnm{Nandi}, \binits{A.}},
\oauthor{\bsnm{Chakrabartty}, \binits{S.}},
\oauthor{\bsnm{Thakur}, \binits{C.S.}}:
Margin propagation based xor-sat solvers for decoding of ldpc codes.
IEEE Transactions on Communications,
1--1
(2024)
\doiurl{10.1109/TCOMM.2024.3519519}
\end{botherref}
\endbibitem

\bibitem[\protect\citeauthoryear{Zhou and Chakrabartty}{2017}]{Zhou2017FNtimer}
\begin{barticle}
\bauthor{\bsnm{Zhou}, \binits{L.}},
\bauthor{\bsnm{Chakrabartty}, \binits{S.}}:
\batitle{Self-powered timekeeping and synchronization using fowler–nordheim tunneling-based floating-gate integrators}.
\bjtitle{IEEE Transactions on Electron Devices}
\bvolume{64}(\bissue{3}),
\bfpage{1254}--\blpage{1260}
(\byear{2017})
\doiurl{10.1109/TED.2016.2645379}
\end{barticle}
\endbibitem

\bibitem[\protect\citeauthoryear{Ye}{2003}]{Gset}
\begin{botherref}
\oauthor{\bsnm{Ye}, \binits{Y.}}:
The Gset Dataset.
\url{https://web.stanford.edu/~yyye/yyye/Gset/}
(2003)
\end{botherref}
\endbibitem

\bibitem[\protect\citeauthoryear{Goto et~al.}{2021}]{SBM_Toshiba}
\begin{barticle}
\bauthor{\bsnm{Goto}, \binits{H.}},
\bauthor{\bsnm{Endo}, \binits{K.}},
\bauthor{\bsnm{Suzuki}, \binits{M.}},
\bauthor{\bsnm{Sakai}, \binits{Y.}},
\bauthor{\bsnm{Kanao}, \binits{T.}},
\bauthor{\bsnm{Hamakawa}, \binits{Y.}},
\bauthor{\bsnm{Hidaka}, \binits{R.}},
\bauthor{\bsnm{Yamasaki}, \binits{M.}},
\bauthor{\bsnm{Tatsumura}, \binits{K.}}:
\batitle{High-performance combinatorial optimization based on classical mechanics}.
\bjtitle{Science Advances}
\bvolume{7}(\bissue{6}),
\bfpage{7953}
(\byear{2021})
\doiurl{10.1126/sciadv.abe7953}
{\href{https://arxiv.org/abs/https://www.science.org/doi/pdf/10.1126/sciadv.abe7953}{{https://www.science.org/doi/pdf/10.1126/sciadv.abe7953}}}
\end{barticle}
\endbibitem

\bibitem[\protect\citeauthoryear{Krzakala and Zdeborová}{2010}]{Krzakala_2010}
\begin{barticle}
\bauthor{\bsnm{Krzakala}, \binits{F.}},
\bauthor{\bsnm{Zdeborová}, \binits{L.}}:
\batitle{Following gibbs states adiabatically —the energy landscape of mean-field glassy systems}.
\bjtitle{Europhysics Letters}
\bvolume{90}(\bissue{6}),
\bfpage{66002}
(\byear{2010})
\doiurl{10.1209/0295-5075/90/66002}
\end{barticle}
\endbibitem

\bibitem[\protect\citeauthoryear{Ricci-Tersenghi}{2010}]{Glassy_without_hard}
\begin{barticle}
\bauthor{\bsnm{Ricci-Tersenghi}, \binits{F.}}:
\batitle{Being glassy without being hard to solve}.
\bjtitle{Science}
\bvolume{330}(\bissue{6011}),
\bfpage{1639}--\blpage{1640}
(\byear{2010})
\doiurl{10.1126/science.1189804}
{\href{https://arxiv.org/abs/https://www.science.org/doi/pdf/10.1126/science.1189804}{{https://www.science.org/doi/pdf/10.1126/science.1189804}}}
\end{barticle}
\endbibitem

\bibitem[\protect\citeauthoryear{Stopfer et~al.}{2003}]{Population_firing_dynamics}
\begin{barticle}
\bauthor{\bsnm{Stopfer}, \binits{M.}},
\bauthor{\bsnm{Jayaraman}, \binits{V.}},
\bauthor{\bsnm{Laurent}, \binits{G.}}:
\batitle{Intensity versus identity coding in an olfactory system}.
\bjtitle{Neuron}
\bvolume{39}(\bissue{6}),
\bfpage{991}--\blpage{1004}
(\byear{2003})
\doiurl{10.1016/j.neuron.2003.08.011}
\end{barticle}
\endbibitem

\bibitem[\protect\citeauthoryear{Goto et~al.}{2019}]{Goto2019SBM}
\begin{barticle}
\bauthor{\bsnm{Goto}, \binits{H.}},
\bauthor{\bsnm{Tatsumura}, \binits{K.}},
\bauthor{\bsnm{Dixon}, \binits{A.R.}}:
\batitle{Combinatorial optimization by simulating adiabatic bifurcations in nonlinear hamiltonian systems}.
\bjtitle{Science Advances}
\bvolume{5}(\bissue{4}),
\bfpage{2372}
(\byear{2019})
\doiurl{10.1126/sciadv.aav2372}
{\href{https://arxiv.org/abs/https://www.science.org/doi/pdf/10.1126/sciadv.aav2372}{{https://www.science.org/doi/pdf/10.1126/sciadv.aav2372}}}
\end{barticle}
\endbibitem

\bibitem[\protect\citeauthoryear{Br\'{e}laz}{1979}]{DSATUR}
\begin{barticle}
\bauthor{\bsnm{Br\'{e}laz}, \binits{D.}}:
\batitle{New methods to color the vertices of a graph}.
\bjtitle{Commun. ACM}
\bvolume{22}(\bissue{4}),
\bfpage{251}--\blpage{256}
(\byear{1979})
\doiurl{10.1145/359094.359101}
\end{barticle}
\endbibitem

\bibitem[\protect\citeauthoryear{Hoos and Stützle}{2000}]{SATLIB}
\begin{bbook}
\bauthor{\bsnm{Hoos}, \binits{H.}},
\bauthor{\bsnm{Stützle}, \binits{T.}}:
\bbtitle{SATLIB: An online resource for research on SAT},
pp. \bfpage{283}--\blpage{292}
(\byear{2000})
\end{bbook}
\endbibitem

\bibitem[\protect\citeauthoryear{Heule et~al.}{2019}]{SAT_competition_2018}
\begin{barticle}
\bauthor{\bsnm{Heule}, \binits{M.}},
\bauthor{\bsnm{Järvisalo}, \binits{M.}},
\bauthor{\bsnm{Suda}, \binits{M.}}:
\batitle{Sat competition 2018}.
\bjtitle{Journal on Satisfiability, Boolean Modeling and Computation}
\bvolume{11},
\bfpage{133}--\blpage{154}
(\byear{2019})
\doiurl{10.3233/SAT190120}
\end{barticle}
\endbibitem

\bibitem[\protect\citeauthoryear{Cheeseman et~al.}{1991}]{hard_problem_phase_transition}
\begin{bchapter}
\bauthor{\bsnm{Cheeseman}, \binits{P.}},
\bauthor{\bsnm{Kanefsky}, \binits{B.}},
\bauthor{\bsnm{Taylor}, \binits{W.M.}}:
\bctitle{Where the really hard problems are}.
In: \bbtitle{Proceedings of the 12th International Joint Conference on Artificial Intelligence - Volume 1}.
\bsertitle{IJCAI'91},
pp. \bfpage{331}--\blpage{337}.
\bpublisher{Morgan Kaufmann Publishers Inc.},
\blocation{San Francisco, CA, USA}
(\byear{1991})
\end{bchapter}
\endbibitem

\bibitem[\protect\citeauthoryear{Prestwich and Lynce}{2006}]{local_search_for_unsat}
\begin{bchapter}
\bauthor{\bsnm{Prestwich}, \binits{S.}},
\bauthor{\bsnm{Lynce}, \binits{I.}}:
\bctitle{Local search for unsatisfiability},
vol. \bseriesno{4121},
pp. \bfpage{283}--\blpage{296}
(\byear{2006}).
\doiurl{10.1007/11814948_28}
\end{bchapter}
\endbibitem

\bibitem[\protect\citeauthoryear{Pierog et~al.}{2015}]{Pierog2015pseudoEnergy}
\begin{barticle}
\bauthor{\bsnm{Pierog}, \binits{T.}},
\bauthor{\bsnm{Karpenko}, \binits{I.}},
\bauthor{\bsnm{Katzy}, \binits{J.M.}},
\bauthor{\bsnm{Yatsenko}, \binits{E.}},
\bauthor{\bsnm{Werner}, \binits{K.}}:
\batitle{Epos lhc: Test of collective hadronization with data measured at the cern large hadron collider}.
\bjtitle{Phys. Rev. C}
\bvolume{92},
\bfpage{034906}
(\byear{2015})
\doiurl{10.1103/PhysRevC.92.034906}
\end{barticle}
\endbibitem

\bibitem[\protect\citeauthoryear{Mehta et~al.}{2022}]{Mehta2022FNDAM}
\begin{botherref}
\oauthor{\bsnm{Mehta}, \binits{D.}},
\oauthor{\bsnm{Rahman}, \binits{M.}},
\oauthor{\bsnm{Aono}, \binits{K.}},
\oauthor{\bsnm{Chakrabartty}, \binits{S.}}:
An adaptive synaptic array using fowler–nordheim dynamic analog memory.
Nature Communications
\textbf{13}(1)
(2022)
\doiurl{10.1038/s41467-022-29320-6}
\end{botherref}
\endbibitem

\bibitem[\protect\citeauthoryear{Mehta~D.}{2020}]{Mehta2020FNlog}
\begin{botherref}
\oauthor{\bsnm{Mehta~D.}, \binits{C.S.} \bsuffix{Aono~A.}}:
A self-powered analog sensor-data-logging device based on fowler-nordheim dynamical systems.
Nature Communications
\textbf{11}(5446)
(2020)
\doiurl{10.1038/s41467-020-19292-w}
\end{botherref}
\endbibitem

\bibitem[\protect\citeauthoryear{Chen et~al.}{2023}]{dendrite_in_3D}
\begin{bchapter}
\bauthor{\bsnm{Chen}, \binits{H.J.-Y.}},
\bauthor{\bsnm{Beauchamp}, \binits{M.}},
\bauthor{\bsnm{Toprasertpong}, \binits{K.}},
\bauthor{\bsnm{Huang}, \binits{F.}},
\bauthor{\bsnm{Le~Coeur}, \binits{L.}},
\bauthor{\bsnm{Nemec}, \binits{T.}},
\bauthor{\bsnm{Wong}, \binits{H.-S.P.}},
\bauthor{\bsnm{Boahen}, \binits{K.}}:
\bctitle{Multi-gate fefet discriminates spatiotemporal pulse sequences for dendrocentric learning}.
In: \bbtitle{2023 International Electron Devices Meeting (IEDM)},
pp. \bfpage{1}--\blpage{4}
(\byear{2023}).
\doiurl{10.1109/IEDM45741.2023.10413707}
\end{bchapter}
\endbibitem

\bibitem[\protect\citeauthoryear{Boahen}{2022}]{Boahen2022}
\begin{barticle}
\bauthor{\bsnm{Boahen}, \binits{K.}}:
\batitle{Dendrocentric learning for synthetic intelligence}.
\bjtitle{Nature}
\bvolume{612}(\bissue{7938}),
\bfpage{43}--\blpage{50}
(\byear{2022})
\doiurl{10.1038/s41586-022-05340-6}
\end{barticle}
\endbibitem

\bibitem[\protect\citeauthoryear{Coja-Oghlan and Panagiotou}{2016}]{Phase_transition_kSAT}
\begin{barticle}
\bauthor{\bsnm{Coja-Oghlan}, \binits{A.}},
\bauthor{\bsnm{Panagiotou}, \binits{K.}}:
\batitle{The asymptotic k-sat threshold}.
\bjtitle{Advances in Mathematics}
\bvolume{288},
\bfpage{985}--\blpage{1068}
(\byear{2016})
\doiurl{10.1016/j.aim.2015.11.007}
\end{barticle}
\endbibitem

\bibitem[\protect\citeauthoryear{Chancellor et~al.}{2016}]{Chancellor2016}
\begin{barticle}
\bauthor{\bsnm{Chancellor}, \binits{N.}},
\bauthor{\bsnm{Zohren}, \binits{S.}},
\bauthor{\bsnm{Warburton}, \binits{P.A.}},
\bauthor{\bsnm{Benjamin}, \binits{S.C.}},
\bauthor{\bsnm{Roberts}, \binits{S.}}:
\batitle{A direct mapping of max k-sat and high order parity checks to a chimera graph}.
\bjtitle{Scientific Reports}
\bvolume{6}(\bissue{1}),
\bfpage{37107}
(\byear{2016})
\doiurl{10.1038/srep37107}
\end{barticle}
\endbibitem

\bibitem[\protect\citeauthoryear{Kirkpatrick et~al.}{1983}]{Kirkpatrick1983}
\begin{barticle}
\bauthor{\bsnm{Kirkpatrick}, \binits{S.}},
\bauthor{\bsnm{Gelatt}, \binits{C.D.}},
\bauthor{\bsnm{Vecchi}, \binits{M.P.}}:
\batitle{Optimization by simulated annealing}.
\bjtitle{Science}
\bvolume{220}(\bissue{4598}),
\bfpage{671}--\blpage{680}
(\byear{1983})
\doiurl{10.1126/science.220.4598.671}
\end{barticle}
\endbibitem

\bibitem[\protect\citeauthoryear{Aramon et~al.}{2019}]{DAU}
\begin{botherref}
\oauthor{\bsnm{Aramon}, \binits{M.}},
\oauthor{\bsnm{Rosenberg}, \binits{G.}},
\oauthor{\bsnm{Valiante}, \binits{E.}},
\oauthor{\bsnm{Miyazawa}, \binits{T.}},
\oauthor{\bsnm{Tamura}, \binits{H.}},
\oauthor{\bsnm{Katzgraber}, \binits{H.G.}}:
Physics-inspired optimization for quadratic unconstrained problems using a digital annealer.
Frontiers in Physics
\textbf{Volume 7 - 2019}
(2019)
\doiurl{10.3389/fphy.2019.00048}
\end{botherref}
\endbibitem

\bibitem[\protect\citeauthoryear{Hen}{2019}]{Equation_planting}
\begin{barticle}
\bauthor{\bsnm{Hen}, \binits{I.}}:
\batitle{Equation planting: A tool for benchmarking ising machines}.
\bjtitle{Phys. Rev. Appl.}
\bvolume{12},
\bfpage{011003}
(\byear{2019})
\doiurl{10.1103/PhysRevApplied.12.011003}
\end{barticle}
\endbibitem

\end{thebibliography}

\newpage
\setcounter{page}{1}
\setcounter{section}{0}
\setcounter{equation}{0}
\renewcommand{\thesection}{S\arabic{section}}

\makeatletter
\renewcommand\theHsection{supp.\thesection}
\makeatother
\setcounter{section}{0}
\renewcommand{\thesection}{S\arabic{section}}
\setcounter{figure}{0}
\renewcommand{\thefigure}{S\arabic{figure}}
\setcounter{table}{0}
\renewcommand{\thetable}{S\arabic{table}}

\makeatletter
\renewcommand\theHsection{supp.\thesection}
\renewcommand\theHfigure{supp.\thefigure}
\makeatother
\resetlinenumber[1]
\noindent

\section*{Supplementary Information for Higher-Order Neuromorphic Ising Machines - Autoencoders and Fowler-Nordheim Annealers are all you need for Scalability}\label{sec_supp}

\section{Modeling Toggle Neurons as ON-OFF Neuron}\label{sec_supp_ON_OFF_neuron}
We note the update rule for the clause output based on the spike events in the latent neurons from Eq.~\eqref{eq_methods_firing_criteria} as,
\begin{equation}
    T_{k,n} \;=\;
    \begin{cases}
        -\,T_{k,n-1}, & \sigma_{k,n}\neq 0,\\
        \;\;T_{k,n-1}, & \sigma_{k,n}=0.
    \end{cases}
    \label{eq_supp_firing_criteria}
\end{equation}
Here, ${T_{k,n}}$ is the clause output of $k$'th clause at time instant $n$. Let, the change in ${T_{k,n}}$ be denoted as $\Delta{T}_{k,n}$, then
\begin{equation}
     T_{k,n} = {T}_{k,n-1}+2\Delta{T}_{k,n},
    \label{eq_supp_T_using_del}
\end{equation}
where $\Delta{T}_{k,n}=\{-1,0,+1\}$ and $\Delta T_{k,n}T_{k,n-1}=\{-1,0\},  \forall k=1...M$ ensures that the clause outputs either flip or remain unchanged.

\noindent
Now, Eq.~\eqref{eq_supp_firing_criteria} can be written case-by-case as 
\begin{equation}
    \Delta T_{k,n}=
    \left\{ 
      \begin{array}{ c l }
        +1, & \quad \textrm{if }T_{k,n-1}=-1 \textrm{  and  }\sigma_{k,n}\neq 0,\\
        -1, & \quad \textrm{if }T_{k,n-1}=+1 \textrm{  and  }\sigma_{k,n}\neq 0,\\
         0, & \quad \textrm{otherwise}.
      \end{array}
    \right.\label{eq_supp_toggle_case}
\end{equation}
We can now decompose the variables differentially as $\Delta T_{k,n}=\Delta T_{k,n}^+-\Delta T_{k,n}^-$, $T_{k,n}=T_{k,n}^+-T_{k,n}^-$, $\Delta T_{k,n}^+$, $\Delta T_{k,n}^-$, $T_{k,n}^+$, $ T_{k,n}^- \ge0$ leads to $T_{k,n}^+=\sum_{j=1}^n\left[\Delta T_{k,j}^+ - \Delta T_{k,j}^-\right]$, $T_{k,n}^-=\sum_{j=1}^n\left[-\Delta T_{k,j}^+ + \Delta T_{k,j}^-\right]$. Eq.~\eqref{eq_supp_toggle_case} is therefore equivalent to
\begin{equation}
    \Delta T_{k,n}^+=
    \left\{ 
      \begin{array}{ c l }
        1, & \quad \textrm{if }T_{k,n-1}^+=0 \textrm{  and  }\sigma_{k,n}\neq 0,\\
        0, & \quad \textrm{otherwise},
      \end{array}
    \right.\label{eq_methods_firing_criteria_on}
\end{equation}
which corresponds to the spiking criterion for an ON neuron and
\begin{equation}
    \Delta T_{k,n}^-=
    \left\{ 
      \begin{array}{ c l }
        1, & \quad \textrm{if }T_{k,n-1}^-=0 \textrm{  and  }\sigma_{k,n}\neq 0,\\
        0, & \quad \textrm{otherwise},
      \end{array}
    \right.\label{eq_methods_firing_criteria_off}
\end{equation}
which corresponds to the spiking criterion for an OFF neuron. We can rewrite the spiking criteria of each latent neurons from Eq.~\eqref{eq_methods_discrete_event_bernoulli_rv} as,
\begin{equation}
    q_{i,n} =
    \begin{cases}
        1, & \text{if } \displaystyle-2\sum_{j=1}^{n-1}\displaystyle\sum_{k=1}^M \widetilde{H}_{k,i}\left(\Delta T_{k,j}^+ - \Delta T_{k,j}^-\right)>\mu_{i,n}, \\
        0, & \text{otherwise}.
    \end{cases}
\end{equation}

\section{The 3R‐3X Problem and Equation Planting}\label{sec_supp_3R3X_Ising}

The 3R‐3X problem (3‐Regular 3‐XORSAT) is a canonical example of an equation‐planting instance used to benchmark Ising‐machine architectures. In this formulation, one constructs a system of \(N\) Boolean equations over \(N\) binary variables \(\{x_{1},x_{2},\dots,x_{N}\}\), each equation involving exactly three distinct variables, and each variable appearing in exactly three distinct equations.  Because of this regularity, 3R‐3X instances possess known ground‐state solutions (hence “planted”) - making them solvable by Gaussian elimination in $\mathcal{O}(n^3)$ complexity. Yet they exhibit the characteristic NP‐hard energy landscape (egg-carton-like)—large Hamming distances between solutions and nearby excited states—making them ideal for stress‐testing Ising solvers~\cite{Equation_planting}.

To, get the problem shown in Fig.~\ref{fig1}.e, we do uniform random generation of such instance where Boolean (mod‐2) form of the \(k\)th equation can be written as
\begin{equation}
  \sum_{i=1}^N H_{k,i}\,x_i \;=\; b_k \quad (\bmod\,2).
  \label{eq:binary_mod2_form}
\end{equation}

Where $\mathbf{H} = \bigl(H_{k,i}\bigr)\in\{0,1\}^{\,M\times N}$ is the same interconnection matrix we use in our higher-order Ising machine (here \(M=N\) in the 3R‐3X setting). As it can be seen in Fig.~\ref{fig1}.e, exactly three entries of each row of \(\mathbf{H}\) are equal to 1 (ensuring each equation contains three variables), and each column of \(\mathbf{H}\) has exactly three entries equal to 1 (ensuring each variable appears in three equations).

To translate the planted‐equation system~\eqref{eq:binary_mod2_form} into an optimization (Ising‐type) cost function, one introduces the binary‐to‐spin mapping
\[
  x_i \;\longmapsto\; s_i \;=\; (-1)^{\,x_i}
  \quad,\qquad
  s_i \in \{\,-1,\,+1\,\}.
\]
So, the LHS in Eq.~\eqref{eq:binary_mod2_form} can be redefined as the clause variable
\begin{equation}
  T_k \;=\; \prod_{i:\,H_{k,i}=1} s_{i}\;,
  \label{eq:Ti_definition}
\end{equation}

so that the Eq.~\eqref{eq:binary_mod2_form} will now look like
\[
  T_k \;=\; (-1)^{\,b_k}
  \quad\Longleftrightarrow\quad
  \prod_{i:\,H_{k,i}=1} s_{i} \;=\; (-1)^{\,b_k}\,.
\]
Thus, we need to minimize the cost (Hamiltonian) function of the form:
\begin{align}
 & \min \sum_{k=1}^M \Bigl(T_k \;-\; (-1)^{b_k}\Bigr)^2
  \label{eq:ising_cost_raw} \\
  &= \min\sum_{k=1}^M \Bigl[\;T_k^2 \;+\;\bigl((-1)^{b_k}\bigr)^2 \;-\; 2\,(-1)^{b_k}\,T_k\Bigr].
  \nonumber
\end{align}
Since \(T_{k}^2 = 1\) and \(\bigl((-1)^{b_k}\bigr)^2 = 1\), it is therefore equivalent to maximizing
\begin{equation}
  \max\sum_{k=1}^M (-1)^{b_k}\,T_k 
  \;=\; 
  \max \sum_{k=1}^M (-1)^{b_k}\,\prod_{i} s_i^{H_{k,i}},
  \label{eq:ising_maximization}
\end{equation}
which reaches the value \(M\) if and only if all clauses \(i\) satisfy \(T_{i}=(-1)^{b_i}\). One can readily observe that Eq.~\eqref{eq:ising_maximization} directly resembles the general higher-order N-spin Hamiltonian (Eq.~\eqref{eq:method_re_index}) we used to derive our higher-order Ising model - where the weight vector $J_k$ is given by $(-1)^{b_k}$.

For the analysis in Fig.~\ref{fig1}\,(f-g), we employed the quadratization gadget from Chancellor et al.~\cite{Chancellor2016} to map the 3R-3X problem onto a second‐order Ising solver.  After applying this gadget to each third‐order term in the original Ising formulation (Eq.~\eqref{eq:ising_maximization}), the objective simplifies to
\begin{align}
  \max \sum_{k=1}^M (-1)^{b_k}\, \Bigl[\,\bigl(s_{k,1}+s_{k,2}+s_{k,3}\bigr)
    &- \bigl(s_{k,1}s_{k,2} + s_{k,2}s_{k,3} + s_{k,3}s_{k,1}\bigr) \nonumber\\
    &\quad -2\,s_{k,a} + 2\,s_{k,a}\bigl(s_{k,1} + s_{k,2} + s_{k,3}\bigr)\Bigr].
  \label{eq:quadratized_3r3x}
\end{align}
Here, \(s_{k,1}\), \(s_{k,2}\), and \(s_{k,3}\) are the three spins in the \(k\)-th clause, and each third‐order clause introduces a single auxiliary spin \(s_{k,a}\). Thus, by aggregating all like-spin tuples, we obtain the final second-order Ising formulation, which directly yields the quadratic Hamiltonian \(\mathbf{Q}\) used by the second-order solver.

\section{MAX-CUT to Ising}\label{sec_supp_MAXCUT_Ising}
 
The MAXCUT problem on an undirected graph \(G = (V,E)\) with nonnegative edge weights \(w_{ij}\) (where \(|V| = N\) and \(|E| = M\)) seeks a partition of \(V\) into two disjoint subsets that maximizes the total weight of the cut edges.  Equivalently, by assigning each vertex \(i\) a spin \(s_i\in\{\pm1\}\), an edge \(\{i,j\}\) contributes only if \(s_i\neq s_j\).  Hence the problem can be written as
\[
\max_{s\in\{\pm1\}^N}
\sum_{\{i,j\}\in E}
w_{ij}\,\frac{1 - s_i s_j}{2}.
\]
\noindent
As in Section~\ref{sec_asynchronous_ising_model}, define a binary matrix \(H\in\{0,1\}^{M\times N}\) by
\[
H_{k,i} = 
\begin{cases}
1,&\text{if vertex }i\text{ is incident to edge }e_k,\\
0,&\text{otherwise}.
\end{cases}
\]
Then each edge \(e_k = \{i,j\}\) yields the clause–monomial
\[
T_k(s) \;=\; \prod_{n=1}^N s_n^{H_{k,n}} \;=\; s_i\,s_j.
\]
In these terms, MAXCUT becomes
\[
\max_{s\in\{\pm1\}^N}
\sum_{k=1}^M
w_{e_k}\,\frac{1 - \displaystyle\prod_{n=1}^N s_n^{H_{k,n}}}{2}.
\]
Dropping the constant \(\tfrac12\sum_k w_{e_k}\) and setting \(J_{e_k} = w_{e_k}/2\) gives the equivalent Ising Hamiltonian
\[
\max_{s\in\{\pm1\}^N}
E_{\mathrm{MAXCUT}}(s)
\;=\;
-\sum_{k=1}^M
J_{e_k}\,\prod_{n=1}^N s_n^{H_{k,n}}.
\]
\\

\section{MAXSAT to Ising}\label{sec_supp_MAXSAT_Ising}

A CNF (conjunctive normal form) formula over Boolean variables \(b_1,\dots,b_N\) is a conjunction of \(M\) clauses—each clause being a disjunction of literals:
\[
\Phi(b)\;=\;\bigwedge_{k=1}^M  
\bigl(\ell_{k,1}\vee \ell_{k,2}\vee \cdots \vee \ell_{k,p}\bigr),
\]
where each literal \(\ell_{k,i}\) is either \(b_i\) or \(\neg b_i\).  In MAXSAT, which is the optimization version of the decision problem boolean SAT, we seek an assignment \(b_i\in\{0,1\}\) that maximizes the number of satisfied clauses.

For the \(k\)th clause with \(p\) literals \(\ell_{k,1},\dots,\ell_{k,p}\), its Boolean indicator is
\[
C_k(b)
\;=\;
1 \;-\;\prod_{i=1}^p\bigl(1-\ell_{k,i}\bigr),
\]
since \(\ell_{k,i}=1\) exactly when that literal is true.  Mapping each literal to a spin  
\[
s_{k,i}\in\{\pm1\},  
\qquad
\ell_{k,i}=\tfrac{1+s_{k,i}}{2},
\]
gives \(1-\ell_{k,i}=(1-s_{k,i})/2\) and hence
\[
C_k(s)
=1 \;-\;\frac{1}{2^p}\,\prod_{i=1}^p(1-s_{k,i})
=1 \;-\;\frac{1}{2^p}
\sum_{\ell=0}^p(-1)^{\ell}
\sum_{1\le i_1<\cdots<i_\ell\le p}
s_{k,i_1}\cdots s_{k,i_\ell}.
\]
Up to the factor \(2^{-p}\) and an additive constant, each clause contributes
\[
\Phi_k(s)
=\sum_{\ell=1}^p(-1)^{\ell-1}
\sum_{1\le i_1<\cdots<i_\ell\le p}
s_{k,i_1}s_{k,i_2}\cdots s_{k,i_\ell},
\]
which equals \(+1\) if the clause is satisfied and \(-(2^p-1)\) otherwise.  The MAXSAT objective is then
\[
\max_{s_{k,i}\,\in\{\pm1\}}
\sum_{k=1}^M\Phi_k(s).
\]

\bigskip
\noindent\textbf{Examples.}
\begin{itemize}
  \item \(p=3\):  
  \[
    \Phi_k(s_{k,1},s_{k,2},s_{k,3})
    = (s_{k,1}+s_{k,2}+s_{k,3})
      - (s_{k,1}s_{k,2}+s_{k,2}s_{k,3}+s_{k,3}s_{k,1})
      + s_{k,1}s_{k,2}s_{k,3}.
  \]
  \item \(p=5\):  
  \[
    \begin{aligned}
      \Phi_k(s_{k,1},\dots,s_{k,5})
      &= \sum_{i=1}^5 s_{k,i}
        - \sum_{1\le i<j\le5}s_{k,i}s_{k,j}
        + \sum_{1\le i<j<r\le5}s_{k,i}s_{k,j}s_{k,r}\\
      &\quad
        - \sum_{1\le i<j<r<\ell\le5}s_{k,i}s_{k,j}s_{k,r}s_{k,\ell}
        + s_{k,1}s_{k,2}s_{k,3}s_{k,4}s_{k,5}.
    \end{aligned}
  \]
\end{itemize}

\noindent
\\Now, to construct the interconnection matrix \(\widetilde{\mathbf{H}}\) required by the algorithm (discussed in Methods Section~\ref{sec_autoencoder_model}), we follow the three steps.

1. Literal-to‐spin mapping:  
   For, each clause \(\Phi_k\), we substitute the spin variables as follows, 
   \begin{align}
   s_{k,i}=
       \begin{cases}
       +\,s_i, & \ell_{k,i}=b_i,\\
       -\,s_i, & \ell_{k,i}=\neg b_i.
     \end{cases}
   \end{align}
     
2. Expansion into a higher–order Ising polynomial:  
   Substituting these spin variables into
   \(\sum_{k=1}^M\Phi_k(s)\) and collecting like terms yields
   \[
     E(\mathbf{s})
     = -\sum_{k=1}^{p}
       \sum_{1 \,\le\, i_1 < \cdots < i_k \,\le\, N}
       J^{(k)}_{\,i_1 i_2 \cdots i_k}
       \;\prod_{m=1}^{k} s_{i_m}\,,
   \]
   where each coefficient \(J^{(k)}_{i_1\cdots i_k}\) arises from collecting multiple spin tuples of the same indices.

3. Assembling \(\widetilde{\mathbf{H}}\):  
   Finally, we follow the procedure of Section~\ref{sec_asynchronous_ising_model} to form the binary matrix \(\widetilde{\mathbf{H}}\).  Each nonzero interaction term \(\;J^{(k)}_{i_1\cdots i_k}\prod_{m=1}^{k} s_{i_m}\)\; corresponds to a row of \(\widetilde{\mathbf{H}}\) with ones in columns \(i_1,\dots,i_k\) (and zeros elsewhere), carrying the weight \(J^{(k)}_{i_1\cdots i_k}\).

\section{Quadratization of MAX-3SAT}\label{sec_supp_quadratization_of_3SAT}
We begin by recalling the cubic Ising formulation for unweighted Max‐3‐SAT (cf.\ the previous section).  Let each clause \(k\) involve spin literals \(s_{k,1},s_{k,2},s_{k,3}\in\{\pm1\}\).  Then
\begin{align}
\max_{s_{k,i}\in\{\pm1\}}
\sum_{k=1}^M
\Phi_k(s)
=& (s_{k,1}+s_{k,2}+s_{k,3})
- \bigl(s_{k,1}s_{k,2}+s_{k,2}s_{k,3}+s_{k,3}s_{k,1}\bigr)
+ s_{k,1}s_{k,2}s_{k,3}.\nonumber
\end{align}
We now introduce one auxiliary spin \(s_{a_k}\) per clause to reduce each cubic term to quadratic form.  After dropping the common additive constant and overall scale, each clause can be written as
\begin{align}\label{eqn_supp_max3sat_quad}
\max_{s_{k,i}\in\{\pm1\}}
\sum_{k=1}^M\Phi_k(s)
&= (s_{a_k}+1)\,(s_{k,1}+s_{k,2}+s_{k,3})\\
&\quad -\;(s_{k,1}s_{k,2}+s_{k,2}s_{k,3}+s_{k,3}s_{k,1})
\;-\;s_{a_k}.\nonumber
\end{align}

This form is obtained by replacing the original cubic term in the boolean-literal-based optimization obejctive polynomial with the standard gadget given in \cite{Freedman2005},
\[
\ell_{k1}\,\ell_{k2}\,\ell_{k3}
\;=\;
\max_{b_{a_k}\in\{0,1\}}
\bigl\{\,b_{a_k}\,\bigl(\ell_{k1}+\ell_{k2}+\ell_{k3}-2\bigr)\bigr\},
\]
where \(b_{a_k}\) is the auxiliary Boolean variable for clause \(k\).  Since this gadget and the original cubic literal term attain the same maximum value for the same variable assignments, any assignment that maximizes \(\Phi_k\) in \eqref{eqn_supp_max3sat_quad} also maximizes the original clause, and vice versa.

Finally, to build the quadratic Hamiltonian matrix \(\mathbf{Q}\) and bias vector \(\mathbf{h}\) used by algorithms such as NeuroSA~\cite{Chen2025}, we do the following:

1. Literal‐to‐spin mapping:  Replace each clause‐local spin \(s_{k,i}\) by the global spin \(s_i\) via
   \[
     s_{k,i} = 
     \begin{cases}
       +\,s_i, & \text{if \quad}\ell_{k,i} = b_i,\\
       -\,s_i, & \text{if \quad}\ell_{k,i} = \neg b_i,
     \end{cases}
     \quad
     \text{and, \quad}s_{a_k}=s_{M+k}.
   \]

2. Collect and assemble:  Substitute into \eqref{eqn_supp_max3sat_quad}, expand, and group like terms to obtain
\[
H(\mathbf{s})
=\sum_{1\le i<j\le M+N}Q_{ij}\,s_i\,s_j
\;+\;\sum_{i=1}^{M+N}h_i\,s_i,
\]
so that the final optimization reads
\[
\max_{\mathbf{s}\in\{\pm1\}^{M+N}}
H(\mathbf{s})
\;=\;
\tfrac12\,\mathbf{s}^TQ\,\mathbf{s}
\;+\;\mathbf{h}^T\mathbf{s}.
\]

\section{Ratio of the size of Interconnection matrix with problem size} \label{sec_supp_resource_vs_size}
\begin{figure}[ht!]
\centering
\includegraphics[width=0.5\textwidth]{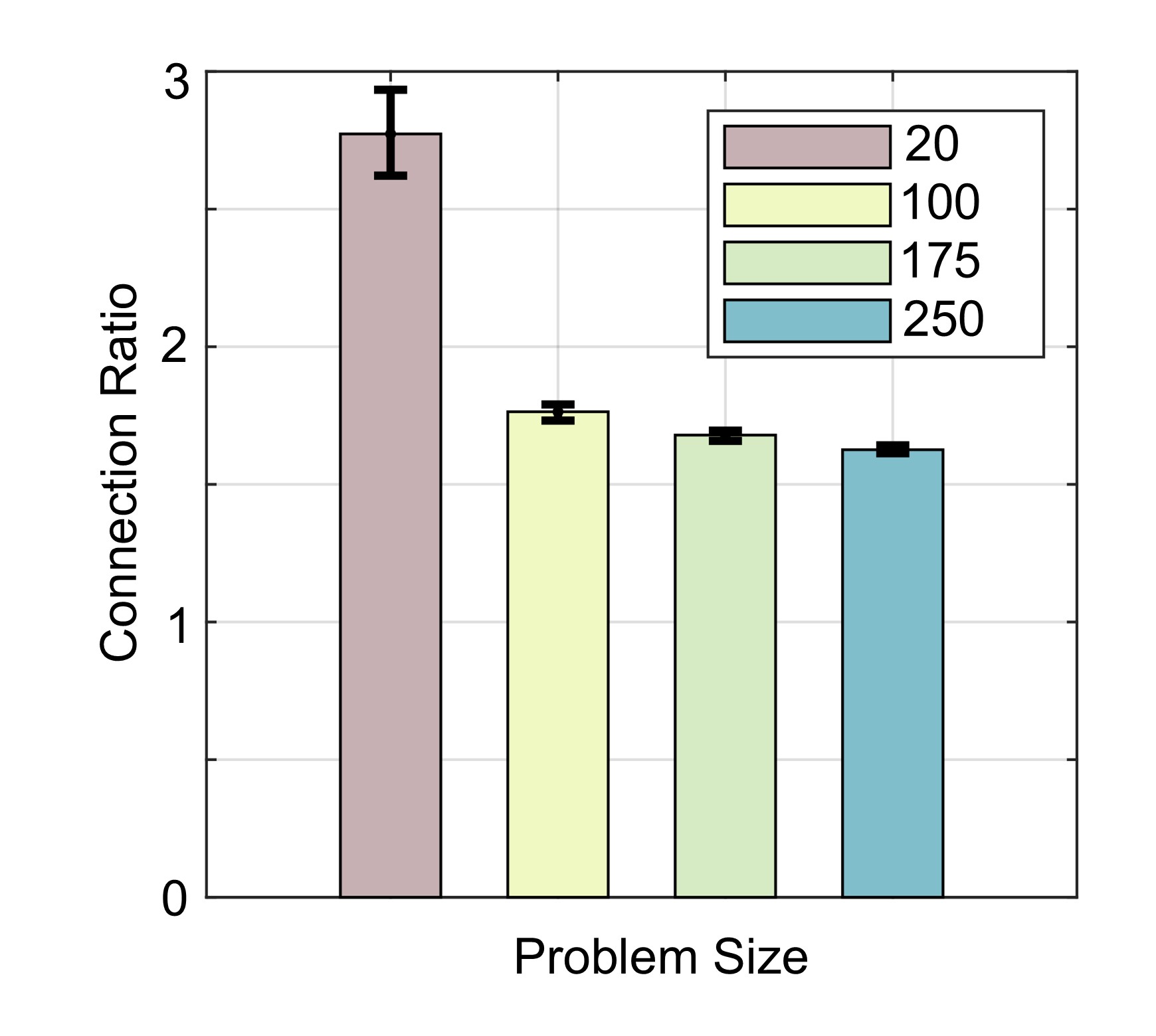}
\caption{{This barplot shows the ratio of connections for our higher-order implementation of MAX-3SAT to the QUBO implementation for different problem sizes. 
} 
}\label{supplementary_fig1}
\end{figure}

Following the discussion in the previous section, for a given MAX-3SAT problem, quadratizing each clause with 3 variables introduces 1 auxiliary variable. Consequently, for a MAX-3SAT problem with $M$ variables and $N$ clauses, the quadratic Hamiltonian matrix $\mathbf{Q}$ will have dimensions $(M+N) \times (M+N)$. In contrast, for our higher-order implementation, we first determine the interconnection matrix $\mathbf{H}$ as described in section~\ref{sec_supp_MAXSAT_Ising}. 

In Fig.~\ref{supplementary_fig1}, we plot the ratio of the size of $\mathbf{Q}$, which represents the connection matrix of the quadratized problem, to the size of $\mathbf{H}$, the connection matrix for the higher-order implementation, for various problem sizes (number of variables: 20, 100, 175, 250) using satisfiable problems from the SATLIB benchmark~\cite{SATLIB}. It should be noted that for a given problem with a fixed number of variables, the size of $\mathbf{H}$ varies depending on the specific problem instance. Therefore, in Fig.~\ref{supplementary_fig1}, we use 16 different instances for each problem size to derive statistics for $\mathbf{H}$, as indicated by the error bars in the figure. One can observe a trend where the ratio of the interconnection matrix gradually stabilizes towards a constant value as the problem size increases. 

Consequently, in Fig.~\ref{fig1}.b, we employ the interconnection ratio for the largest problem size to generate the plot of the interconnection ratio against the interaction order of the original objective polynomial.

\begin{algorithm}
\caption{Pseudocode for the implementation without graph coloring}\label{alg:pseudocode_without_color}
\begin{algorithmic}[1]
  \State \textbf{Input:} $\mathbf{H}\in\{0,1\}^{M\times N}$, $\mathbf{J}\in\mathbb{R}^M$, max iterations $\texttt{MAX\_ITER}$
  \State $\widetilde{\mathbf{H}} \; \gets\;\mathrm{diag}(\mathbf{J})\,\mathbf{H}\,$
  \State Initialize parameters $t$, $\Delta t$, $\alpha$, $\beta$, $B$, $\epsilon$
  \State Draw initial variables $\mathbf{X}\sim\mathrm{Uniform}\{0,1\}^N$
  \State Compute clause‐outputs $\mathbf{T}\gets\mathrm{calculate\_clause\_outputs}\bigl(\mathbf{H},\,2\mathbf{X}-1\bigr)$
  \Statex \Comment{$\mathbf{T} \in \{-1,+1\}^{M}$}
  \State $iter\gets 1$
  \While{$iter \le \texttt{MAX\_ITER}$}
    \State Sample uniforms $\mathbf{u}\sim U(0,1)^N$
    \State $\boldsymbol{\mu}\gets 
      \dfrac{\beta\,\log\!\bigl(B\,\mathbf{u}+\epsilon\bigr)}
            {\log\!\bigl(1 + \alpha\,t\bigr)}$
    \Comment{Threshold for latent neurons}
    \State $\mathbf{q}\gets \widetilde{\mathbf{H}}^{\mathsf{T}}\,\mathbf{T}$
    \Comment{Aggregate inputs to latent neuron}
    \State $\texttt{mask}\gets (\mathbf{q} < \boldsymbol{\mu})$
    \Comment{Spiking criteria}
    \If{$\exists\,j:\texttt{mask}_j$}
      \State Select $v\sim\mathrm{Uniform}\{\,j:\texttt{mask}_j\}$
      \Comment{Rejection‐free sampling}
      \State $\boldsymbol{\sigma}\gets \mathbf{H}(:,v)
$
      \Comment{Clauses containing spin $v$}
      \ForAll{$j$ with $\sigma_j\neq 0$}
        \State $T_j\gets -\,T_j$  
      \EndFor
      \Comment{Flip affected clause‐outputs}
    \EndIf
    \State $iter\gets iter + 1$, \quad $t\gets t + \Delta t$
  \EndWhile
  \State \textbf{Output:} Final clause‐outputs $\mathbf{T}$
\end{algorithmic}
\end{algorithm}

\section{Algorithmic Implementation}\label{sec_supp_algorithmic_implementation}

We consider two variants of our higher‐order Ising machine: one using graph coloring and one without. In the colored version, events on conditionally independent latent neurons of the same color can be processed simultaneously, whereas in the uncolored version a global arbiter selects which events to accept to preserve the ergodicity of the simulated annealing process. So in both cases, we only employ the annealed exponential random variable \(\mathcal{N}^E_n\) as the threshold \(\mu_n\) for each latent neurons, as detailed in Methods Section~\ref{sec_methods_sync_impl}.

For the implementation without coloring, we adopt a rejection‐free sampling strategy—common in solvers such as DAU~\cite{DAU}—in which all latent neurons are tested against their thresholds in parallel. And then, we choose one of the latent neurons uniformly at random only from the list of neurons that will potentially give an event. This “parallel‐trial” scheme enhances the acceptance probability, since the chance of at least one valid flip among \(N\) trials typically exceeds that of a single‐trial update. The detailed pseudocode for this variant is provided in Algorithm~\ref{alg:pseudocode_without_color}.

\begin{algorithm}
\caption{Graph‐Colored Higher-Order Ising Machine}\label{alg:graph_colored_algo}
\begin{algorithmic}[1]
  \State \textbf{Input:}
    $\mathbf{H}\in\{0,1\}^{M\times N}$,\quad
    $\mathbf{J}\in\mathbb{R}^M$,\quad
    max iterations \texttt{MAX\_ITER},%
  \Statex \quad\quad\quad\quad color groups $\{\mathbf{G}_{1},\dots,\mathbf{G}_{R}\}$
  \State $\widetilde{\mathbf{H}} \; \gets\;\mathrm{diag}(\mathbf{J})\,\mathbf{H}\,$
  \State Precompute column sum 
    $\;\mathbf{C_{sum}}\gets \widetilde{\mathbf{H}}^{\mathsf T}\,\mathbf{1}^{M}$
  \Comment{where $\mathbf{1}^{M}=(1,\dots,1)^\mathsf T$}
  \State Initialize parameters 
    $t,\;\Delta t,\;\alpha,\;\beta,\;B,\;\varepsilon$
  \State Draw initial variable assignment 
    $\;\mathbf{X}\sim\mathrm{Uniform}\{0,1\}^{N}$
  \State Compute clause‐outputs 
    $\;\mathbf{T}\gets \mathrm{calculate\_clause\_outputs}(\mathbf{H},\,\mathbf{X})$
    \Comment{$\mathbf{T} \in \{0,1\}^{M}$}
  \State $iter\gets 1$
  \While{$iter \le \texttt{MAX\_ITER}$}
    \For{$r = 1$ to $R$}
      \State $\mathbf{V}\gets \mathbf{G}_r,\;m\gets |\mathbf{V}|$
      \Comment{Variables of color $r$}
      \State Sample noise 
        $\;\mathbf{u}\sim U(0,1)^{m}$
      \State
        $          \boldsymbol{\mu}
          \;\gets\;
          \frac{\beta\,\log\bigl(B\,\mathbf{u} + \varepsilon\bigr)}
               {\log\bigl(1 + \alpha\,t\bigr)}
        $
      \Comment{Thresholds for latent neurons of color $r$}
      \State 
        $\;\mathbf{q}_{\mathrm{cal}}
          \;\gets\;
          2\,(\widetilde{\mathbf{H}}^{\mathsf T}\mathbf{T}) - \mathbf{C_{sum}}$
        \Comment{Aggregate inputs to latent neurons}
      \State 
        $\;\mathbf{q}_{\mathbf{V}}
          \;\gets\;
          (\mathbf{q}_{\mathrm{cal}}[\mathbf{V}] < \boldsymbol{\mu})$
      \Comment{Spiking criteria for neurons in $G_r$}
      \State 
        $
          \boldsymbol{\sigma}
          \;\gets\;
          (\mathbf{H}{(:,\mathbf{V})}\,.\,\mathbf{q}_{\mathbf{V}}) \;\neq\; \mathbf{0}
        $
      \Comment{Calculate parity indicator}
      \State 
        $\;\mathbf{T}
          \;\gets\;
          \mathbf{T}\;\oplus\;\boldsymbol{\sigma}$
      \Comment{$\oplus$: Denotes element-wise addition modulo 2}
      \State $t\gets t + \Delta t,\quad iter\gets iter + 1$
    \EndFor
  \EndWhile
  \State \textbf{Output:} Final clause‐outputs $\mathbf{T}$
\end{algorithmic}
\end{algorithm}

The graph‐colored implementation leverages sparsity to perform massively parallel updates on neurons of the same color (see Algorithm~\ref{alg:graph_colored_algo}). In hardware realizations, rather than using bipolar clause outputs \(T_k\in\{-1,+1\}\), we may employ binary outputs in \(\{0,1\}\). Consequently, Eq.~\eqref{eq_methods_discrete_event_bernoulli_rv} can be reformulated as
\begin{equation}
    q_{i,n} =
    \begin{cases}
        1, & \text{if }
            -2\displaystyle\sum_{k=1}^{M} \widetilde{H}_{k,i}\,T_{k,n-1}
            + \displaystyle\sum_{k=1}^{M} \widetilde{H}_{k,i}
            > \mu_{i,n},\\[8pt]
        0, & \text{otherwise},
    \end{cases}
    \label{eq_supp_discrete_event_bernoulli_rv}
\end{equation}
where the second summation $\sum_{k=1}^{M} \widetilde{H}_{k,i}$
is the precomputed sum of the \(i\)th column of \(\widetilde{\mathbf{H}}\) (see line 2 of Algorithm~\ref{alg:graph_colored_algo}).

In either variant, \(\texttt{MAX\_ITER}\) denotes the total number of discrete‐time steps, \(\Delta t\) is the sampling interval for the continuous FN annealer dynamics, and \(\alpha,\beta\) are FN annealer hyperparameters (cf.\ Methods Section~\ref{sec_methods_FN_annealer}). In the colored algorithm, after processing each color group we advance the simulation time \(t \leftarrow t + \Delta t\) and increment the iteration counter, ensuring the same annealing schedule as the variant without coloring when all FN hyperparameters are equal.

\begin{figure}[ht!]
\centering
\includegraphics[width=\textwidth]{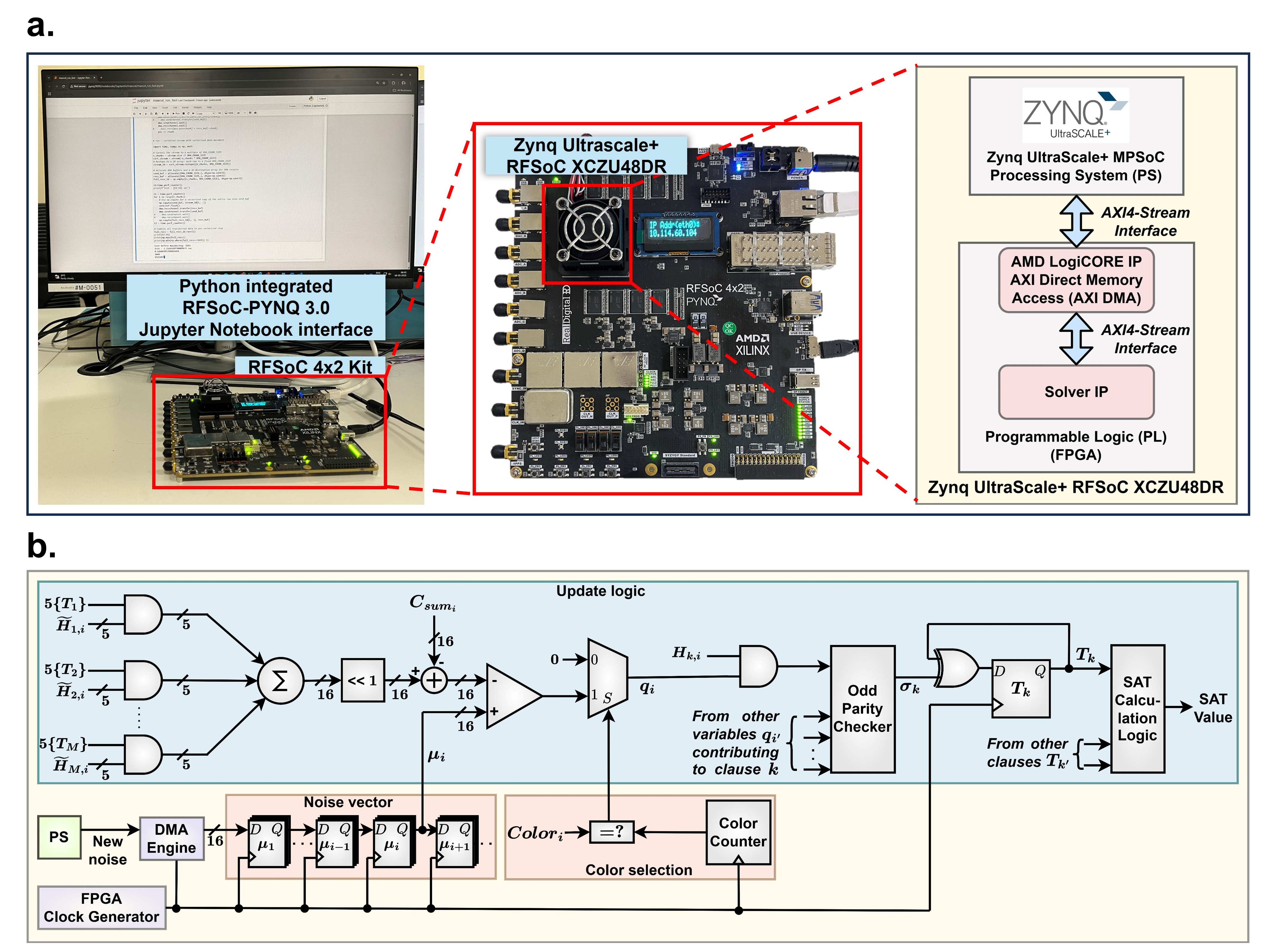}
\caption{\textbf{FPGA-based digital implementation of the algorithm:} (a) The interfacing unit to communicate between the processing system (PS) on board and the programmable logic (PL) of the FPGA. (b) Digital hardware implementation logic of the algorithm.
} 
\label{supplementary_fig2}
\end{figure}
\section{Implementation on FPGA} \label{supp_sec_FPGA}
Here, we describe an FPGA-based implementation of the algorithm using the RFSoC 4x2 FPGA Kit featuring the Zynq Ultrascale+ RFSoC XCZU48DR-2FFVG1517E Core. The basic architecture of the design is shown in Fig.~\ref{supplementary_fig2}.

\noindent
(a) \textbf{Interfacing unit:}
The RFSoC board features the Zynq UltraScale+ RFSoC XCZU48DR device, which integrates a powerful Arm Cortex-A53 64-bit quad-core processing system (PS) alongside user-programmable logic (PL) on the same silicon. We use the on-board PS, programmed via PYNQ’s Python interface, to configure, drive and communicate with our PL design. A custom Python Master module initializes the slave FPGA hardware and streams the configuration bits and generated 16-bit random noise samples into the PL, emulating the Fowler-Nordheim annealer behavior. The communication between the software and hardware layers is carried over a fourth-generation Advanced eXtensible Stream Interface (AXI4-Stream), supporting 32-bit read or write transfers. Inside the FPGA PL fabric, an AMD LogiCORE AXI Direct Memory Access (AXI DMA) IP core handles both-way data communication, enabling seamless writes to and reads from the FPGA without requiring any external components.

\noindent
(b) \textbf{FPGA unit:}
The hardware design is made of two components: a solver core, that executes the Algorithm~\ref{alg:graph_colored_algo} and an AXI4-stream wrapper that interface with the AXI DMA engine.
The solver core maintains a `M' bit register storing the clause output values ($T_1,\dots,T_M$), and `L' parallel 16-bit shift registers that buffer recent noise samples, where `L' equals the maximum number of variables in any color group. Upon startup, the Python Master initiates a DMA transfer to load initial values into both the T registers and noise buffers. When initialization completes, asserting the RUN signal engages the computation loop.
On each clock edge, the noise buffers shift right by one position, and the newest 16‑bit noise sample—generated by the Python Master and provided on the same AXI4 stream through continuous DMA transfer—is inserted into the first buffer. This design minimises PL resource usage by reusing buffered noise samples over successive cycles to compare for different variables without losing the randomness while maintaining a time-varying noise schedule. Concurrently, a color counter increments each clock, wrapping after reaching the predefined total number of colors, thereby selecting the next color group of latent neurons to update. For the selected color, all clauses requiring a flip—computed according to Eq.~\eqref{eq_methods_discrete_event_bernoulli_rv},\eqref{eqn_methods_sigma_cal}, and \eqref{eq_methods_firing_criteria}—are evaluated in parallel via combinational logic, and the resulting flips to the T registers take effect on the subsequent clock edge.
After each noise injection, the DMA receive channel is enabled to capture the 16‑bit SAT result generated by the solver core, streaming these values back to system memory for offline analysis. To detect a successful solution, the design compares each SAT value against the target clause count, `C'; if they match, the SOLVED signal is asserted. At any time, the current state of the clauses can be read out by asserting a READ signal, which will pause computation and latch the current T register values for sequential readout.
The python master includes a programmable counter that begins counting the clock edges the moment RUN is asserted and halts when SOLVED goes high; multiplying this cycle count by the clock period yields the time-to-solution metric. Note that this measurement excludes communication overheads such as DMA interrupt latency and software-based noise generation during initialization.
While performing the experiment of running the hardware for a fixed number of iterations, the entire process is controlled through the Python Master. After initializing the FPGA hardware, the master continuously streams new noise samples into the FPGA via a DMA write buffer. Internally, the FPGA logic is designed to consume one noise sample per clock cycle: the AXI4-Stream interface guarantees that exactly one valid sample is accepted on each iteration or clock cycle. Consequently, the total number of iterations executed on the FPGA is simply equal to the number of noise samples dispatched by the master. As each noise sample is sent, the Python controller’s counter increments; similarly, as each sample is processed, the corresponding SAT value is streamed back over the DMA read channel, providing a direct confirmation of the iteration count on the FPGA side. Once the final noise sample has been transmitted, the AXI wrapper automatically disables the solver core and freezes the contents of all T-registers. If desired, the master may issue a READ command over DMA at any point to halt computations and sequentially read out the T-register values on subsequent clock cycles.
To measure the time required to reach a specified number of satisfied clauses, we load the target SAT count (C) into the Python master at configuration time. On each iteration, the FPGA slave streams back the computed SAT value to the Python master over DMA, which it compares against the preloaded target. When the two values match, the Python master stops its iteration counter at that instant; the recorded count, together with the clock frequency, yields the FPGA’s time-to-solution for the given SAT target. If the initial DMA latency (which is negligible) and software noise generation time (at Pre-RUN or CONFIG stage) are excluded, this calculated TTS almost matches with the actual time elapsed during the RUN phase, measured by the Python master using \texttt{time.perf\_counter\_ns()}. The mismatch is random and in orders of magnitude smaller than the actual TTS.

\begin{table}[htb]
\centering
\scriptsize
\setlength{\tabcolsep}{4pt}      
\renewcommand{\arraystretch}{0.8} 
\caption{Comparison of best cuts and time-to-solution (TTS) on G-set instances.}
\label{tab:gset-cuts}
\begin{tabular}{lclrrrrrrrr}
\toprule
\textbf{Instance} & \textbf{Weight} & \textbf{Type}     & \textbf{SOTA}
  & \multicolumn{2}{c}{\textbf{NeuroSA}}
  & \multicolumn{2}{c}{\textbf{dSBM}}
  & \multicolumn{2}{c}{\textbf{N-HOIM}} \\
\cmidrule(lr){5-6}\cmidrule(lr){7-8}\cmidrule(lr){9-10}
        &        &          &     
  & \textbf{Cuts}     & \textbf{TTS (s)}
  & \textbf{Cuts}     & \textbf{TTS (s)}
  & \textbf{Cuts}     & \textbf{TTS (s)} \\
\midrule
G4   & $+1$             & random   & 11\,646
     & 11\,641  &   540
     & 11\,646  & 0.059
     & 11\,646  & 0.03 \\
G11  & $+1$/\,${-1}$   & toroidal &    564
     &    564   &   525
     &    564   & 0.020
     &    564   & 0.10 \\
G15  & $+1$             & planar   &  3\,050
     &  3\,049  &   480
     &  3\,050  & 1.200
     &  3\,050  & 0.17 \\
\bottomrule
\end{tabular}
\end{table}

\begin{table}[ht]
    \centering
    \caption{Comparison of dSBM vs.\ N-HOIM on G-set.}
    \label{tab:dsbm_vs_nhoim}
    \begin{tabular}{l c c  c c c}
        \toprule
        & \multicolumn{2}{c}{\textbf{dSBM}} 
        & \multicolumn{3}{c}{\textbf{N-HOIM}} \\
        \cmidrule(l){2-3} \cmidrule(l){4-6}
        \textbf{Dataset} 
        & \textbf{$P_S$} 
        & \textbf{TTS (s)} 
        & \textbf{$\Delta$} 
        & \textbf{$P_S$} 
        & \textbf{TTS (s)} \\
        \midrule
        \multirow{3}{*}{G4}
          & \multirow{3}{*}{0.0784} 
          & \multirow{3}{*}{0.059}    
          & $3\times10^{-6}$  & 0.554 & 0.353715 \\
          & 
          &          
          & $3\times10^{-5}$ & 0.254 & 0.107146 \\
          & 
          &          
          & $3\times10^{-4}$ & 0.113 & 0.030091 \\
        \addlinespace
        \multirow{3}{*}{G11}
          & \multirow{3}{*}{0.3304} 
          & \multirow{3}{*}{0.020}    
          & $3\times10^{-6}$ & 0.998 & 0.134839 \\
          & 
          &          
          & $3\times10^{-5}$ & 0.498 & 0.110878 \\
          & 
          &          
          & $3\times10^{-4}$ & 0.075 & 0.100058 \\
        \addlinespace
        \multirow{3}{*}{G15}
          & \multirow{3}{*}{0.0192} 
          & \multirow{3}{*}{1.200}    
          & $3\times10^{-6}$ & 0.067 & 10.47815 \\
          & 
          &          
          & $3\times10^{-5}$ & 0.065 & 1.15033 \\
          & 
          &          
          & $3\times10^{-4}$ & 0.053 & 0.169688 \\
        \bottomrule
    \end{tabular}
\end{table}

\section{Detailed MAX‐CUT Benchmark Results and Annealing Parameter Selection}
\label{supp_sec_compare_Gset}
Table~\ref{tab:gset-cuts} presents the detailed results of our FPGA implementation benchmark for solving MAX‐CUT problems, as illustrated in Fig.~\ref{fig2}.c. The full TTS computation procedure is described in Methods Section~\ref{sec_methods_benchmark_dataset}.  

For these experiments, the FN‐annealer parameters (see Methods Section~\ref{sec_methods_FN_annealer}) were set as follows: the exponential noise mean \(\mathcal{N}_n^E\) was fixed at \(-1\) for all graph instances. The annealing schedule in Eq.~\eqref{eq_methods_optimal_temp2} can be rewritten as
\begin{equation}
    \tau(t) \;=\; \frac{A}{\,C\log\bigl(1 + \frac{t}{C}\bigr)\,},
    \label{eq_supp_optimal_temp2}
\end{equation}
where \(A = \tau_{0} \,C\). For all experiments in this work, we chose \(C = 8\times 10^{4}\). The parameter \(A\) was selected heuristically to be slightly larger than the maximum \(\Delta E_{i}\) observed across all spins over a few iterations. Specifically, for MAX‐CUT benchmarking, we set \(A\) ahead of time as follows: \(A = 30\) for G4, \(A = 20\) for G11, and \(A = 30\) for G15.

The sampling period \(\Delta\) provides a notion of time step analogous to the parameter \(\Delta_{t}\) used in dSBM~\cite{SBM_Toshiba}. Since \(\Delta\) is sensitive to performance, we chose it on a per‐instance basis. In dSBM, each graph is run with five different timesteps, and the best result is selected. Similarly, we ran each graph with three different values of \(\Delta = \{\,3\times10^{-4},\,3\times10^{-5},\,3\times10^{-6}\}\) and reported the results for the best performing \(\Delta\). For each graph instance, we performed 1000 independent trials to estimate the success probability \(P_{S}\) and then plotted the median TTS. Table~\ref{tab:dsbm_vs_nhoim} provides the detailed \(P_{S}\) values for each \(\Delta\), along with the reported numbers from dSBM. Note that the total simulation runtime for each experiment (the time it'll run for if the stopping criteria is not met) follows directly from \(\Delta\), since they are inversely related (a smaller \(\Delta\) leads to a longer simulation time). For each chosen \(\Delta\), we set the simulation length to hundred times the reciprocal of \(\Delta\)'s order of magnitude; for example, if \(\Delta = 3\times10^{-4}\), the simulation runs for \(100\times10^{4}\) iterations.

For the FPGA experiments in Fig.~\ref{fig4}\,(a–c), we selected the annealing parameters in Methods Section~\ref{sec_methods_FN_annealer} using the same procedure described above.

From Table~\ref{tab:dsbm_vs_nhoim}, we observe that for G15, each ten‐fold slowing of the annealing schedule (i.e., ten times longer simulation time) yields a ten‐fold increase in TTS (in seconds), calculated using
\[
  \mathrm{TTS} \;=\; T_{\mathrm{comp}}\;\frac{\log\bigl(1 - 0.99\bigr)}{\log\bigl(1 - P_{S}\bigr)}.
\]
This occurs because, for G15, the success probability \(P_{S}\) does not improve with increased simulation time. In other words, using the formula above alone fails to capture the observation from~\cite{Chen2025} that achieving marginal improvements in solution quality requires exponentially more computation time. Therefore, for all TTS‐related experiments other than MAX‐CUT benchmarking, we use the following procedure: if a run fails to reach the target solution quality, we assign it a large TTS value, then compute the median TTS across all runs and display appropriate error bars to indicate the variance.
\begin{table}[htb]
\centering
\scriptsize
\setlength{\tabcolsep}{6pt}        
\renewcommand{\arraystretch}{0.9}   
\caption{Best‐SAT results on the unsatisfied SATLIB instances.}
\label{tab:supp-satlib-results}
\begin{tabular}{lcr}
\toprule
\textbf{Test set}    & \textbf{Instance} & \textbf{Best SAT} \\
\midrule
\multirow{4}{*}{uuf100-430}   & 1 & 429 \\
                              & 2 & 429 \\
                              & 3 & 429 \\
                              & 4 & 428 \\
\midrule
\multirow{4}{*}{uuf125-538}   & 1 & 537 \\
                              & 2 & 537 \\
                              & 3 & 537 \\
                              & 4 & 537 \\
\midrule
\multirow{4}{*}{uuf150-645}   & 1 & 644 \\
                              & 2 & 644 \\
                              & 3 & 644 \\
                              & 4 & 643 \\
\midrule
\multirow{4}{*}{uuf175-753}   & 1 & 752 \\
                              & 2 & 752 \\
                              & 3 & 752 \\
                              & 4 & 752 \\
\midrule
\multirow{4}{*}{uuf200-860}   & 1 & 858 \\
                              & 2 & 858 \\
                              & 3 & 859 \\
                              & 4 & 859 \\
\midrule
\multirow{4}{*}{uuf225-960}   & 1 & 959 \\
                              & 2 & 959 \\
                              & 3 & 958 \\
                              & 4 & 958 \\
\midrule
\multirow{4}{*}{uuf250-1065}  & 1 & 1064 \\
                              & 2 & 1063 \\
                              & 3 & 1064 \\
                              & 4 & 1063 \\
\bottomrule
\end{tabular}
\end{table}
\section{Detailed result for Unsatisfiable benchmark} \label{supp_sec_unsatisfiable_result}
The Table~\ref{tab:supp-satlib-results} summarizes the best SAT achieved by our neuromorphic higher-order Ising machine on the unsatisfiable SATLIB instances of sizes 100–250 \cite{SATLIB}.

\section{Mapping to Dendritic computing model}\label{autoencoder_to_dendritic}
\begin{figure}[ht!]
\centering
\includegraphics[width=0.5\textwidth]{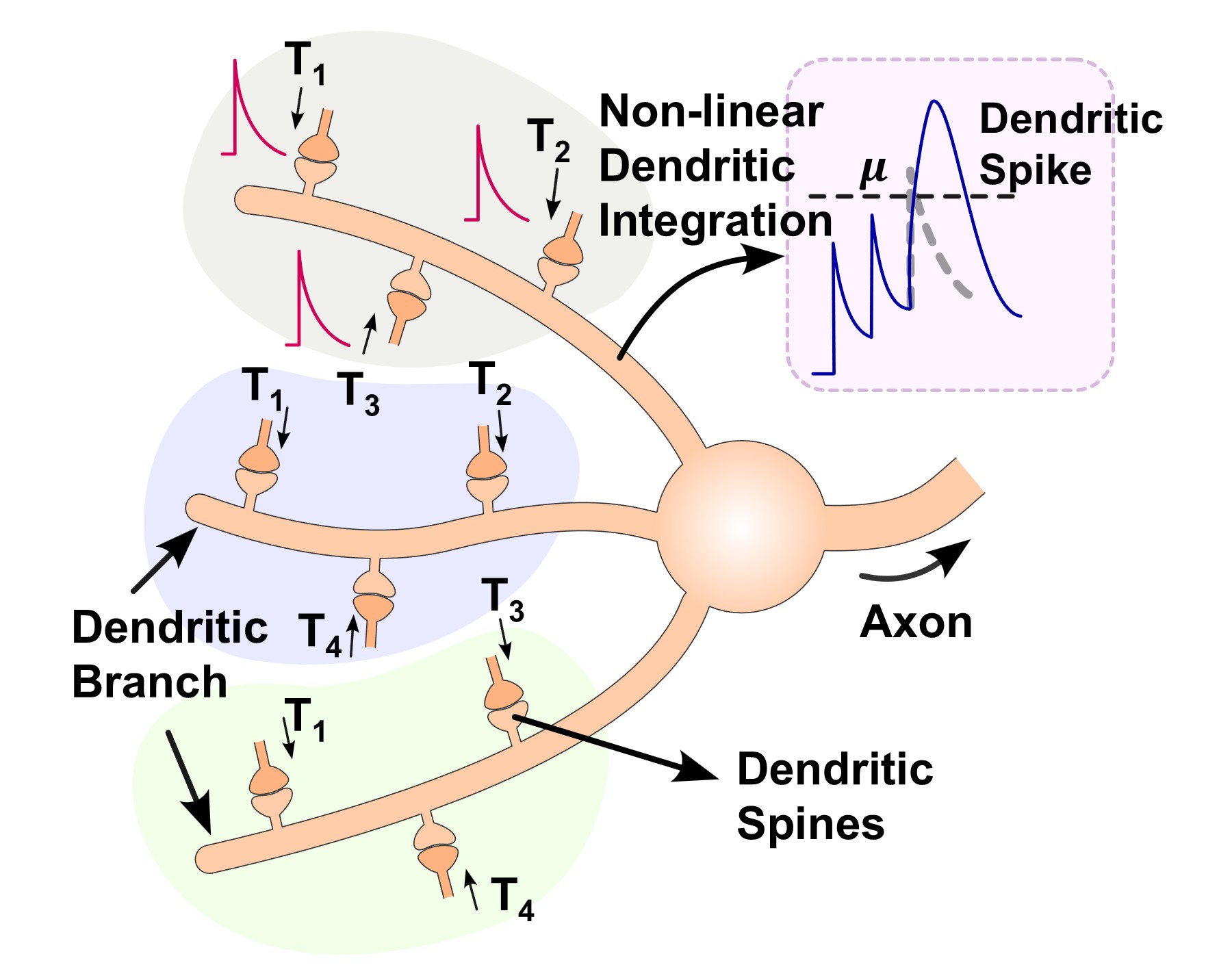}
\caption{\textit{Illustration of higher-order computation in dendritic branches:} The soma accumulates inputs from three different dendritic branches to generate one of the clause outputs ($T_k$) having a third-order interaction of three contributing variables. Each dendritic branch aggregates three synaptic inputs ($T_k$) (where each input can correspond to any order of interaction of dendritic branches) via dendritic spines and performs non-linear amplification after thresholding.
}\label{supplementary_dendrite}
\end{figure}

This section details how the proposed autoencoder architecture can be interpreted as a dendritic computing model, as introduced in Discussion section~\ref{sec_discussion}. Fig.~\ref{supplementary_dendrite} illustrates this mapping for the clause \(T_1\) within the architecture depicted in Fig.~\ref{fig1}.d. In this mapping, each dendritic branch shown in Fig.~\ref{supplementary_dendrite} corresponds directly to one neuron in the latent layer.

The weighted aggregation of preceding clause outputs \(T_{k,n-1}\) at each latent neuron during the encoder stage corresponds directly to synaptic integration along an individual dendritic branch.  
Similarly, the soma representing a given clause, which collects inputs from the associated dendritic branches, is equivalent to decoding the clause output \(T_{k,n}\) from its contributing latent neurons.  
When the total synaptic input on a dendritic branch exceeds a threshold, the membrane potential is nonlinearly amplified rather than simply summing passively. Similar to our architecture (Fig.~\ref{fig1}.d), the soma also applies a somatic threshold \(\theta\) to filter out spurious fluctuations in the dendritic branches. This ensures that once the summed dendritic potential crosses \(\theta\), we interpret it as a discrete dendritic spike event.  
We can see this mechanism in the inset of Fig.~\ref{supplementary_dendrite}.  

A notable feature of practical dendritic computing-based architectures is its sensitivity to temporal dynamics and location-dependent attenuation which can strongly influence whether a dendritic branch will spike. However, for our purposes, we abstract away these fine-grained temporal details. Specifically, we assume that as long as the cumulative input arriving at a branch within a predefined integration window surpasses the branch threshold, the exact order or precise timing of individual synaptic events need not be tracked. This can be readily seen from the inset of Fig.~\ref{supplementary_dendrite}. In other words, our mapping imposes a coarse-grained synchrony requirement—each dendritic branch integrates inputs over a fixed time window and treats them equivalently regardless of arrival order. This simplification yet preserves the essential computational property of dendrites (nonlinear, local thresholding). Beyond this assumption of synchronized integration windows, all other operations in our autoencoder map directly onto known stages of dendritic computation.

\end{document}